\documentclass[preprint,14pt]{elsarticle}

\usepackage{mathrsfs}
\usepackage{amssymb}
\usepackage{amsmath}
\usepackage{multirow}
\allowdisplaybreaks[4]
\usepackage{bm}
\usepackage{color}
\usepackage{exscale}
\usepackage{relsize}
\usepackage{graphicx}
\usepackage{wrapfig}
\usepackage{multicol}
\usepackage{verbatim}
\usepackage[nodots]{numcompress}

\usepackage{float}
\usepackage{graphicx}
\usepackage{subcaption}

\usepackage{algorithm}
\usepackage{algorithmic}

\usepackage[nodots]{numcompress}
\usepackage[colorlinks,linkcolor=blue]{hyperref}

\journal{}

\begin{document}

\begin{frontmatter}
	
	
	
	\title{SO-PIFRNN: Self-optimization physics-informed Fourier-features randomized neural network for solving partial differential equations}
	
	\author[label1]{Jiale Linghu}\ead{mathlhjl@163.com}
	\author[label1]{Weifeng Gao}\ead{gaoweifeng2004@126.com}
	\author[label1]{Hao Dong}\ead{donghao@mail.nwpu.edu.cn}
	\author[label4]{Yufeng Nie}\ead{yfnie@nwpu.edu.cn}
	
	\address[label1]{School of Mathematics and Statistics, Xidian University, Xi'an 710071, PR China}
	\address[label4]{School of Mathematics and Statistics, Northwestern Polytechnical University, Xi'an 710129, PR China}
	
	\begin{abstract}
		
		This study proposes a self-optimization physics-informed Fourier-features randomized neural network (SO-PIFRNN) framework, which significantly improves the numerical solving accuracy of PDEs through hyperparameter optimization mechanism. The framework employs a bi-level optimization architecture: the outer-level optimization utilizes a multi-strategy collaborated particle swarm optimization (MSC-PSO) algorithm to search for optimal hyperparameters of physics-informed Fourier-features randomized neural network, while the inner-level optimization determines the output layer weights of the neural network via the least squares method. The core innovation of this study is embodied in the following three aspects: First, the Fourier basis function activation mechanism is introduced in the hidden layer of neural network, which significantly enhances the ability of the network to capture multi-frequency components of the solution. Secondly, a novel derivative neural network method is proposed, which improves the calculation accuracy and efficiency of PIFRNN method. Finally, the MSC-PSO algorithm of the hybrid optimization strategy is designed to improve the global search ability and convergence accuracy through the synergistic effect of dynamic parameter adjustment, elitist and mutation strategies. Through a series of numerical experiments, including multiscale equations in complex regions, high-order equations, high-dimensional equations and nonlinear equations, the validity of SO-PIFRNN is verified. The experimental results affirm that SO-PIFRNN exhibits superior approximation accuracy and frequency capture capability.
		
	\end{abstract}
	
	
\end{frontmatter}

\section{Introduction}

Partial differential equations (PDEs), as a cornerstone of mathematical physics modeling, hold significant application value across engineering and scientific domains, including fluid dynamics, multiscale analysis of composite materials and electromagnetic field computations \cite{cai2021physics, linghu2025higher, linghu2025higherDRM, khan2022physics}. Traditional numerical methods—such as the finite element method (FEM), finite difference method (FDM) and finite volume method (FVM) rely on structured mesh discretization. However, their accuracy is inherently limited by grid quality and the order of discretization schemes, leading to computational bottlenecks in complex geometries. These approaches encounter computational bottlenecks when addressing complex geometric domains or high-dimensional problems. Over the past few years, the neural network technology has been developed rapidly. These methods not only circumvent the grid-generation dependency inherent to traditional methods but also demonstrate remarkable adaptability to high-dimensional, nonlinear and inverse problems.

In recent years, the method of solving PDEs based on neural network has become a prominent research frontier. Physics-informed neural networks (PINNs) \cite{raissi2019physics} have established a hybrid framework that synergistically integrates physical laws with data-driven learning by embedding residual constraints from PDE governing equations, boundary conditions, and initial conditions into the loss function architecture. To address specific challenges encountered by PINNs in engineering applications, researchers have proposed a variety of improved architectures: Parareal Physics-informed neural network (PPINN) \cite{meng2020ppinn}, Bayesian Physics-informed neural networks (B-PINNs) \cite{yang2021b}, Extended physics-informed neural networks (XPINNs) \cite{jagtap2020extended}, gradient-enhanced Physics-informed neural networks (gPINNs) \cite{yu2022gradient}. E et al.\cite{yu2018deep} proposed the deep Ritz method (DRM), which converts the strong form of PDEs into variational formulations through energy functional minimization and embeds these principles into neural network loss functions. This framework demonstrates significant efficacy in solving high-dimensional problems and eigenvalue problems. Building upon the deep Ritz method, Ming et al.\cite{liao2019deep} proposed the deep Nitsche method, which rigorously enforces boundary conditions through weak formulation embedding. In the context of complex geometries, the researcher has proposed various hybrid frameworks that integrate traditional numerical methods with emerging strategies \cite{sun2023binn, sukumar2022exact, sheng2021pfnn}. These frameworks aim to address discretization challenges arising from irregular boundary conditions. Notably, the neural network method exhibit inherent limitations in resolving multiscale phenomena due to spectral bias in standard activation functions. To address this limitation, some research through spectral analysis to solution spaces to develop Fourier feature-enhanced architectures \cite{wang2021eigenvector, ng2024spectrum} to effectively compute multiscale problems. Although the methods of physical information-driven classes have been applied to compute various equations. However, a major drawback of neural network-based approaches compared to traditional numerical methods is the high computational cost associated with the training phase. The traditional neural network methods require continuous gradient descent updates to optimize weights and biases, which typically consume significant computational time.

The randomized neural network (RNN) method, also known as the extreme learning machine (ELM) method, distinguishes itself from traditional fully connected neural network methods by randomly assigning and fixing the hidden layer weights and biases. The weights of output layer are directly determined via the least squares method. This approach eliminates the need for iterative gradient descent optimization, thereby achieving rapid weight determination and enhancing computational efficiency. Dwivedi et al.\cite{dwivedi2020physics} proposed a physics-informed extreme learning machine method capable of efficiently solving partial differential equations. Dong et al.\cite{dong2021local, dong2022numerical} developed local extreme learning machines (locELM) framework designed to efficiently solve both linear and nonlinear PDEs. Chen et al.\cite{chi2024random, chen2022bridging} proposed the random feature method (RFM) to solve interface problems and complex geometry problems. Automatic differentiation methods are widely used for computing derivatives of equations. However, with the increase of the calculated derivative, the automatic differentiation method will generate huge computational costs. Wang et al.\cite{sun2024local, shang2023randomized, li2025local} employed finite difference methods to approximate derivatives, thereby enhancing computational efficiency in derivative calculations. They developed a weak-form randomized neural network integrated with Petrov-Galerkin methods for solving PDEs and a strong-form local randomized neural network to address interface problems.

The architecture of neural networks and the configuration of hyperparameters significantly influence computational accuracy. At present, the design of neural network parameters primarily relies on existing knowledge and empirical experience. Nevertheless, in the area of automated optimization, several innovative research approaches have emerged in recent years, including random search algorithms, population-based evolutionary computation algorithms and reinforcement learning frameworks \cite{li2020random, linghu2024self, zoph2016neural}. Wang et al.\cite{wang2024pinn} introduced a novel framework termed NAS-PINN, which integrates neural architecture search (NAS) into the PINN framework. By combining these methodologies, the proposed approach automatically identifies optimal neural architectures for solving PDEs. This study \cite{escapil2023hyper} proposes a Gaussian process-based Bayesian optimization framework for hyperparameter optimization (HPO) in PINNs, specifically applied to bounded-domain Helmholtz equation solutions. Zhang et al.\cite{zhang2024discovering} proposed an innovative evolutionary computation method to investigate the structural optimization of PINN models and the design of parameterized activation functions. This approach aims to enhance the approximation accuracy and accelerate the convergence rate for solving PDEs. In the framework of PINNs, the number and distribution of sample points within the computational domain and on the boundaries significantly influence the computational accuracy of the surrogate model. These studies \cite{nabian2021efficient, wu2023comprehensive} delve into the impact of sample point strategies on computational accuracy. Under the current research paradigm, neural architecture search methods primarily focus on the application of the PINNs framework in the field of solving PDEs. In contrast, there is still a research gap in the algorithmic innovation of RNN for solving PDEs. Within the RNN algorithmic computational framework, numerical accuracy is influenced by several interrelated factors: neural network hyperparameters, the distribution of sample points in the computational domain and on boundaries, and the weighting of boundary conditions in linear systems. These factors collectively determine the accuracy of the model during the solution process.

In this paper, we propose a self-optimization physics-informed Fourier-features randomized neural network framework. By incorporating evolutionary computation methods, physics-informed Fourier-features randomized neural network enables an automatic search for optimal hyperparameters, thereby enhancing the accuracy of the surrogate model. Extensive numerical validations demonstrate the effectiveness of the proposed method. The primary contributions of this study can be summarized as follows:
\begin{itemize}
	\item The self-optimization physics-informed Fourier-features randomized neural network (SO-PIFRNN) framework is proposed, which can automatically search the hyperparameters within the PIFRNN method to improve the calculation accuracy. Additionally, the introduction of Fourier activation functions improves the ability of the neural network to approximate multi-frequency and multiscale solutions.
	\item The derivative neural network method are proposed for computing derivatives in PDEs. By comparing it with automatic differentiation and finite difference methods, the numerical experiments demonstrate the superior accuracy and efficiency of the proposed derivative neural network method.
	\item The multi-strategy collaborated particle swarm optimization algorithm is introduced, which can improve the global search ability and convergence efficiency.
\end{itemize}

The remainder of this paper is structured as follows. In section \ref{sec:2}, we first introduce the physics-informed Fourier-features randomized neural network (PIFRNN) and derivative neural network methodology. Next, a multi-strategy collaborated particle swarm optimization (MSC-PSO) algorithm is presented. Finally, the SO-PIFRNN framework implementation is described in detail. Section \ref{sec:3} systematically validates the proposed method through extensive numerical experiments, including multiscale equations, high-order equations, high-dimensional equations and nonlinear equations. The concluding remarks and future research directions are provided in section \ref{sec:4}.

\section{Methodology}
\label{sec:2}
In this section, the physics-informed Fourier-features randomized neural network (PIFRNN) is introduced for solving the 2D Poisson equation. Additionally, we develop a novel derivative neural network method to enhance both computational efficiency and accuracy. Then, a multi-strategy collaborated particle swarm optimization algorithm (MSC-PSO) is proposed to improve the search ability and convergence speed of the algorithm. Finally, we provide a detailed overview of the complete algorithm framework for the self-optimization physics-informed Fourier-features randomized neural network (SO-PIFRNN).

\subsection{Physics-informed Fourier-features randomized neural network (PIFRNN)}
\label{sec:21}
The randomized neural network, also known as extreme learning machine (ELM), is a type of fully connected neural network. In contrast to traditional fully connected neural networks, for a randomized neural network with $M$ hidden layers, the weights and biases of the first $M-1$ layers are randomly initialized and fixed, while only the weights of the final layer are determined via least squares method. In this study, we focus on the single hidden layer randomized neural network with $N$ neurons in the hidden layer, which is defined as follows:
\begin{equation}
	\label{eq:1}
	\begin{aligned}
		&{\varphi _i}(\bm{x}) = \rho \Big( {\sum\limits_{j = 1}^k {{w_{ij}}{x_j} + {b_i}} } \Big),\quad 1 \le i \le N,\\
		&{u_\rho }(\bm{x}) = \sum\limits_{i = 1}^N {{\alpha _i}{\varphi _i}(\bm{x})},
	\end{aligned}
\end{equation}
where $\bm{x}=\left[ {{x_1};{x_2};...;{x_k}} \right]$ denotes the input of neural network, $\rho \left( \bullet \right)$ is the activation function, ${w_{ij}}$ and ${b_i}$ are the weights and biases of the hidden layer that are randomly generated, ${\varphi _i}(\bm{x})$ represents the output of the $i$-th neuron of the hidden layer, $\bm{\alpha}=\left[ {{\alpha_1};{\alpha_2};...;{\alpha_N}} \right]$ is the weights of the output layer that are calculated by the least squares method.

Then, we introduce the physics-informed randomized neural network method for solving Poisson equations. This approach integrates the governing equations and boundary conditions, into the linear system. The weights of the output layer are then determined through the least squares method. Specifically, we consider the following two-dimensional Poisson equation:
\begin{equation}
	\label{eq:2}
	\begin{aligned}
		& -\Big(\frac{{{\partial ^2}{u}}}{{\partial {x_{1}^2}}} + \frac{{{\partial ^2}{u}}}{{\partial {x_{2}^2}}}\Big)(\bm{x}) = f(\bm{x}), \quad \bm{x} \in \Omega,\\
		&u(\bm{x}) = g(\bm{x}), \quad \bm{x} \in \partial \Omega.
	\end{aligned}
\end{equation}
where $\Omega={\left[ {0,1} \right]^2}$. Given $N_1$ points sampled within the domain and $N_2$ points sampled on the boundary. Substitute ${u_\rho }(\bm{x})$ into the Eq.\eqref{eq:2}, we obtain the following linear system:
\begin{equation}
	\label{eq:3}
	\begin{bmatrix}
		\mathcal{F}\\
		{\lambda }\mathcal{B}
	\end{bmatrix} \bm{\alpha}  = \begin{bmatrix}
		\mathfrak{F}\\
		{\lambda }\mathfrak{B}
	\end{bmatrix} ,
\end{equation}
where the coefficients $\lambda$ represents the tunable hyperparameters of the linear system, strategically designed to balance the relative contributions of the governing equations and boundary conditions. $\mathcal{F}$ and $\mathcal{B}$ are matrices of order $N_1 \times N$ and $N_2 \times N$, respectively. $\mathfrak{F}$ and $\mathfrak{B}$ are vectors of order $N_1$ and $N_2$. Specifically,
\begin{equation}
	\label{eq:4}
	\begin{aligned}
		&{{\mathcal F}_{ij}} = -\Big(\frac{{{\partial ^2}{\varphi _j}}}{{\partial {x_{i1}^2}}} + \frac{{{\partial ^2}{\varphi _j}}}{{\partial {x_{i2}^2}}}\Big)(\bm{x_i}), \qquad {\mathfrak{F}_i} = f(\bm{x_i}),\\
		&{{\mathcal{B}}_{ij}} = {\varphi _j}(\bm{x_i}), \qquad\qquad\qquad\qquad\;\;\; {\mathfrak{B}_i} = g(\bm{x_i}).
	\end{aligned}
\end{equation}

Subsequently, the least squares method is employed to solve the linear system \eqref{eq:3}. The calculated parameter $\bm{\alpha}$ is substituted into the randomized neural network \eqref{eq:1}, and ${u_\rho }$ represents the solution of the Poisson equation. Compared with traditional neural network methods, such as the physics-informed neural network, the deep Ritz methd and the deep Galerkin method, the PIFRNN avoids the need for complex nonlinear iterative processes. Instead, it can obtain the solution of the equation using only the least squares method, significantly reducing computational time while maintaining high accuracy.

In the randomized neural network, $\mathcal{N}_{\rho} = \text{span}\{\varphi_1, \varphi_2, \ldots, \varphi_{N}\}$ constitutes the functional vector space. Therefore, the $\varphi_i$ plays a pivotal role in this framework. To enhance the capability of randomized neural networks in approximating multi-frequency functions, it is crucial to incorporate multi-frequency components into $\mathcal{N}_{\rho}$. Existing studies have demonstrated that incorporating Fourier embedding layers $\left[ {{x}, {sin(x)}, {cos(x)}} \right]$ and multiscale coordinate encoding into neural network inputs enables the networks to handle multi-frequency functions more effectively \cite{wang2021eigenvector, tancik2020fourier}. This approach achieves a reduction in the spectral bias through input-space frequency translation. We employ sin activation function to construct neural architectures in this study. The sin activation function has excellent frequency capture ability and distinct periodicity, and its derivative cos function also has completely consistent properties. In the architecture of randomized neural networks, the weights and biases of hidden layer is initialized via a uniform distribution sampling strategy. Therefore, by modulating the parameter range of the uniform distribution interval $\left[ -\omega, \omega \right]$, spectral domain modulation in the parameter space can be achieved. As demonstrated in Eq.\eqref{eq:1}, this parameter tuning process intrinsically constructs an implicit multiscale mapping of coordinate inputs, thereby enhancing the spectral representation capacity of the function approximation space.

\begin{figure}[!htb]
	\centering
	\begin{minipage}[t]{0.48\textwidth}
		\centering
		\includegraphics[width=\textwidth]{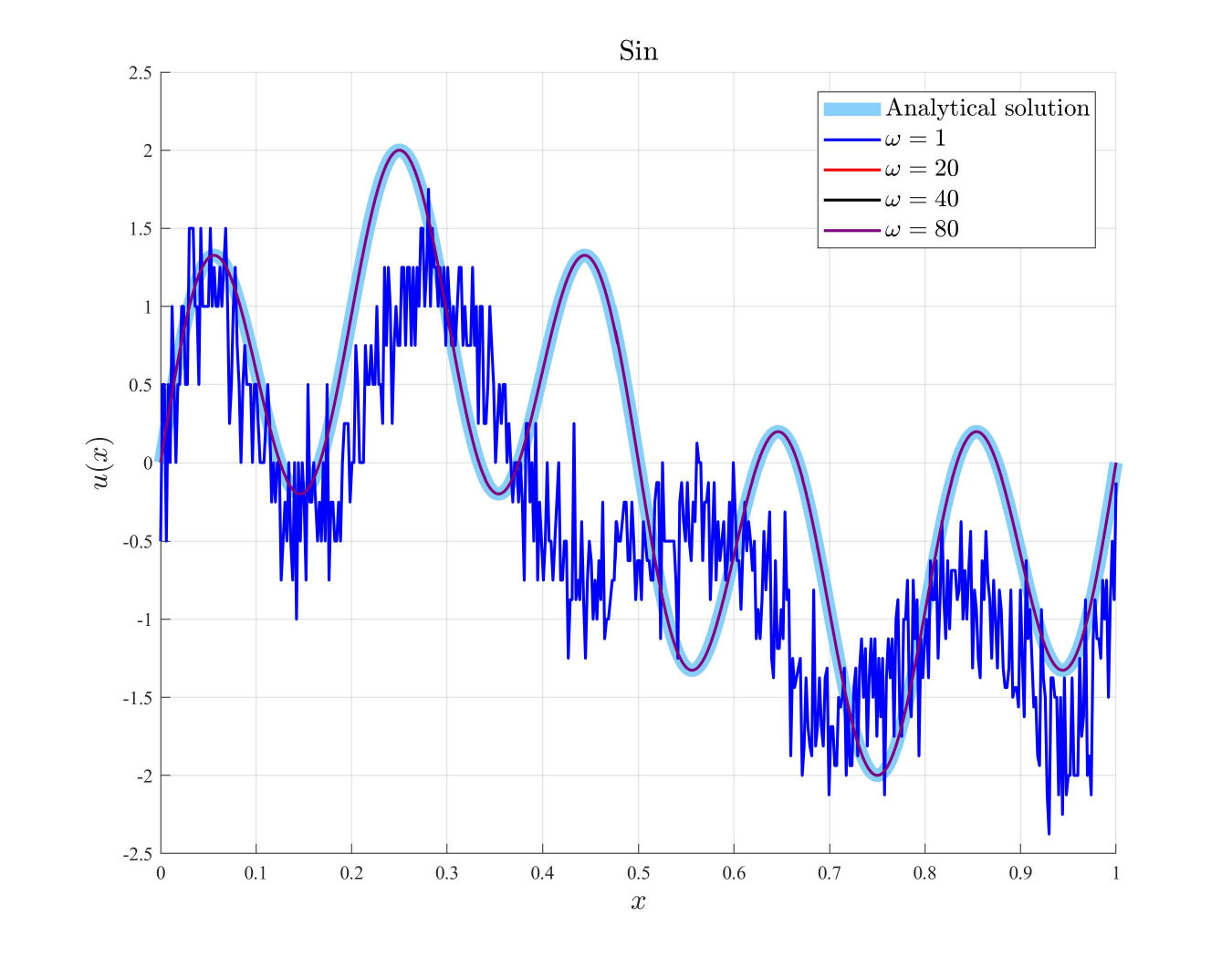}
		(a)
	\end{minipage}
	\hfill
	\begin{minipage}[t]{0.48\textwidth}
		\centering
		\includegraphics[width=\textwidth]{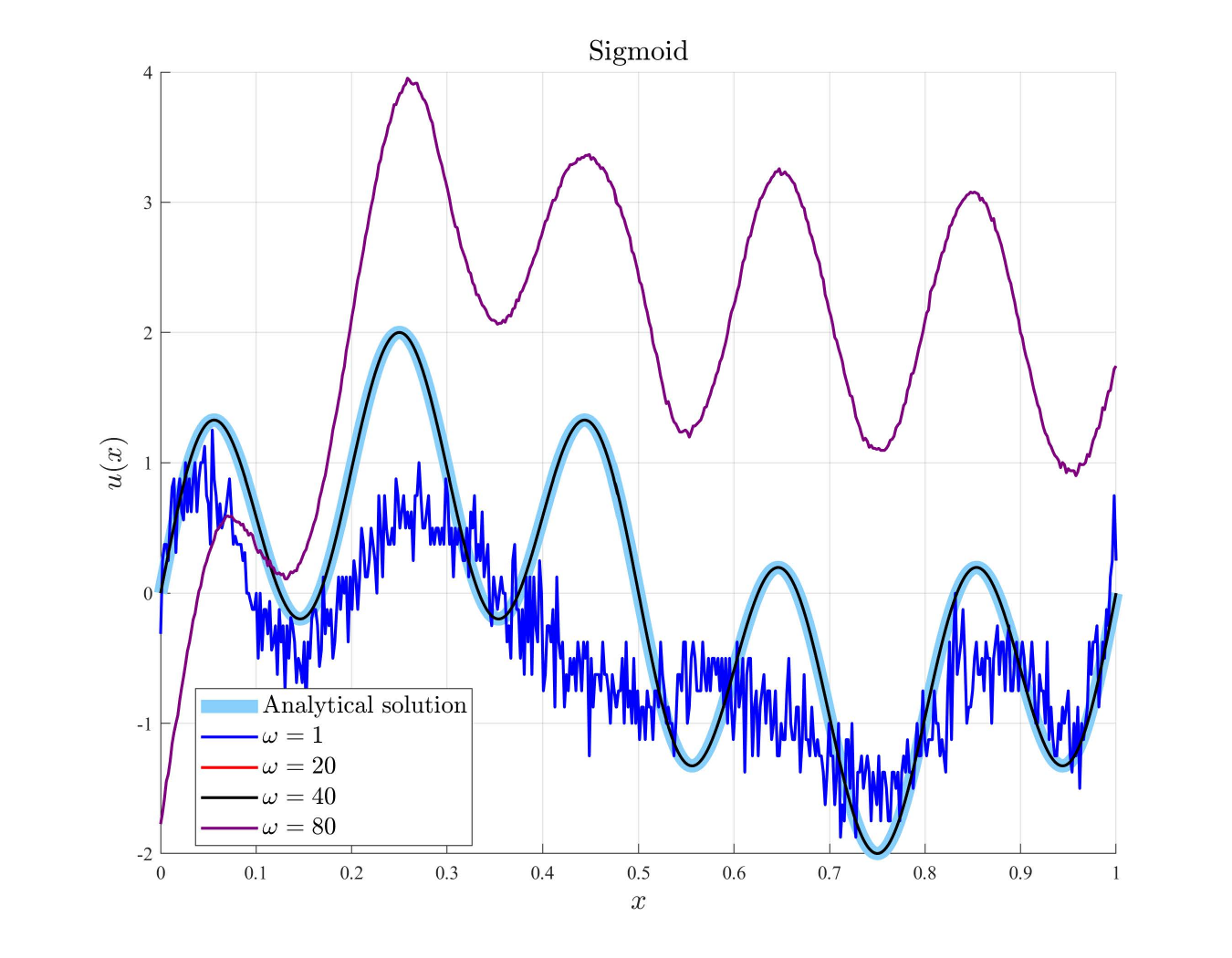}
		(b)
	\end{minipage}
	
	\vspace{0.5em}
	
	\begin{minipage}[t]{0.48\textwidth}
		\centering
		\includegraphics[width=\textwidth]{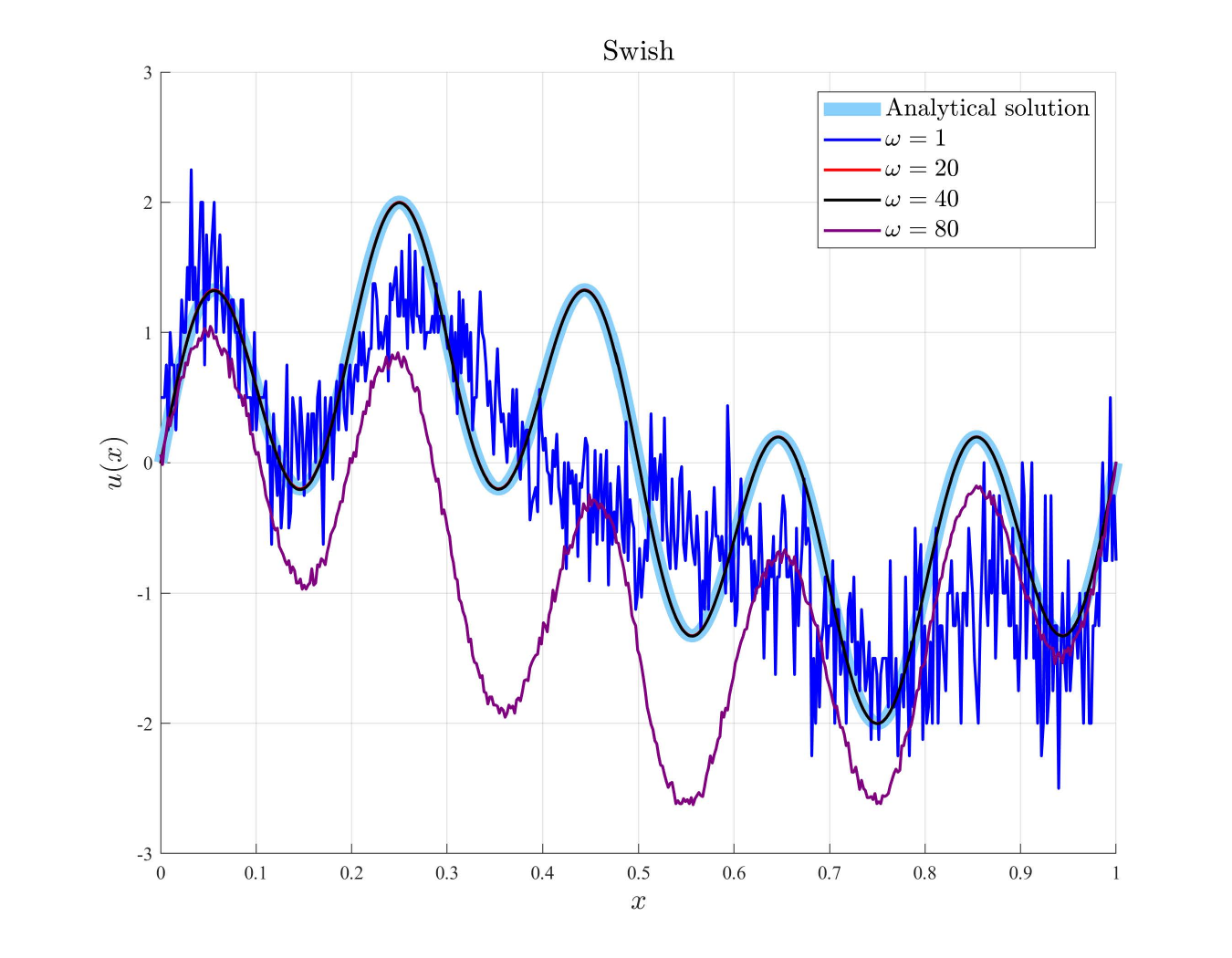}
		(c)
	\end{minipage}
	\hfill
	\begin{minipage}[t]{0.48\textwidth}
		\centering
		\includegraphics[width=\textwidth]{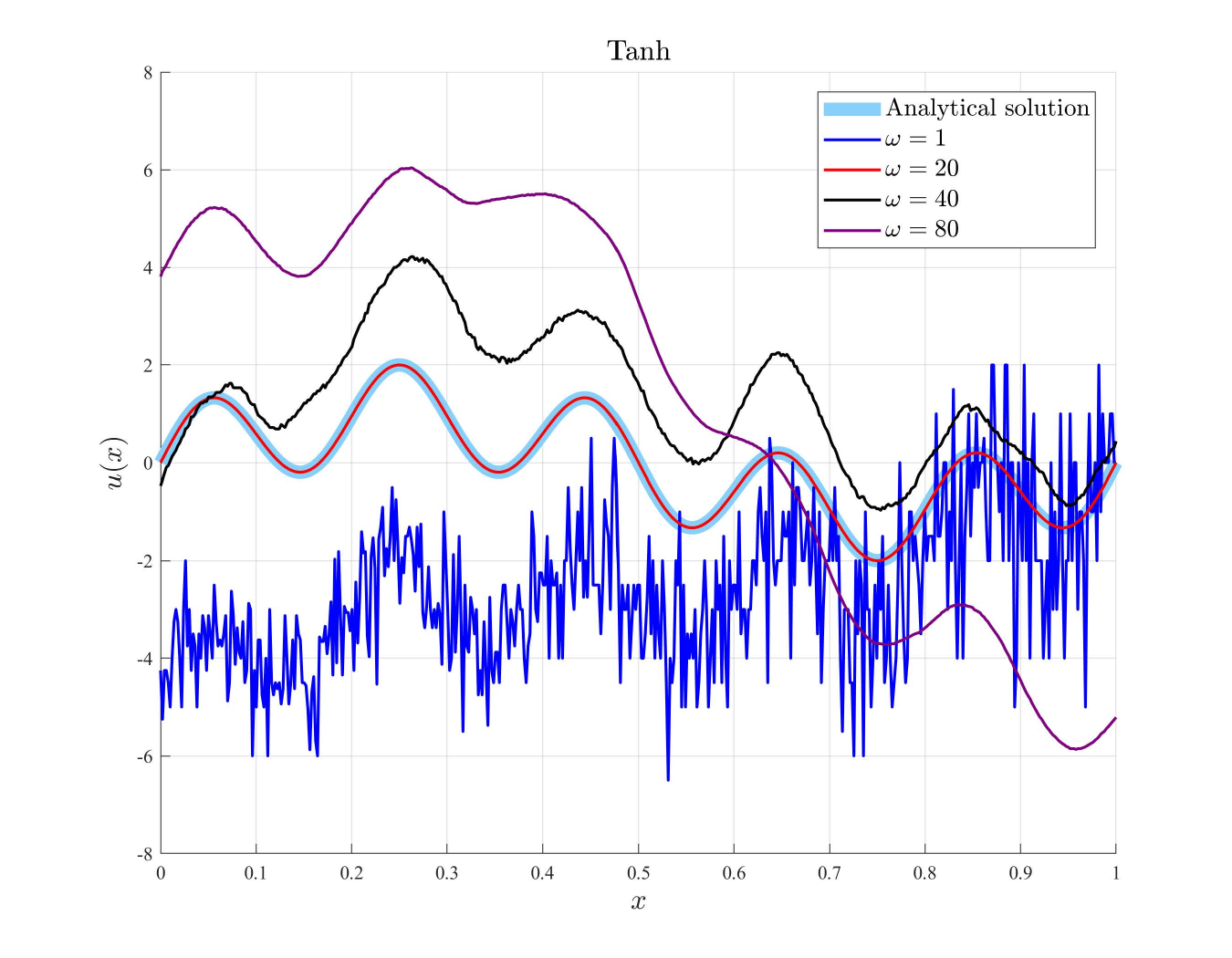}
		(d)
	\end{minipage}
	
	\caption{The computational results using different activation functions and the hidden layer parameter $\omega$ when $\kappa=10$. (a) Sin. (b) Sigmoid. (c) Swish. (d) Tanh.}
	\label{fig:f1}
\end{figure}

\begin{figure}[!htb]
	\centering
	\begin{minipage}[t]{0.48\textwidth}
		\centering
		\includegraphics[width=\textwidth]{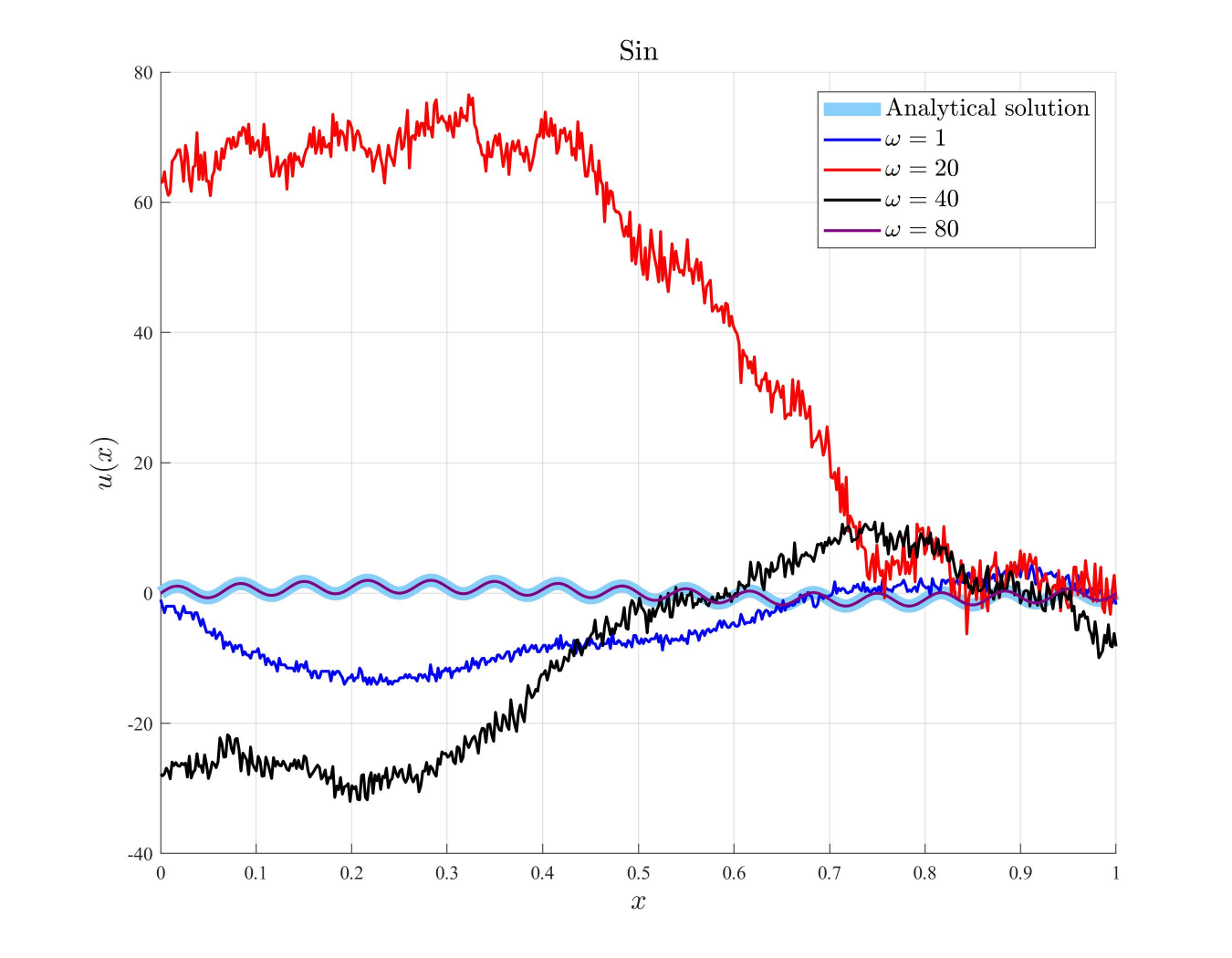}
		(a)
	\end{minipage}
	\hfill
	\begin{minipage}[t]{0.48\textwidth}
		\centering
		\includegraphics[width=\textwidth]{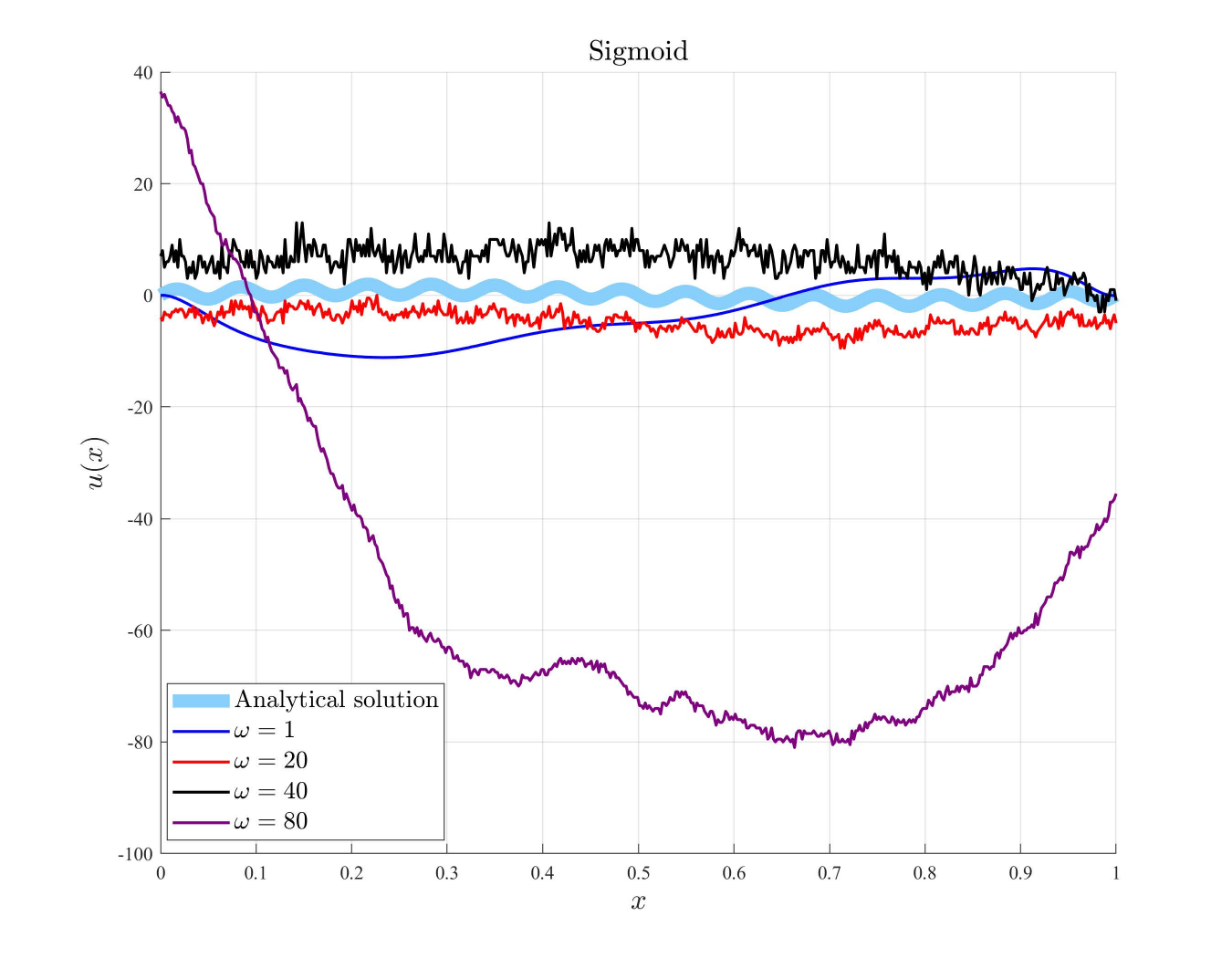}
		(b)
	\end{minipage}
	
	\vspace{0.5em}
	
	\begin{minipage}[t]{0.48\textwidth}
		\centering
		\includegraphics[width=\textwidth]{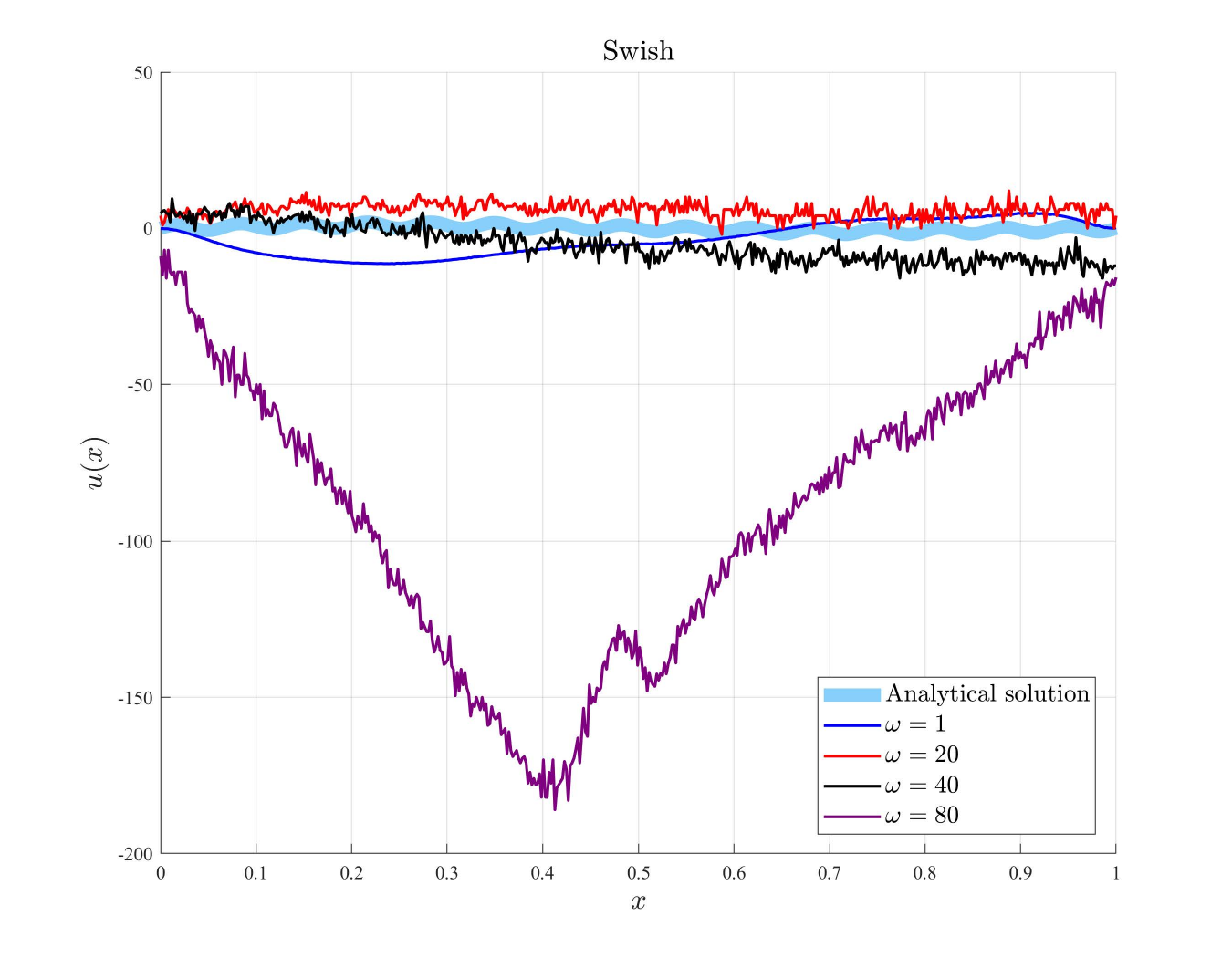}
		(c)
	\end{minipage}
	\hfill
	\begin{minipage}[t]{0.48\textwidth}
		\centering
		\includegraphics[width=\textwidth]{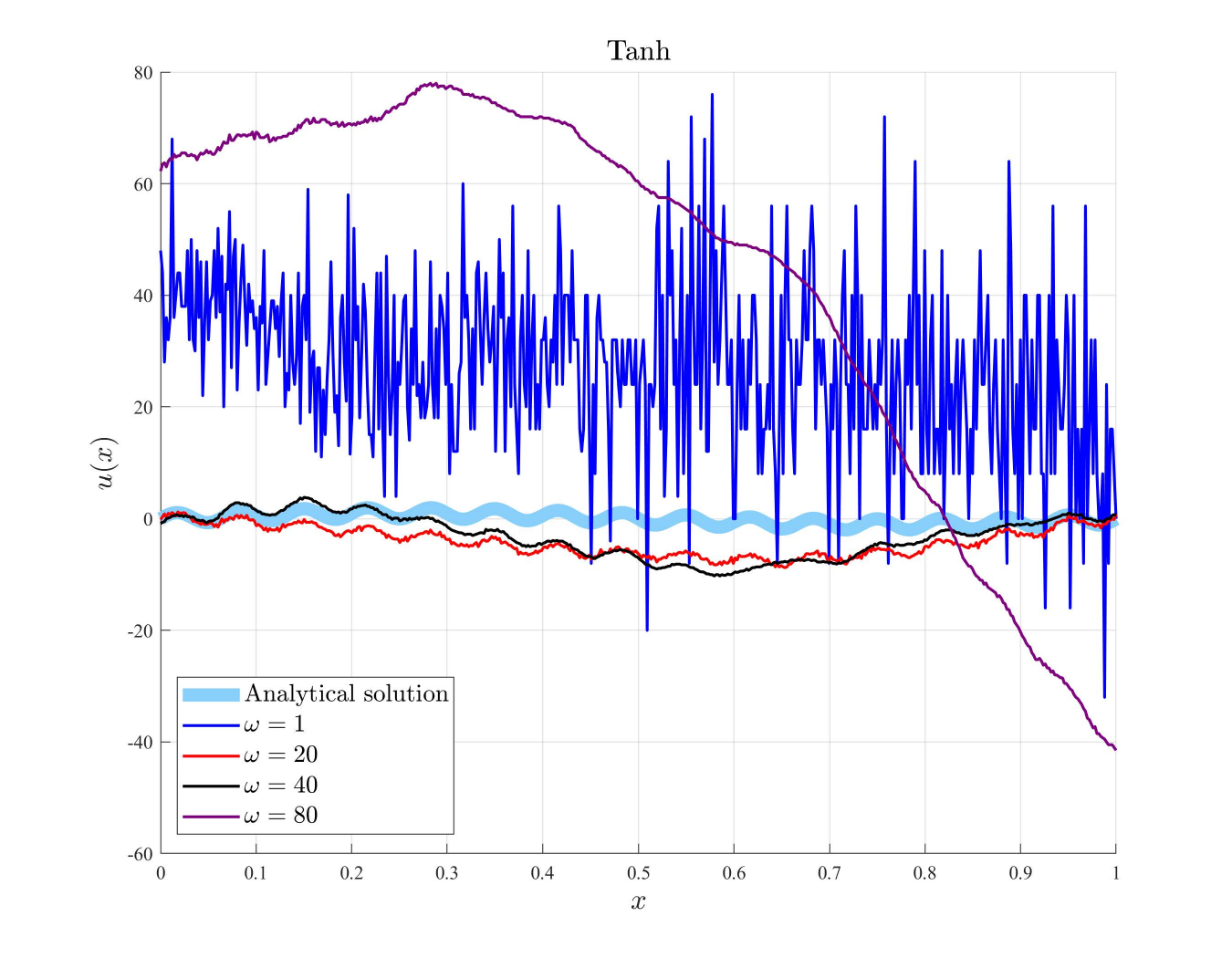}
		(d)
	\end{minipage}
	
	\caption{The computational results using different activation functions and the hidden layer parameter $\omega$ when $\kappa=30$. (a) Sin. (b) Sigmoid. (c) Swish. (d) Tanh.}
	\label{fig:f2}
\end{figure}

We use the PIFRNN with 100 neuron to compute the one-dimensional Poisson equation $- \Delta u\left( x \right) = f\left( x \right),x \in \left[ {0,1} \right]$, which has a multi-frequency nature. Assuming that the solution takes the form of $u\left( x \right) = \sin \left( {2\pi x} \right) + \sin \left( {\kappa \pi x} \right)$. Figs. \ref{fig:f1} and \ref{fig:f2} present the computational results using different activation functions and hidden layer parameters $\omega$. It can be seen that for the low-frequency problem $\kappa=10$, the sin activation function yields poor approximation only when $\omega=1$. In contrast, sigmoid, swish and tanh activation functions perform poorly when $\omega$ is either too large or too small. For the high-frequency problem $\kappa=30$, the sin activation function only provides good approximation when $\omega=80$, while the other activation functions cannot achieve high-precision approximation of high-frequency functions.

In solving the linear system \eqref{eq:3}, the discretized approximation of the differential operator constitutes a key computational bottleneck. The current mainstream approach adopts the gradient calculation paradigm based on automatic differentiation method (ADM) of neural networks. Although this technique can maintain machine precision, it incurs a high time complexity, especially when calculating high-order derivatives. To break through this computational complexity barrier, some research \cite{sun2024local, shang2023randomized, li2025local} focuses on the finite difference method (FDM) approximation: constructing the central difference schemes for the first-order and second-order derivatives as follows:
\begin{equation*}
	\begin{aligned}
		&\frac{\partial u(x_1, x_2)}{\partial x_1} = \frac{u(x_1 + \Delta h, x_2) - u(x_1 - \Delta h, x_2)}{2\Delta h},\\
		&\frac{\partial^2 u(x_1, x_2)}{\partial x_1^2} = \frac{u(x_1 + \Delta h, x_2) - 2u(x_1, x_2) + u(x_1 - \Delta h, x_2)}{\Delta h ^2},
	\end{aligned}
\end{equation*}
where $\Delta h$ is the step size, set the step size $\Delta h = 1.0\times10^{-10}$, $\Delta h = 1.0\times10^{-5}$ of first-order and second-order central difference, respectively. Nevertheless, the finite difference method introduces truncation errors and exhibits a significant escalation in computational complexity when applied to high-order derivatives or high-dimensional problems.

In this study, a novel derivative neural network is proposed for single hidden layer randomized neural networks. It can efficiently and accurately calculate derivatives, being suitable for high-order and high-dimensional problems. For ${{\mathcal F}_{ij}}$, the second derivative of $\bm{x}$ is required. According to Eqs.\eqref{eq:1} and \eqref{eq:4}, ${{\mathcal F}_{ij}}$ can be expressed as follows:
\begin{equation*}
	{{\cal F}_{ij}} =  - \left( {w_{j1}^2{\rho  ''}\left( {{w_{j1}}{x_{i1}} + {w_{j2}}{x_{i2}} + {b_j}} \right) + w_{j2}^2{\rho  ''}\left( {{w_{j1}}{x_{i1}} + {w_{j2}}{x_{i2}} + {b_j}} \right)} \right)(\bm{x_i}),
\end{equation*}
where $\rho  ''$ is the second derivative of the activation function. We define the second-order derivative neural network as follows:
\begin{equation*}
	{\phi _{i}^j} = \frac{{{\partial ^2}{\varphi _i}(\bm{x})}}{{\partial x_j^2}} = w_{ij}^2\rho ''\Big(\sum\limits_{j = 1}^k {{w_{ij}}{x_j} + {b_i}} \Big),\quad 1 \le i \le N.
\end{equation*}

\begin{figure}[!htb]
	\centering
	\includegraphics[width=0.99\linewidth]{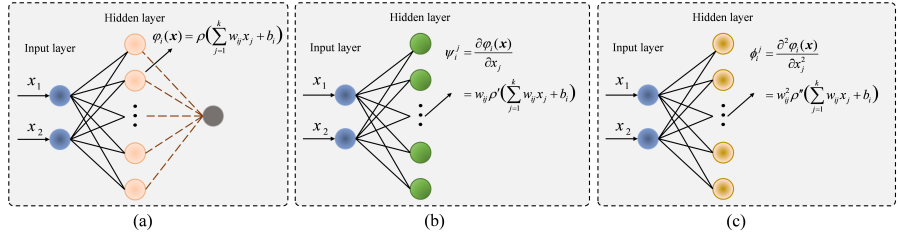}
	\caption{The network architecture diagram. (a) Original network. (b) First-order derivative network. (c) Second-order derivative network.}
	\label{fig:Dl}
\end{figure}

Similarly, the first-order derivative and high-order derivatives of function have similar definitions. Fig.\ref{fig:Dl} shows the original network, the first-order derivative network and the second-order derivative network. Notably, the sin activation function employed in Fourier-features randomization neural networks has the following properties: odd-order derivatives is cos function, whereas even-order derivatives preserve the original sin functional form. Therefore, it is only necessary to construct first- and second- order derivative neural networks. For the $n-th$ order derivative, the corresponding derivative neural network is selected based on the knowledge of the derivatives of the sin function. By constructing derivative neural networks, FPINNs achieve rapid and precise derivative computations in linear systems while maintaining extensibility to high-order and high-dimensional problems. In numerical experiments, we demonstrate the advantages of the method of constructing derivative neural networks by comparing the automatic differentiation method of neural networks and the finite-difference method.

\subsection{A multi-strategy collaborated particle swarm optimization (MSC-PSO) algorithm}

Intelligent optimization algorithms mimic natural or biological processes to search without gradient information, enabling global exploration in high-dimensional hyperparameter spaces. In this section, we first outline the PSO algorithm workflow, and then delve into specific strategies: dynamic parameter adjustment, elitist and mutation. Finally, the multi-strategy collaborated particle swarm optimization algorithm (MSC-PSO) framework is presented.

\subsubsection{Particle swarm optimization (PSO) algorithm}

Particle swarm optimization (PSO) is a classical intelligent optimization algorithm. By simulating social behavior and information-sharing among individuals within a group, the PSO effectively explores the search space, making it highly capable in complex, multi-peak optimization problems. The particle swarm size is $\mathcal{M}$, and the position vector $\mathbf{x}$ and velocity vector $\mathbf{v}$ of each particle are randomly initialized in the search space.
\begin{equation*}
	{x_{i,j}} \sim U\left( {{X_{\min ,j}},{X_{\max ,j}}} \right), \quad {v_{i,j}} \sim U\left( {{-V_{\max ,j}},{V_{\max ,j}}} \right),
\end{equation*}
where $x_{i,j}$ and $v_{i,j}$ represent the position and velocity of the $i$-th particle in the $j$-th dimension, respectively. $X_{\min ,j}$ and $X_{\max ,j}$ represent the upper and lower limits of the position of the particle in the $j$-th dimension. The upper limit of the velocity is ${V_{\max ,j}} = 0.2\left( {{X_{\max ,j}} - {X_{\min ,j}}} \right)$.

Set the maximum number of iterations $T_{max}$. During the iteration process, the formula of $i$-th particle position $\mathbf{x}_i$ and velocity $\mathbf{v}_i$ update is as follows:
\begin{equation*}
	\label{eq:ms1}
	\mathbf{v}^{t+1}_i = \eta  \mathbf{v}^{t}_i + c_1 r_1\left(\mathbf{p}_i - \mathbf{x}^t_i  \right) + c_2 r_2\left(\mathbf{p}_g - \mathbf{x}^t_i  \right),
\end{equation*}
\begin{equation*}
	\label{eq:ms2}
	\mathbf{x}^{t+1}_i = \mathbf{x}^{t}_i + \mathbf{v}^{t+1}_i,
\end{equation*}
where $\mathbf{x}^{t}_i$ and $\mathbf{x}^{t+1}_i$ represent the position of generations $t$ and $t+1$, $\mathbf{v}^{t}_i$ and $\mathbf{v}^{t+1}_i$ represent the velocity of generations $t$ and $t+1$. $\eta $ represents the inertia weight, which is used to balance global exploration and local exploitation. $c_1$ and $c_2$ are learning factors, which represent the ability of the particle to learn from the individual optimal position and the global optimal position, respectively. $r_1$ and $r_2$ are random number in $\left(0, 1\right)$. $\mathbf{p}_i$ is the $i$-th particle individual optimal position. $\mathbf{p}_g$ is the global optimal position.

\subsubsection{The dynamic parameter adjustment strategy}

To address the rigid constraints of classical PSO in managing the exploration-exploitation trade-off, a dynamic parameter adaptive regulation strategy is introduced \cite{melin2013optimal}. This strategy optimizes algorithmic search behavior through the co-evolution of inertia weight $\eta $ and learning factors $\left(c_1, c_2\right)$, enabling phased behavioral shifts: prioritizing global exploration capacity in initial iterations to expand solution space coverage, while progressively intensifying local exploitation precision as evolutionary generations advance to concentrate on high-quality solution regions. The parameter change rules are as follows:
\begin{equation}
	\label{eq:ms3}
	\begin{aligned}
		&\eta  \left( t \right) = {\eta  _{\max }} - \frac{{t\left( {{\eta  _{\max }} - {\eta  _{\min }}} \right)}}{{{T_{\max }}}},\\
		&{c_1}\left( t \right) = {c_{1,\max }} - \frac{{t\left( {{c_{1,\max }} - {c_{1,\min }}} \right)}}{{{T_{\max }}}},\\
		&{c_2}\left( t \right) = {c_{2,\min }} + \frac{{t\left( {{c_{2,\max }} - {c_{2,\min }}} \right)}}{{{T_{\max }}}},
	\end{aligned}
\end{equation}
where set ${\eta  _{\max }}=0.9$, ${\eta  _{\min }}=0.4$, ${c_{1,\max }}=2.5$, ${c_{1,\min }}=0.5$, ${c_{2,\max }}=2.5$ and ${c_{2,\min }}=0.5$.

\subsubsection{The elitist strategy}

In the classical PSO algorithm, particles achieve iterative updates by tracking the individual best position $\mathbf{p}_i$ and the global best position $\mathbf{p}_g$. However, this mechanism is prone to population diversity depletion and premature convergence. In order to improve the search efficiency of the population and maintain population diversity, an elite strategy is introduced \cite{lim2014adaptive}. Specifically, the elite particles are designated as the top 20\% of particles with the lowest fitness values in the current generation. After the introduction of the elite particle strategy, the update formula for velocity is:
\begin{equation}
	\label{eq:ms4}
	\mathbf{v}^{t+1}_i = \eta \left( t \right) \mathbf{v}^{t}_i + c_1\left( t \right) r_1\left(\mathbf{p}_i - \mathbf{x}^t_i  \right) + c_2\left( t \right) r_2\left(\mathbf{p}_g - \mathbf{x}^t_i  \right) + c_3 r_3\left(\mathbf{p}^t_e - \mathbf{x}^t_i  \right),
\end{equation}
where $c_3=0.4$ represents the learning ability of particles towards elite populations, $r_3$ is random number in $\left(0, 1\right)$. $\mathbf{p}^t_e$ is the average position of the elite populations.

\subsubsection{The mutation strategy}

In PSO, the mutation strategy serves as a pivotal technical approach to mitigate premature convergence and enhance global search capability. Its core idea is to break the deterministic pattern of the particle position update formula by artificially introducing random perturbations, thereby enhancing the population diversity. The mutation operation is expressed as:
\begin{equation}
	\label{eq:ms5}
	\mathbf{x}^{t+1}_i = \mathbf{x}^{t}_i + \mathbf{v}^{t+1}_i + \Delta \mathbf{x}_i^t ,
\end{equation}
where $\Delta \mathbf{x}_i^t$ follows a normal distribution $\Delta \mathbf{x}_i^t \sim N\left( {0,\bm{\sigma} _t^2} \right)$, the standard deviation ${\bm{\sigma} _t} = {\bm{\sigma} _{\max }}{e^{ - {t \mathord{\left/{\vphantom {t {{T_{\max }}}}} \right.\kern-\nulldelimiterspace} {{T_{\max }}}}}}$, and ${\sigma _{\max ,j}} = 0.1\left( {{X_{\max ,j}} - {X_{\min ,j}}} \right)$.

\subsubsection{Multi-strategy collaborated particle swarm optimization (MSC-PSO) framework}

Aiming at the defects of the traditional PSO algorithm, such as the lack of population diversity and premature convergence in complex optimization problems, this study proposes a multi-strategy collaborated particle swarm optimization (MSC-PSO). This framework synergistically integrates dynamic parameter adaptation, elite preservation mechanisms and mutation operators strategies to achieve intelligent optimization of the exploration-exploitation trade-off. The algorithm framework of MSC-PSO is as follows:

\begin{algorithm}[!h]
	\caption{MSC-PSO framework}
	\label{alg:MSC-PSO}
	\begin{algorithmic}[1]
		\STATE Initialization: the number of maximum iterations $T_{max}$, the number of particles $\mathcal{M}$, the position $\mathbf{x}^{1}_i$ and velocity $\mathbf{v}^{1}_i$ of particles, and the function of fitness value $f_{val}$. The optimal position of each particle $\mathbf{p}_i$, the elite position $\mathbf{p}^1_e$ and the global optimal position $\mathbf{p}_g$.
		\WHILE{$t<=T_{max}$}
		\STATE Update $\eta$, $c_1$ and $c_2$ by Eq.\eqref{eq:ms3}
		\FOR{$i=1$ to $\mathcal{M}$}
		\STATE Computed velocity $\mathbf{v}^{t+1}_i$ by Eq.\eqref{eq:ms4}
		\STATE Restrict $v^{t+1}_{ij}$ in $[{-{V_{\max }},{V_{\max }}}]$
		\STATE Computed position $\mathbf{x}^{t+1}_i$ by Eq.\eqref{eq:ms5}
		\STATE Restrict $x^{t+1}_{ij}$ in $[{{X_{\min ,j}},{X_{\max ,j}}}]$
		\STATE Computed the fitness value $f_{val}(\mathbf{x}^{t+1}_i)$
		\STATE Uptate $\mathbf{p}_i$
		\ENDFOR
		\STATE t=t+1
		\STATE Update $\mathbf{p}^t_e$
		\STATE Update $\mathbf{p}_g$
		\ENDWHILE		
	\end{algorithmic}
\end{algorithm}

\subsection{Self-optimization physics-informed Fourier-features randomized neural network (SO-PIFRNN)}

\begin{figure}[!htb]
	\centering
	\includegraphics[width=0.95\linewidth]{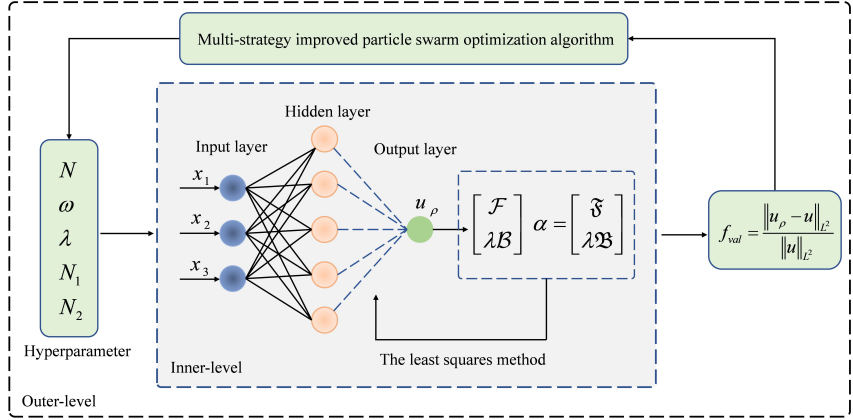}
	\caption{The SO-PIFRNN algorithm bi-level optimization flowchart.}
	\label{fig:Fl}
\end{figure}

The numerical accuracy of the physics-informed Fourier-features randomized neural network is affected by the coupling effect of multi-dimensional hyperparameters, including the hyperparameters of the randomized neural network, the number of sample points and the hyperparameters of the linear system. Traditional hyperparameter selection often rely heavily on expert prior knowledge and trial-and-error approaches, which can be time-consuming and resource-intensive. To overcome these limitations, this study introduces a novel self-optimization physics-informed Fourier-features randomized neural network. This approach aims to reduce dependency on manual tuning and enhance optimization efficiency by integrating self-optimization mechanisms with physics-informed neural network. The algorithm framework is illustrated in Fig.\ref{fig:Fl}.

The algorithm framework adopts a bi-level optimization architecture. The outer-level optimization employs the MSC-PSO to tune hyperparameters critical for the physics-informed Fourier-features randomized neural network in solving PDEs. The inner-level optimization utilizes the least squares method to compute the weights of output layer of PIFRNN. The $L^2$ relative error, denoting the variance between the solution computed by FPIRNN and the analytical solution, is quantified to be \cite{wang2024pinn}:
\begin{equation*}
	{f_{val}} = \frac{{{{\left\| {{u_\rho } - u} \right\|}_{{L^2}}}}}{{{{\left\| u \right\|}_{{L^2}}}}},
\end{equation*}
which serves as the fitness value that provides feedback to the outer-level optimizer. The outer-level optimizer subsequently adjusts the hyperparameters for the inner-level computation based on the fitness value provided by the inner-level.

\section{Numerical experiments and discussions}
\label{sec:3}
In this section, the effectiveness of the self-optimization physics-informed Fourier-features randomized neural network (SO-PIFRNN) is validated through numerical experiments involving multiscale problems in complex geometries, one-dimensional wave equations, Kirchhoff-Love thin plate problems, high-dimensional problems, Lam\'e equations and two-dimensional nonlinear Helmholtz equations. To validate the robustness of the SO-PIFRNN framework, three comparative experiments are designed:
\begin{itemize}
	\item To evaluate the frequency-domain feature extraction capability of the PIFRNN method, comprehensive numerical comparisons are performed among sin activation functions and conventional functions, including sigmoid, swish and tanh activation functions.
	\item The multi-strategy collaborated particle swarm optimization (MSC-PSO) algorithm is implemented for hyperparameter optimization, with systematic performance comparisons conducted against three baseline methods: random search (Rand), genetic algorithm (GA) and traditional particle swarm optimization (PSO).
	\item The accuracy and efficiency of the derivative neural network (Derivative-NN) method are compared with the automatic differentiation method (ADM) and the finite difference method (FDM).
\end{itemize}
All numerical experiments are conducted using double-precision arithmetic and fixed computer random numbers, with computations executed on an Intel(R) Xeon(R) Gold 5218R CPU @ 2.1GHz and an NVIDIA GeForce RTX 3080 GPU.

\subsection{Example 1: multiscale problems in complex geometries}
\label{sec:31}

\begin{figure}[!htb]
	\centering
	\includegraphics[width=0.4\linewidth]{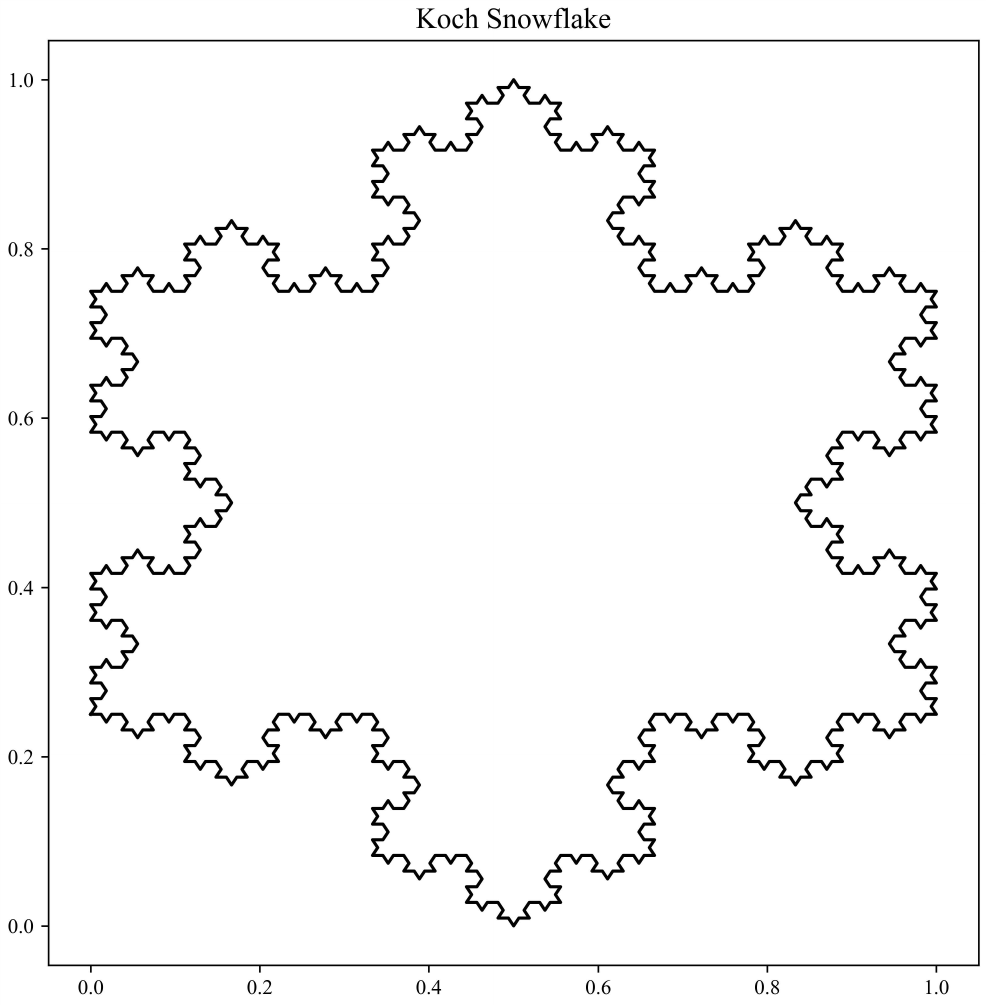}
	\caption{The computation domain for a Koch Snowflake after 5 iterations.}
	\label{fig:ks}
\end{figure}

The Koch snowflake domain, a classical complex geometric domain characterized by its fractal boundary, exemplifies the evolution from a simple equilateral triangle to an infinitely intricate structure through iterative refinement. In this experiment, we investigate the two-dimensional Poisson equation \eqref{eq:2} defined on the Koch snowflake domain \cite{sheng2021pfnn, koch1904courbe}, as illustrated in Fig.\ref{fig:ks}.

We address the Poisson equation \eqref{eq:2} by implementing a source term of
\begin{equation*}
	f(x) = 2{\gamma ^2}\sin (\gamma {x_1})\sin (\gamma {x_2}),\quad \gamma  = 15\pi.
\end{equation*}
The analytical solution can be expressed as:
\begin{equation*}
	u(x) = \sin (\gamma {x_1})\sin (\gamma {x_2}).
\end{equation*}
The Dirichlet boundary condition $g(x) = {\left. {u(x)} \right|_{\partial \Omega }}$ are inherently prescribed by the analytical solution.

Subsequently, the SO-PIFRNN method is employed to solve the Poisson equation on the complex geometries. Specifically, the original Poisson equation \eqref{eq:2} is discretized into the linear system \eqref{eq:3}, where the hyperparameters include: the number of hidden layer neurons $N$, the range of the weight and bias of the input layer of the neural network $\omega$, the coefficient of the boundary constraint term $\lambda$, the number of sampling points $N_1$ in the domain and the number of boundary sampling points $N_2$. The optimization range of each hyperparameter is shown in Table \ref{tab:c1}.

\begin{table}[!htb]
	\centering
	\caption{The optimization range of hyperparameter.}\label{tab:c1}
	\begin{tabular}{ccccc}
		\hline
		Hyperparameters	& $N$ & $\omega$ & $\lambda$ & $N_1$, $N_2$ \\ \hline
		Range	& $\left[ {10,{\rm{ 2000}}} \right]$ &  $\left[ {0.0001,{\rm{ 100}}}\right]$  &  $\left[ {0.0001,{\rm{ 10000}}} \right]$   &    $\left[  {10,{\rm{ 3000}}}\right]$  \\ \hline
	\end{tabular}
\end{table}

\begin{figure}[!htb]
	\centering
	\begin{minipage}[t]{0.24\textwidth}
		\centering
		\includegraphics[width=\textwidth]{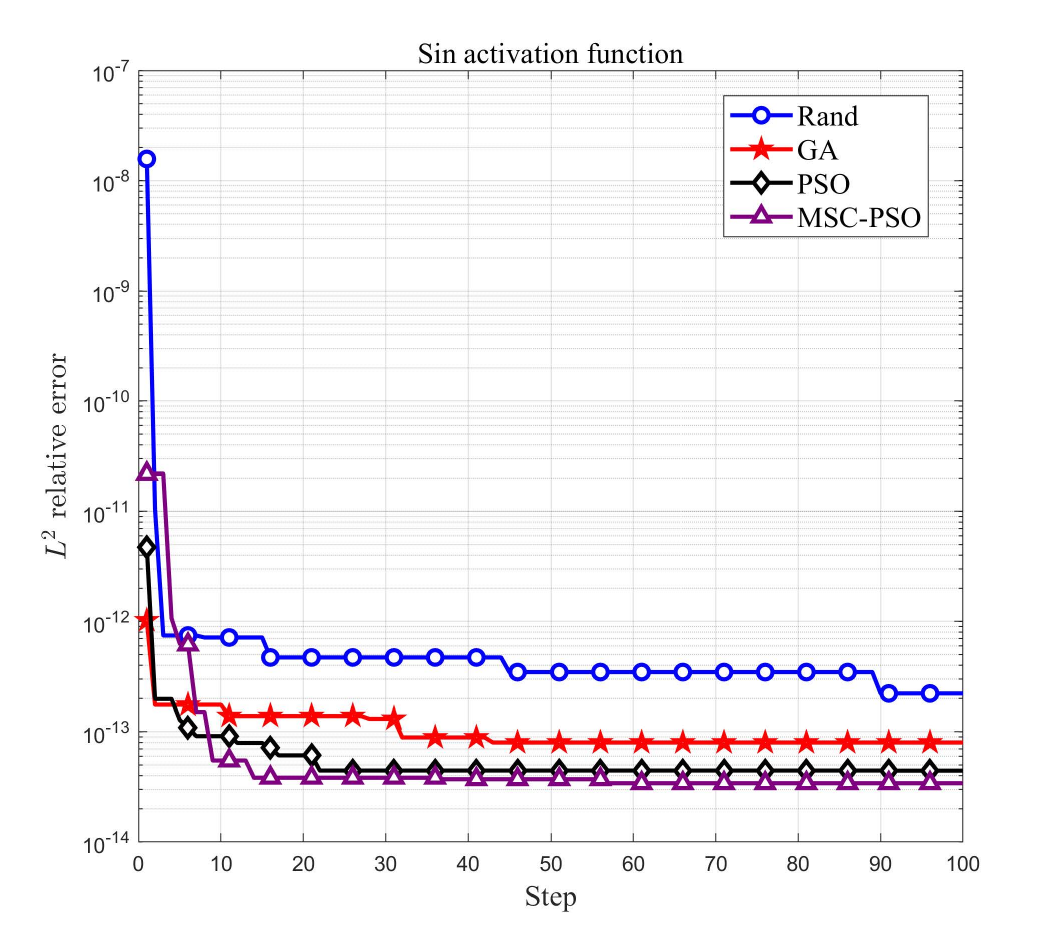}
		(a)
	\end{minipage}
	\hfill
	\begin{minipage}[t]{0.24\textwidth}
		\centering
		\includegraphics[width=\textwidth]{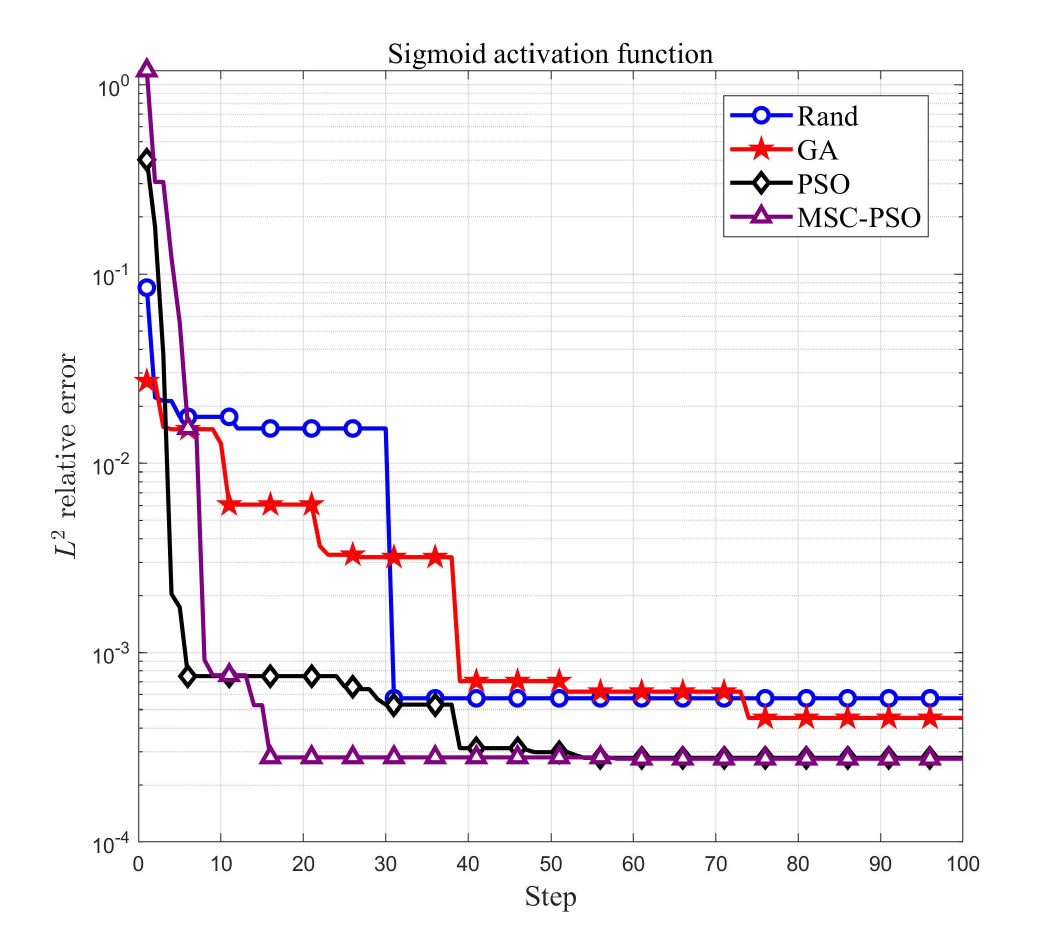}
		(b)
	\end{minipage}
	\hfill
	\begin{minipage}[t]{0.24\textwidth}
		\centering
		\includegraphics[width=\textwidth]{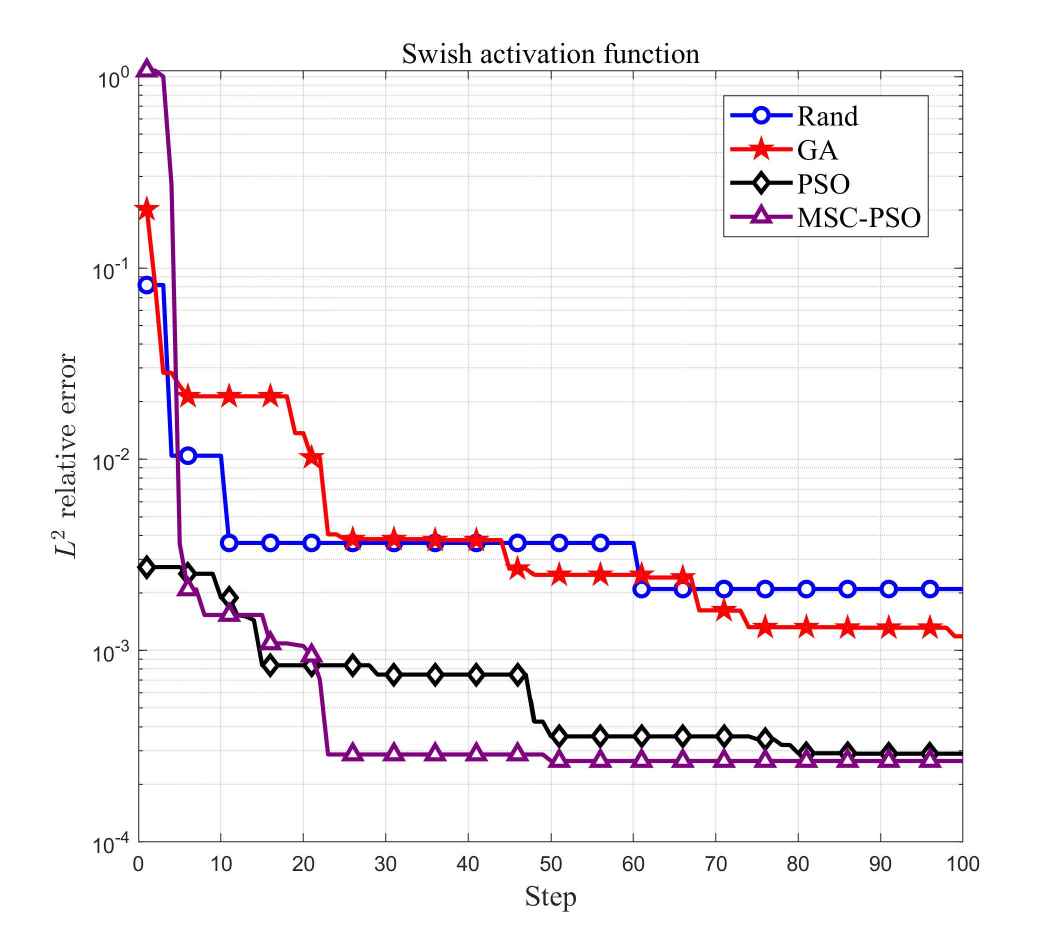}
		(c)
	\end{minipage}
	\hfill
	\begin{minipage}[t]{0.24\textwidth}
		\centering
		\includegraphics[width=\textwidth]{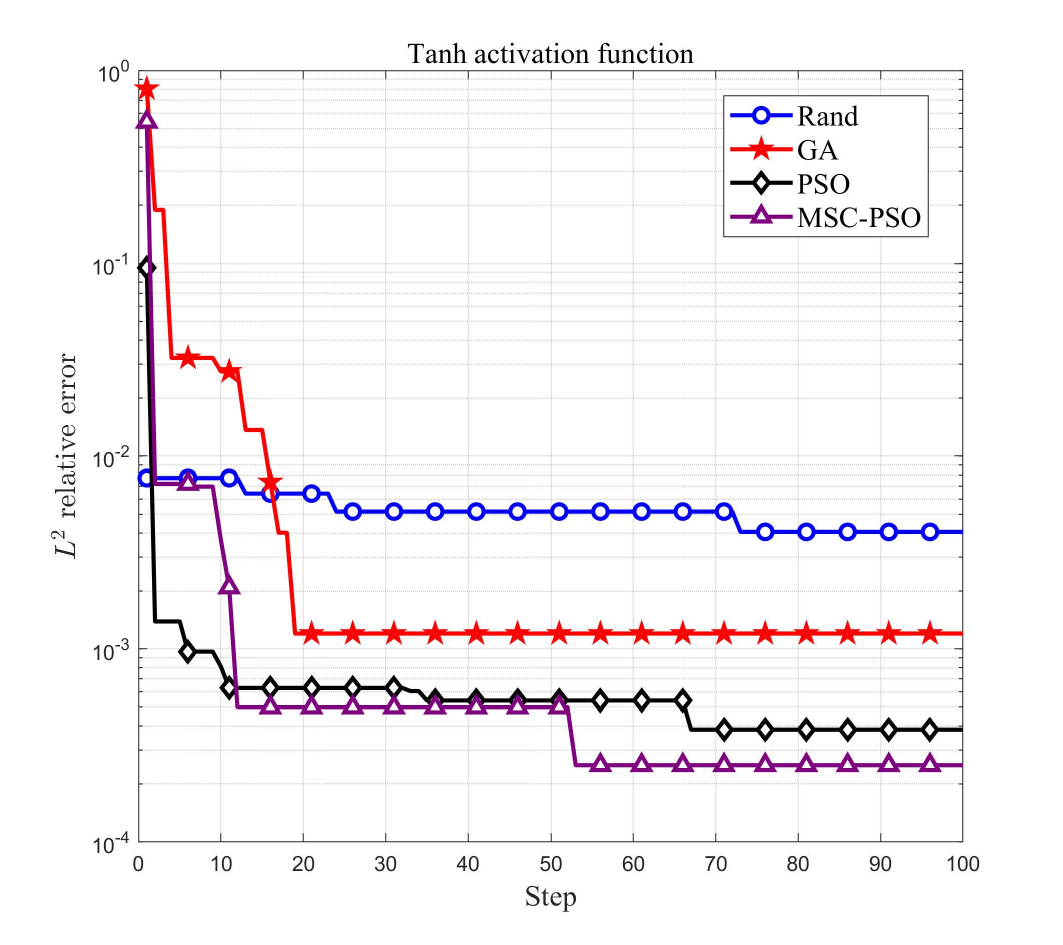}
		(d)
	\end{minipage}
	\caption{The $L^2$ relative error of different activation functions under intelligent different optimization algorithms. (a) Sin. (b) Sigmoid. (c) Swish. (d) Tanh.}
	\label{fig:c1}
\end{figure}

\begin{figure}[!htb]
	\centering
	\includegraphics[width=1\linewidth]{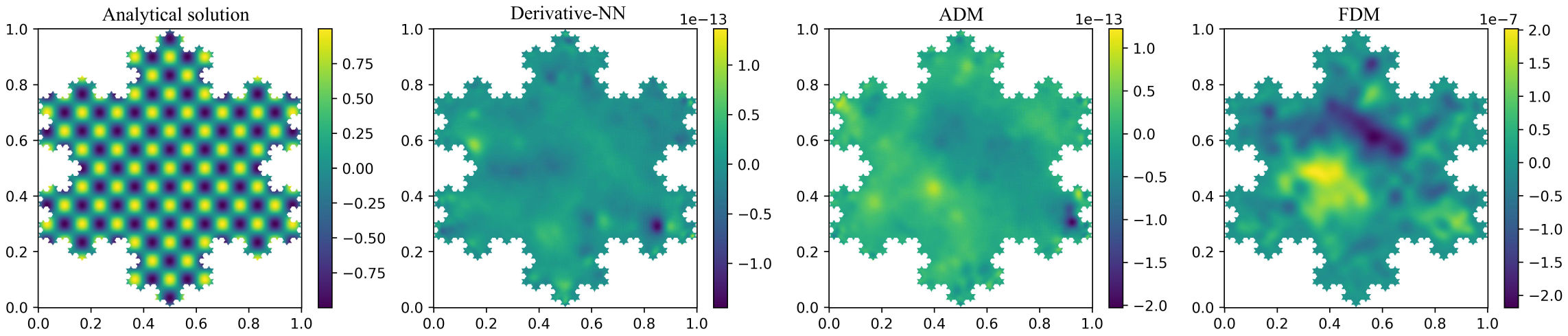}
	\caption{The analytical solution (first column) of multiscale problems in complex geometries and absolution error (last three columns).}
	\label{fig:c2}
\end{figure}

Fig.\ref{fig:c1} presents the convergence profiles of $L^2$ relative error for the Poisson equation solutions obtained using different activation functions and intelligent optimization algorithms. As illustrated in the Fig.\ref{fig:c1}, the MSC-PSO algorithm developed in this work demonstrates accelerated convergence rates and identifies hyperparameter combinations that yield higher computational accuracy. In particular, when using the SO-PIFRNN framework, combined with the frequency domain feature extraction capability of sin activation function, its numerical solution accuracy is improved by ten orders of magnitude compared with sigmoid, swish and tanh activation functions. The optimal hyperparameters of the SO-PIFRNN method are: $N = 2000$, $\omega=69.583$, $\lambda_1 = 2002.105$, $N_1 = 2650$ and $N_2 = 3000$. Table \ref{tab:c2} systematically compares the computational accuracy and efficiency of the derivative neural network (Derivative-NN) method with the automatic differentiation method (ADM) of neural networks and the finite difference method (FDM) under the optimal hyperparameter configuration. From the table, it can be observed that the $L^2$ relative error of Derivative-NN and ADM is nearly identical. Due to truncation errors, the $L^2$ relative error of FDM is significantly higher than those of the other two methods. The Derivative-NN can quickly compute derivatives, achieving computational efficiency comparable to FDM. Compared to ADM, it improves computational efficiency by more than 90\%. Fig.\ref{fig:c2} further shows the analytical solution and the absolute errors between the three numerical methods and the analytical solution. 

\begin{table}[!htb]
	\centering
	\caption{Comparison of calculation accuracy and efficiency.}\label{tab:c2}
	\begin{tabular}{cccc}
		\hline
		& Derivative-NN & ADM & FDM \\ \hline
		$L^2$ relative error & $3.415\times10^{-14}$   &$5.144\times10^{-14}$   &$1.261\times10^{-7}$  \\
		Time/s               & 1.029  &22.282  & 1.089  \\ \hline
	\end{tabular}
\end{table}

\subsection{Example 2: one-dimensional wave equation}

We delve into the one-dimensional wave equation, which is represented in the form of
\begin{equation*}
	\begin{aligned}
		& \frac{{{\partial ^2}u(x,t) }}{{\partial {t}^2}}- 100\frac{{{\partial ^2}u(x,t) }}{{\partial {x}^2}}=0,\quad (x,t) \in (0,1) \times (0,1),\\
		&u(0,t) = u(1,t) = 0,\quad t \in [0,1],\\
		&u(x,0) = \sin (\pi x) + \sin (2\pi x),\quad x \in [0,1],\\
		&\frac{{{\partial}u(x,0) }}{{\partial {t}}} = 0,\quad x \in [0,1].
	\end{aligned}
\end{equation*}
According to d'Alembert's formula \cite{evans2022partial}, the solution $u(x,t)$ can be expressed as
\begin{equation*}
	u(x,t) = \sin (\pi x)\cos (10\pi t) + \sin (2\pi x)\cos (20\pi t).
\end{equation*}

Subsequently, the wave equation is discretized into a linear system. The SO-PIFRNN method inherits the advantages of neural network approaches, enabling direct solution of the wave equation within a unified spatio-temporal framework while circumventing error accumulation caused by time-stepping iterations inherent in traditional numerical methods. Notably, the initial conditions are treated as special boundary conditions within the spatio-temporal domain. The initial and boundary conditions of the original equation can be divided into two types: Dirichlet boundary condition and Neumann boundary condition. Assuming that $N_1$ points are selected within the space-time domain, $N_2$ points are located on the Dirichlet boundary, and $N_3$ points are positioned on the Neumann boundary, the linear system can be expressed as:
\begin{equation}
	\label{eq:10}
	\begin{bmatrix}
		\mathcal{F}\\
		{\lambda _1}\mathcal{B}^1\\
		{\lambda _2}\mathcal{B}^2
	\end{bmatrix} \bm{\alpha}  = \begin{bmatrix}
		\mathfrak{F}\\
		{\lambda _1}\mathfrak{B}^1\\
		{\lambda _2}\mathfrak{B}^2
	\end{bmatrix},
\end{equation}
where $\lambda _1$ and $\lambda _2$ represent the weights of Dirichlet boundary condition and Neumann boundary condition. $\mathcal{F}$, $\mathcal{B}^1$ and $\mathcal{B}^2$ are matrices of order $N_1 \times N$, $N_2 \times N$ and $N_3 \times N$, respectively. $\mathfrak{F}$, $\mathfrak{B}^1$ and $\mathfrak{B}^2$ are vectors of order $N_1$, $N_2$ and $N_3$. Table \ref{tab:m1} presents the optimization range for each hyperparameter.

\begin{table}[!htb]
	\centering
	\caption{The optimization range of hyperparameter.}\label{tab:m1}
	\begin{tabular}{ccccc}
		\hline
		Hyperparameters	& $N$ & $\omega$ & $\lambda_1$, $\lambda_2$ & $N_1$, $N_2$, $N_3$ \\ \hline
		Range	& $\left[ {10,{\rm{ 2000}}} \right]$ &  $\left[ {0.0001,{\rm{ 100}}}\right]$  &  $\left[ {0.0001,{\rm{ 10000}}} \right]$   &    $\left[  {10,{\rm{ 3000}}}\right]$ \\ \hline
	\end{tabular}
\end{table}

\begin{figure}[!htb]
	\centering
	\begin{minipage}[t]{0.24\textwidth}
		\centering
		\includegraphics[width=\textwidth]{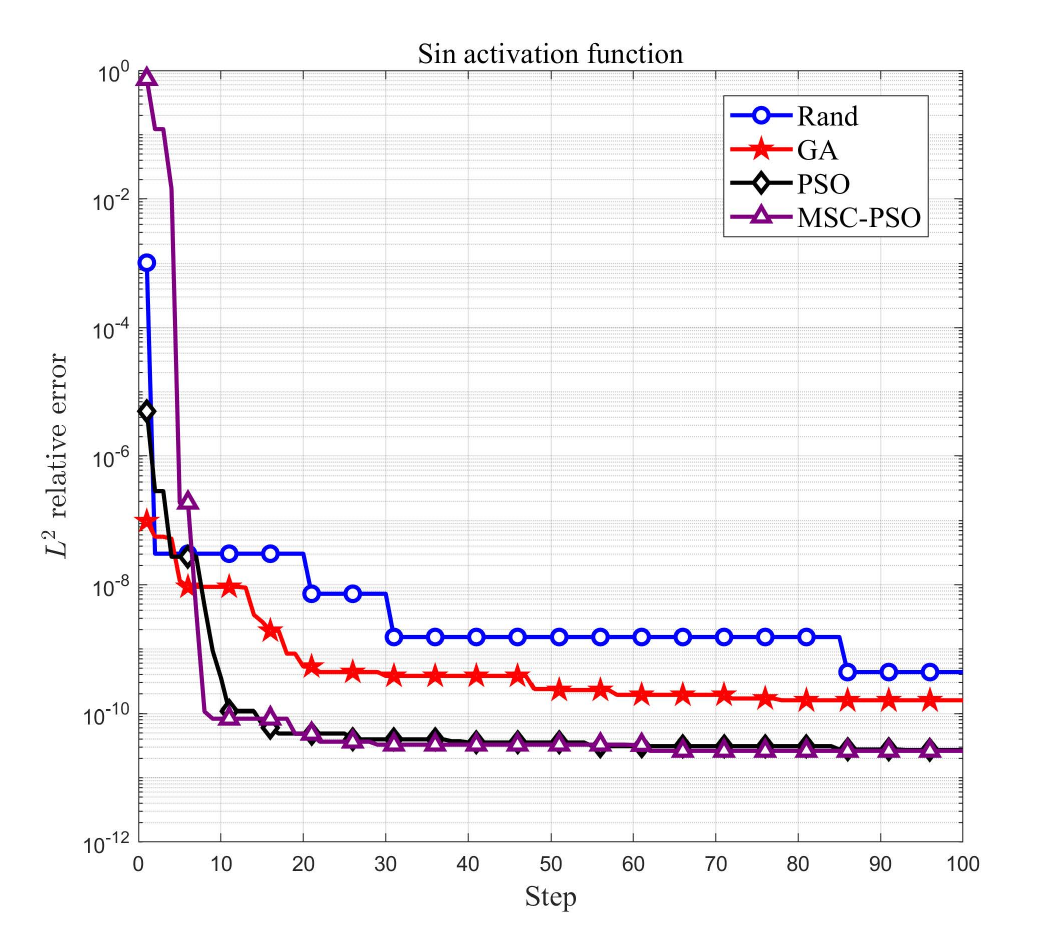}
		(a)
	\end{minipage}
	\hfill
	\begin{minipage}[t]{0.24\textwidth}
		\centering
		\includegraphics[width=\textwidth]{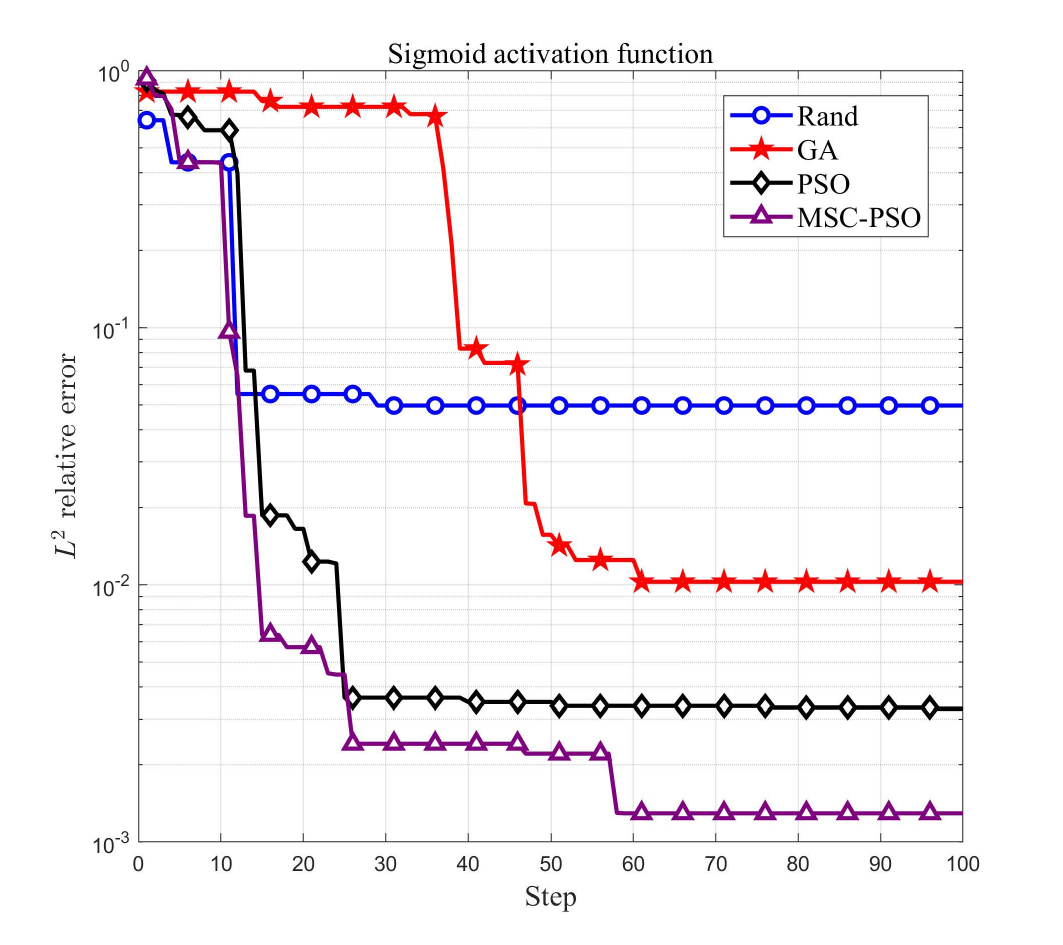}
		(b)
	\end{minipage}
	\hfill
	\begin{minipage}[t]{0.24\textwidth}
		\centering
		\includegraphics[width=\textwidth]{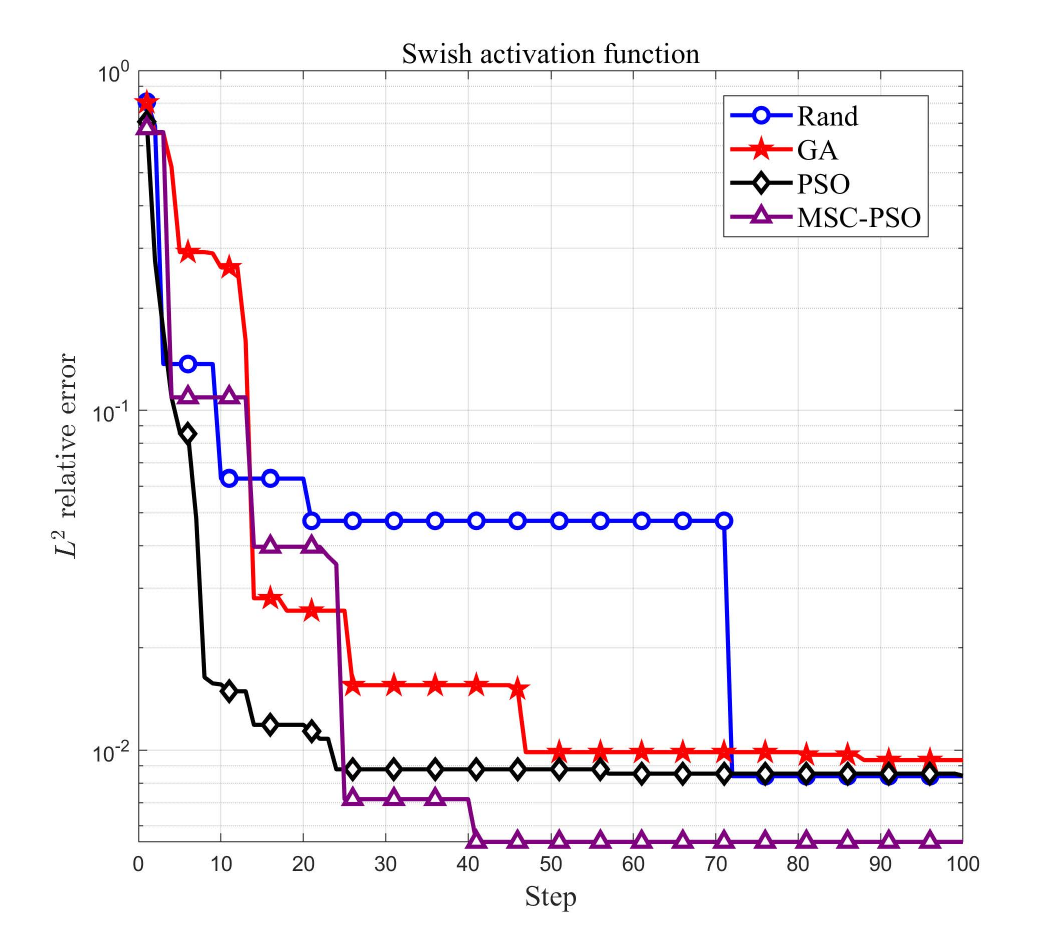}
		(c)
	\end{minipage}
	\hfill
	\begin{minipage}[t]{0.24\textwidth}
		\centering
		\includegraphics[width=\textwidth]{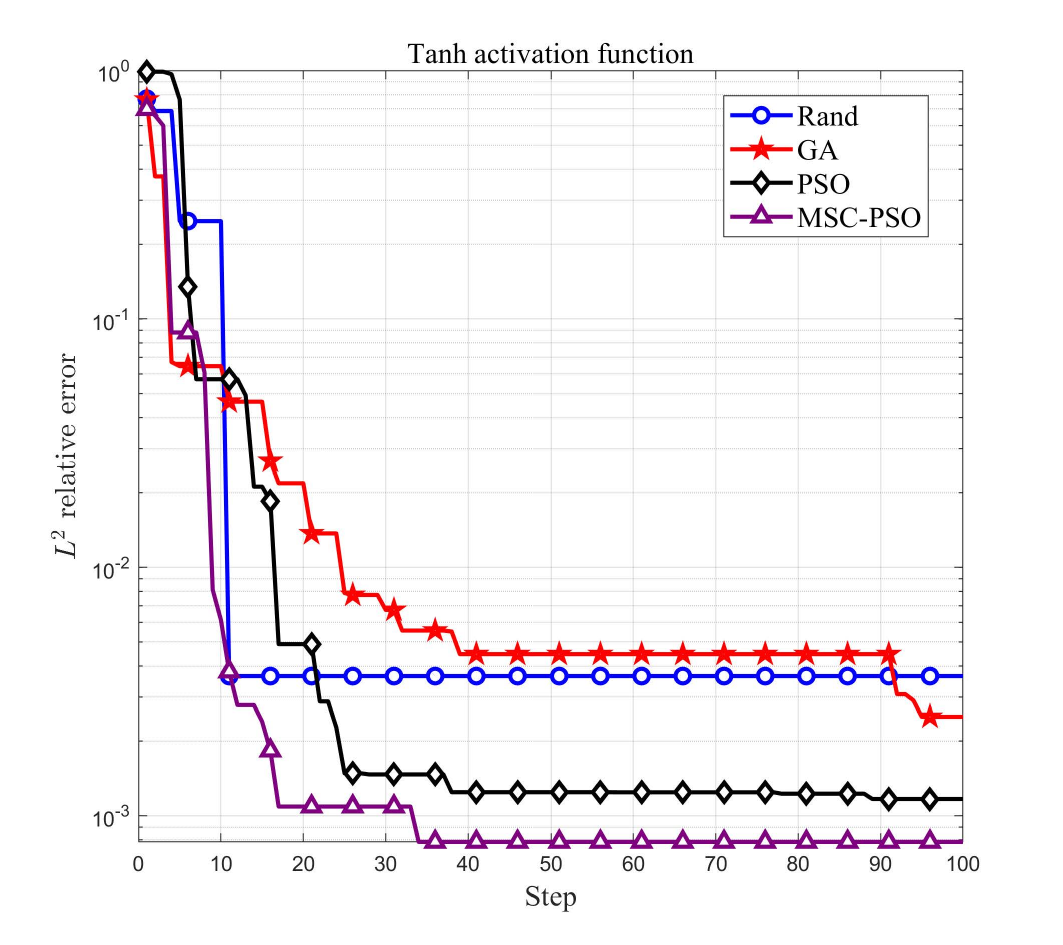}
		(d)
	\end{minipage}
	\caption{The $L^2$ relative error of different activation functions under different intelligent optimization algorithms. (a) Sin. (b) Sigmoid. (c) Swish. (d) Tanh.}
	\label{fig:m1}
\end{figure}

\begin{figure}[!htb]
	\centering
	\includegraphics[width=1\linewidth]{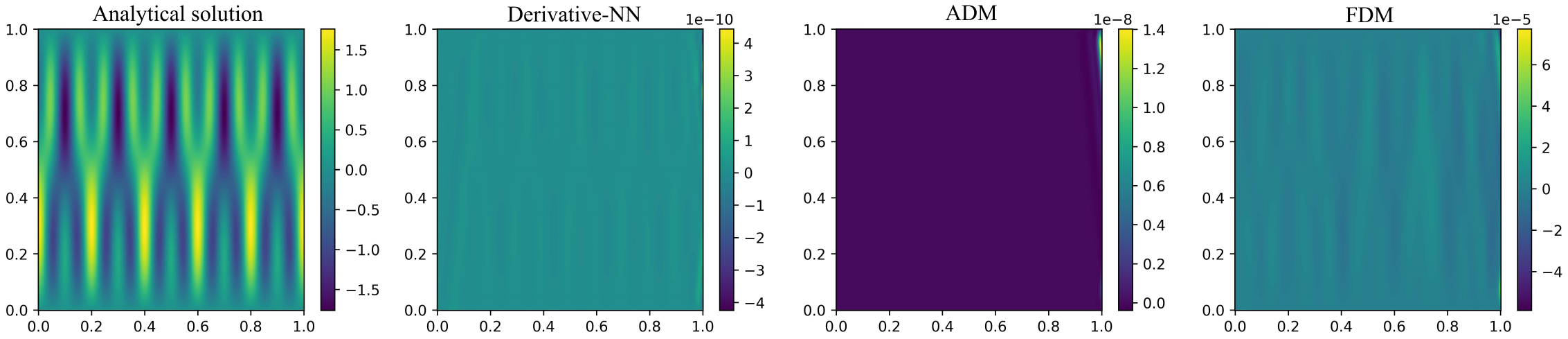}
	\caption{The analytical solution (first column) of wave equation and absolution error (last three columns).}
	\label{fig:m2}
\end{figure}

Fig.\ref{fig:m1} illustrates the convergence characteristics of $L^2$ relative error in wave equation solutions achieved through various activation functions and intelligent optimization algorithms. Notably, the MSC-PSO algorithm exhibits accelerated convergence rates and identifies hyperparameter combinations yielding superior computational precision. Particularly noteworthy is the SO-PIFRNN framework integrated with sin activation function, which demonstrates unprecedented numerical accuracy improvements of seven orders of magnitude compared to conventional activation functions (sigmoid/swish/tanh), attributable to its spectral feature extraction capability in frequency domains. The optimized hyperparameters for SO-PIFRNN are determined as: $N = 2995$, $\omega=70.255$, $\lambda_1 = 9438.998$, $\lambda_2 = 9622.454$, $N_1 = 3000$, $N_2 = 3000$ and $N_3 = 50$. Table \ref{tab:m2} provides systematic comparisons of computational accuracy and efficiency between Derivative-NN, ADM and FDM under optimal hyperparameter conditions. The results reveal nearly identical $L^2$ relative errors between Derivative-NN and ADM. In contrast, FDM demonstrates significantly elevated $L^2$ relative errors due to inherent truncation errors. The Derivative-NN achieves computational efficiency comparable to FDM while maintaining ADM-level accuracy. Fig.\ref{fig:m2} further validates these results through comparative visualizations of analytical solutions versus absolute error distributions. The Derivative-NN solution demonstrates superior fidelity in resolving frequency-domain features, with maximum pointwise errors confined below $4.438\times10^{-10}$, outperforming both ADM $(1.402\times10^{-8})$ and FDM $(7.725\times10^{-5})$.

\begin{table}[!htb]
	\centering
	\caption{Comparison of calculation accuracy and efficiency.}\label{tab:m2}
	\begin{tabular}{cccc}
		\hline
		& Derivative-NN & ADM & FDM \\ \hline
		$L^2$ relative error & $2.624\times10^{-11}$   &$7.142\times10^{-10}$   &$5.568\times10^{-6}$  \\
		Time/s               & 1.649  &122.708  & 2.007  \\ \hline
	\end{tabular}
\end{table}

\subsection{Example 3: Kirchhoff-Love thin plate problems}

In this subsection, based on Kirchhoff-Love thin plate theoretical framework \cite{timoshenko1959theory, reddy2006theory, roy2023deep}, the SO-PIFRNN is used to solve the mechanical response problem of linear elastic plates under transverse distributed loads. A thin plate, as a representative planar structural element, is characterized by in-plane dimensions (length and width) significantly exceeding its thickness (typically with a thickness-to-span ratio below 1/20), thereby allowing reduction to a two-dimensional planar problem. A rectangular plate with dimensions $a \times b$ and thickness $h$ is considered. According to the basic hypothesis of Kirchhoff-Love theory, the static equilibrium of thin plates under transverse loading is governed by a fourth-order partial differential equation:
\begin{equation}
	\label{eq:ho1}
	\mathcal{D}\left( {\frac{{{\partial ^4}u}}{{\partial {x_1}^4}} + 2\frac{{{\partial ^4}u}}{{\partial {x_1}^2\partial {x_2}^2}} + \frac{{{\partial ^4}u}}{{\partial {x_2}^4}}} \right)\left( \bm{x} \right) = q\left( \bm{x} \right),
\end{equation}
with displacement boundary conditions and moment boundary conditions:
\begin{equation}
	\label{eq:ho22}
	\begin{aligned}
		& u = 0, \quad -\mathcal{D} \left( \frac{\partial^2 u}{\partial x_1^2} + \nu \frac{\partial^2 u}{\partial x_2^2} \right) = 0, \quad for \; x_1 = 0 \; and \; x_1 = a, \\
		& u = 0, \quad -\mathcal{D} \left( \frac{\partial^2 u}{\partial x_2^2} + \nu \frac{\partial^2 u}{\partial x_1^2} \right) = 0, \quad for \; x_2 = 0 \; and \; x_2 = b,
	\end{aligned}
\end{equation}
where $u\left( \bm{x} \right)$ represents the transverse deflection, $q\left( \bm{x} \right)$ corresponds to the transverse load distribution, $\mathcal{D}=\frac{Eh^3}{12(1-\mu^2)}$ denotes the flexural rigidity of the plate, $E$, $\nu$ and $h$ respectively signify the Young's modulus, Poisson's ratio and thickness of the plate.

Assume the distributed transverse load applied to the plate is expressed as
\begin{equation*}
	q\left(\bm{x} \right) =q_0 \sin \frac{{\pi {x_1}}}{a}\sin \frac{{\pi {x_2}}}{b}.
\end{equation*}
The analytical solution of plate deflection $u$ is expressed as
\begin{equation*}
	u\left( \bm{x} \right) = \frac{q_0}{{\mathcal{D}{\pi ^4}{{\left( {\frac{1}{{{a^2}}} + \frac{1}{{{b^2}}}} \right)}^2}}}\sin \frac{{\pi {x_1}}}{a}\sin \frac{{\pi {x_2}}}{b}.
\end{equation*}
The solution variables also include bending moment, twisting moment and shearing forces:
\begin{equation}
	\begin{aligned}
		& M_{x_1} = -\mathcal{D} \left( \frac{\partial^2 u}{\partial x_1^2} + \nu \frac{\partial^2 u}{\partial x_2^2} \right)= \frac{q_0}{\pi^2 \left(\frac{1}{a^2} + \frac{1}{b^2}\right)^2} \left(\frac{1}{a^2} + \frac{\nu}{b^2}\right) \sin \frac{\pi x_1}{a} \sin \frac{\pi x_2}{b},\\
		& M_{x_2} = -\mathcal{D} \left( \frac{\partial^2 u}{\partial x_2^2} + \nu \frac{\partial^2 u}{\partial x_1^2} \right)= \frac{q_0}{\pi^2 \left(\frac{1}{a^2} + \frac{1}{b^2}\right)^2} \left(\frac{\nu}{a^2} + \frac{1}{b^2}\right) \sin \frac{\pi x_1}{a} \sin \frac{\pi x_2}{b},\\
		& M_{x_1x_2} = \mathcal{D} (1-\nu) \frac{\partial^2 u}{\partial x_1 \partial x_2}= \frac{q_0 (1-\nu)}{\pi^2 \left(\frac{1}{a^2} + \frac{1}{b^2}\right)^2 ab} \left(\frac{\nu}{a^2} + \frac{1}{b^2}\right) \cos \frac{\pi x_1}{a} \cos \frac{\pi x_2}{b},\\
		& Q_{x_1} = -\mathcal{D} \frac{\partial}{\partial x_1} \left( \frac{\partial^2 u}{\partial x_1^2} + \frac{\partial^2 u}{\partial x_2^2} \right)= \frac{q_0}{\pi a \left(\frac{1}{a^2} + \frac{1}{b^2}\right)} \cos \frac{\pi x_1}{a} \sin \frac{\pi x_2}{b},\\
		& Q_{x_2} = -\mathcal{D} \frac{\partial}{\partial x_2} \left( \frac{\partial^2 u}{\partial x_1^2} + \frac{\partial^2 u}{\partial x_2^2} \right)= \frac{q_0}{\pi b \left(\frac{1}{a^2} + \frac{1}{b^2}\right)} \sin \frac{\pi x_1}{a} \cos \frac{\pi x_2}{b}.
	\end{aligned}
\end{equation}

Given that bending moments, twisting moments and transverse shear forces can all be derived through differentiation of the deflection field, this study only employs the SO-PIFRNN to solve the Eq \eqref{eq:ho1} to calculate the deflection field. Then calculate the bending moments, twisting moments and transverse shear forces through deflection field. Specifically, $N_1$ collocation points are distributed within the computational domain, $N_2$ and $N_3$ points are assigned to displacement boundary conditions and moment boundary conditions, respectively. Penalty weights $\lambda_1$ and $\lambda_2$ are introduced as weighting factors for displacement and moment boundary conditions, respectively. The governing equation \eqref{eq:ho1} and boundary conditions \eqref{eq:ho22} are discretized into a linear system for numerical solution. Based on the existing knowledge, it can be analyzed that the frequency of the displacement solution is relatively low. Therefore, we narrow the range of the hyperparameter $\omega$ of the neural network. Table \ref{tab:ho1} presents the optimization ranges of all hyperparameters.

\begin{table}[!htb]
	\centering
	\caption{The optimization range of hyperparameter.}\label{tab:ho1}
	\begin{tabular}{ccccc}
		\hline
		Hyperparameters	& $N$ & $\omega$ & $\lambda_1$, $\lambda_2$ & $N_1$, $N_2$, $N_3$ \\ \hline
		Range	& $\left[ {10,{\rm{ 2000}}} \right]$ &  $\left[ {0.0001,{\rm{ 10}}}\right]$  &  $\left[ {0.0001,{\rm{ 10000}}} \right]$   &    $\left[  {10,{\rm{ 3000}}}\right]$ \\ \hline
	\end{tabular}
\end{table}

Within the FDM framework, this work employs high-order difference schemes presented in Eqs.\eqref{eq:ho2} and \eqref{eq:ho3} to approximate fourth-order derivatives and fourth-order mixed partial derivatives. Set the step $\Delta h = 3\times10^{-3}$.
\begin{equation}
	\label{eq:ho2}
	\begin{aligned}
		&\frac{\partial^4 u(x_1, x_2)}{\partial x_1^4} = \frac{\mathfrak{u}_1 + \mathfrak{u}_2 + 6u(x_1, x_2)}{\Delta h ^4},\\
		&\mathfrak{u}_1 = u(x_1 + 2\Delta h, x_2) + u(x_1 - 2\Delta h, x_2),\\
		&\mathfrak{u}_2 = -4((x_1 + \Delta h, x_2) + (x_1 - \Delta h, x_2)).
	\end{aligned}
\end{equation}

\begin{equation}
	\label{eq:ho3}
	\begin{aligned}
		&\frac{\partial^4 u(x_1, x_2)}{\partial x_1^2x_2^2} = \frac{\mathfrak{u}_1 + \mathfrak{u}_2 + \mathfrak{u}_3 +\mathfrak{u}_4 + 4u(x_1, x_2)}{\Delta h ^4},\\
		&\mathfrak{u}_1 = u(x_1 + \Delta h, x_2 + \Delta h) + u(x_1 - \Delta h, x_2 - \Delta h),\\
		&\mathfrak{u}_2 = -2(u(x_1 + \Delta h, x_2) + u(x_1 - \Delta h, x_2)),\\
		&\mathfrak{u}_3 = u(x_1 - \Delta h, x_2 + \Delta h) + u(x_1 + \Delta h, x_2 - \Delta h),\\
		&\mathfrak{u}_4 = -2(u(x_1, x_2 + \Delta h) + u(x_1, x_2 - \Delta h)).
	\end{aligned}
\end{equation}

\begin{figure}[!htb]
	\centering
	\begin{minipage}[t]{0.24\textwidth}
		\centering
		\includegraphics[width=\textwidth]{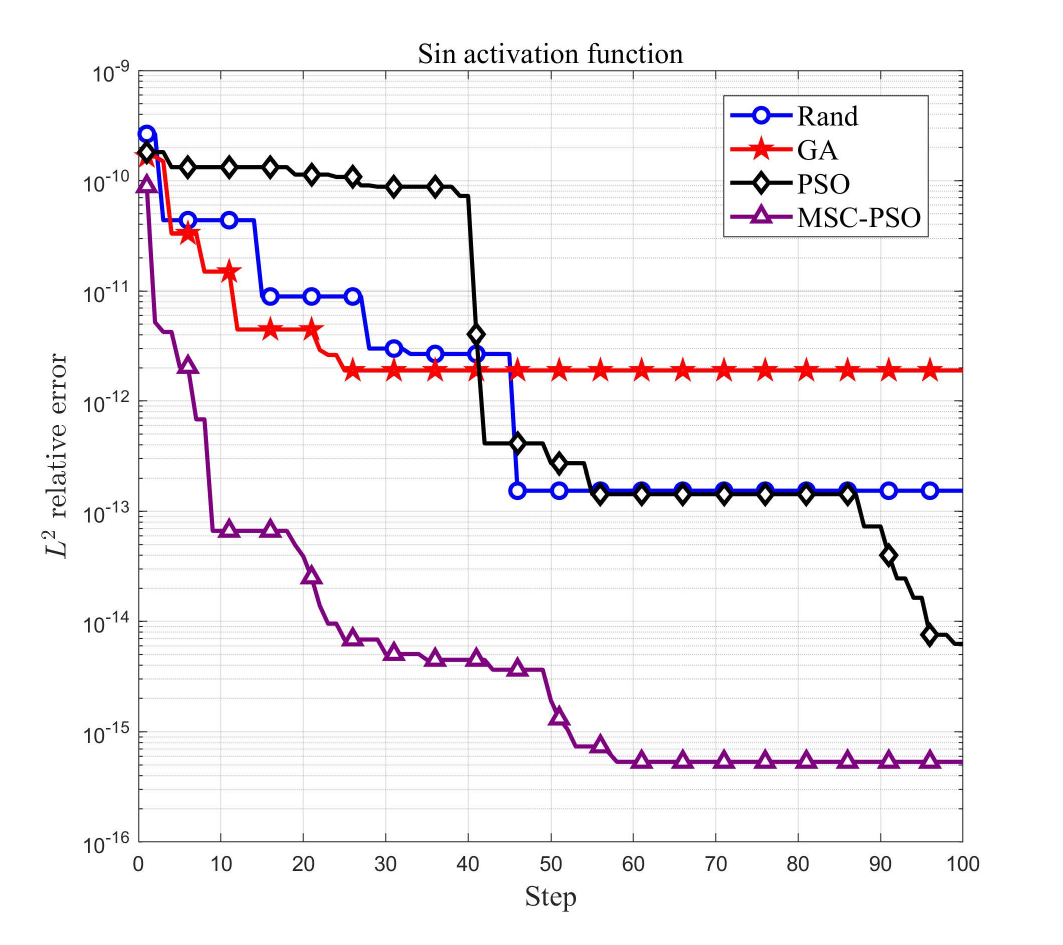}
		(a)
	\end{minipage}
	\hfill
	\begin{minipage}[t]{0.24\textwidth}
		\centering
		\includegraphics[width=\textwidth]{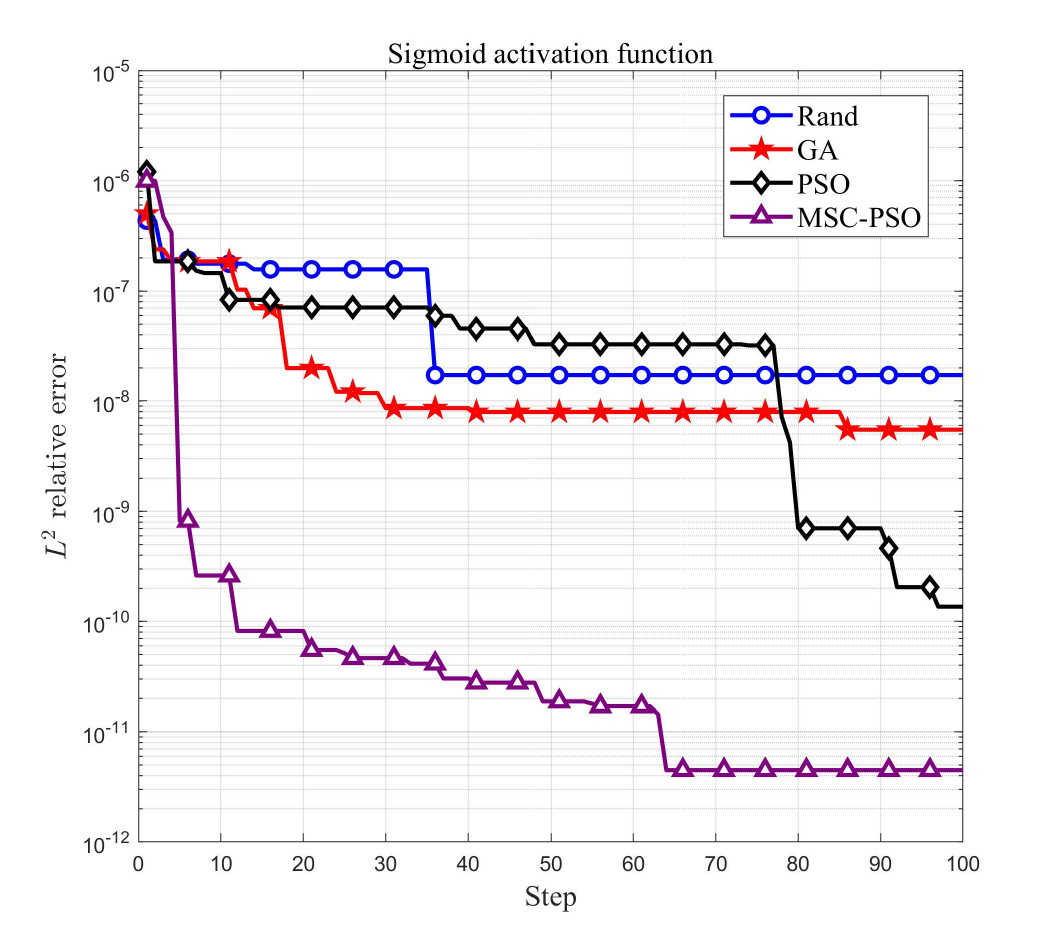}
		(b)
	\end{minipage}
	\hfill
	\begin{minipage}[t]{0.24\textwidth}
		\centering
		\includegraphics[width=\textwidth]{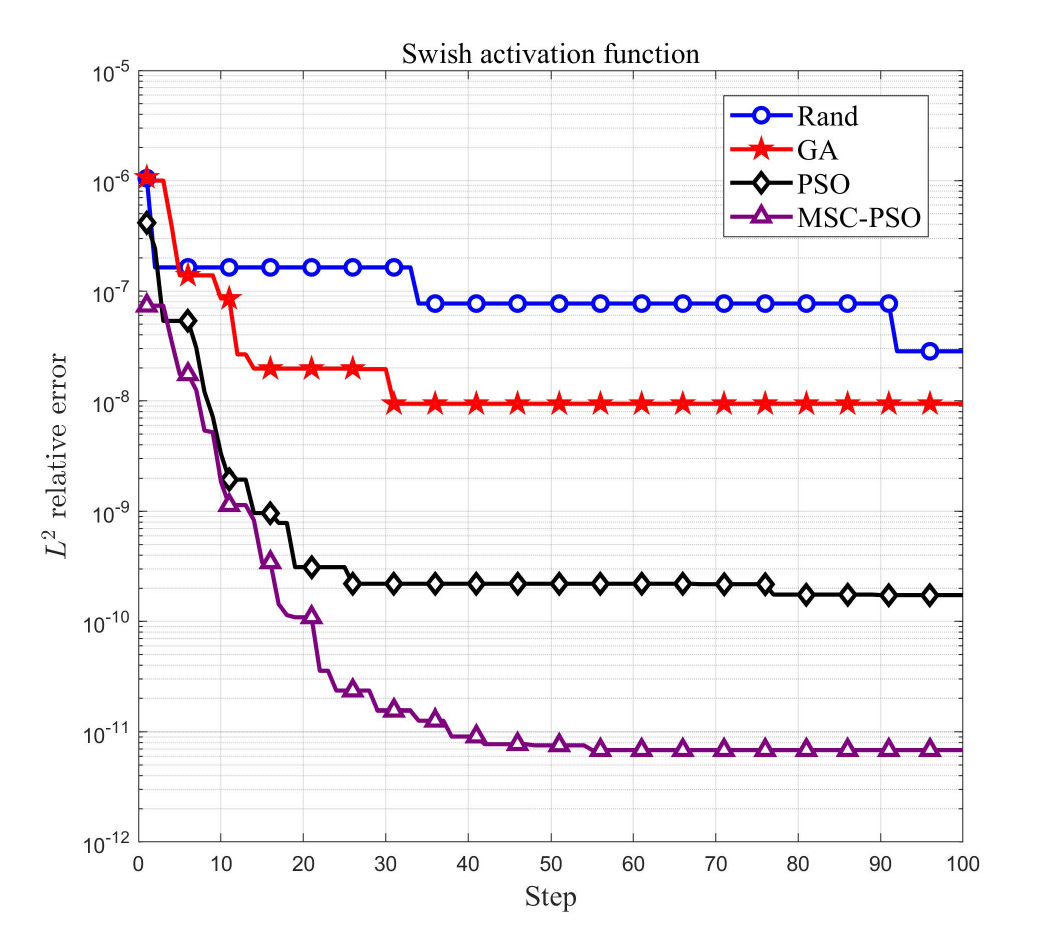}
		(c)
	\end{minipage}
	\hfill
	\begin{minipage}[t]{0.24\textwidth}
		\centering
		\includegraphics[width=\textwidth]{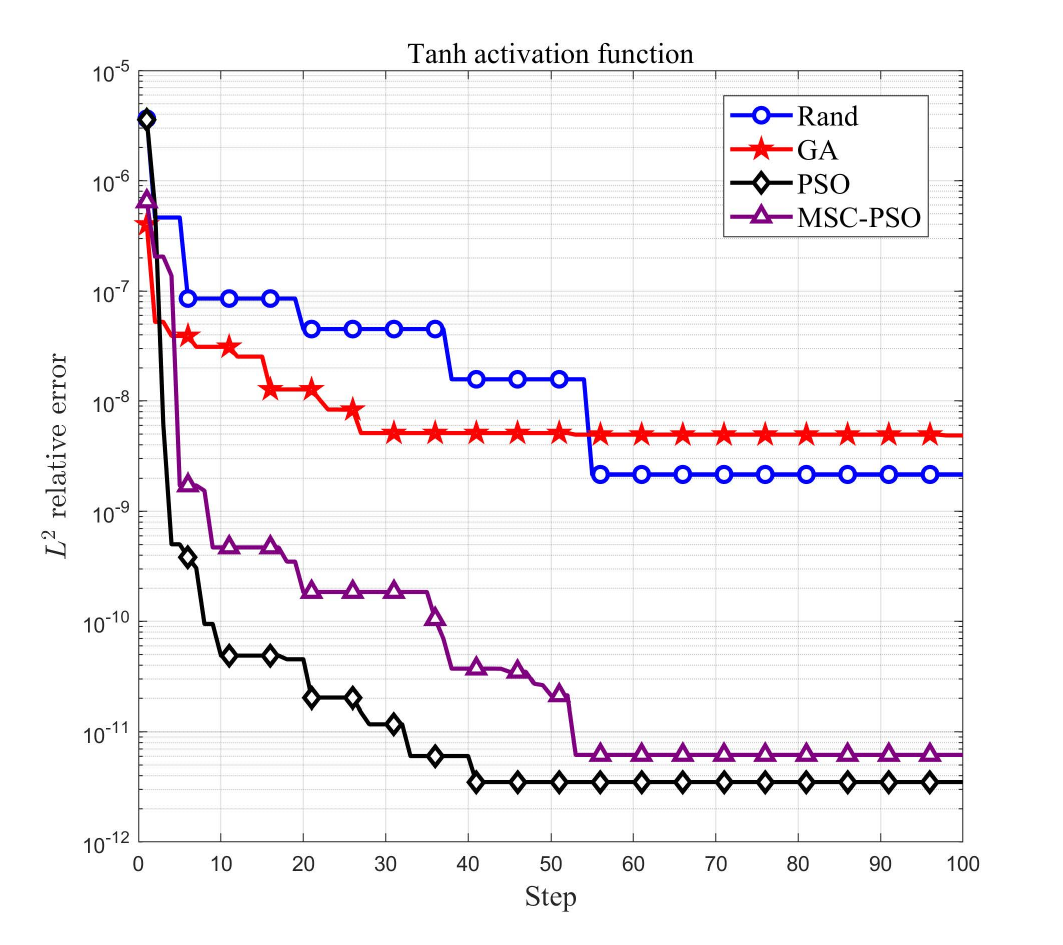}
		(d)
	\end{minipage}
	\caption{The $L^2$ relative error of different activation functions under different intelligent optimization algorithms. (a) Sin. (b) Sigmoid. (c) Swish. (d) Tanh.}
	\label{fig:ho1}
\end{figure}

\begin{figure}[!htb]
	\centering
	\includegraphics[width=1\linewidth]{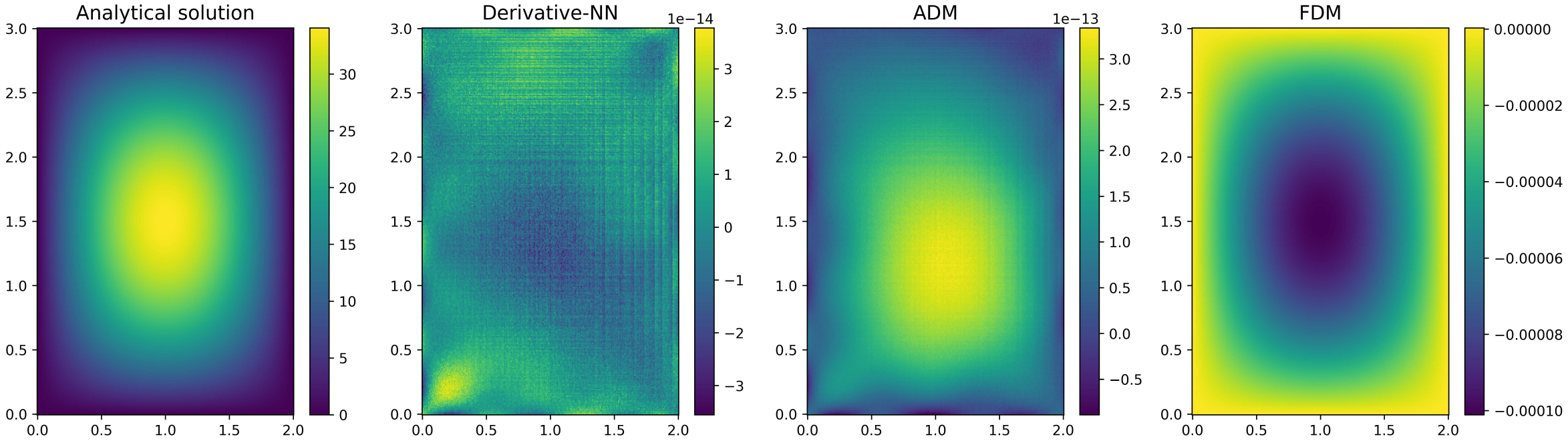}
	\caption{The analytical solution (first column) of plate deflection of Kirchhoff-Love thin plate problems and absolution error (last three columns).}
	\label{fig:ho2}
\end{figure}

In this experiment, set $a=2$, $b=3$, $h=0.01$, $E=2.6\times10^{2}$, $\mu=0.25$ and $q_0=1\times10^{-2}$. Fig.\ref{fig:ho1} depicts the convergence profiles of the $L^2$ relative error for solutions generated by distinct activation functions and evolutionary computation techniques. Notably, the MSC-PSO algorithm exhibits enhanced convergence kinetics and identifies hyperparameter configurations that optimize numerical precision. The accuracy of sin activation function used in SO-PIFRNN framework is much higher than that of other activation functions. The optimized hyperparameters for SO-PIFRNN are determined as: $N=707$, $\omega=2.607$, $\lambda_1=0.002$, $\lambda_2=8.441$, $N_1=787$, $N_2=1508$ and $N_3=2951$. Table \ref{tab:ho2} provides a systematic performance comparison among Derivative-NN, ADM and FDM under optimized hyperparameters. The accuracy and efficiency of Derivative-NN are better than those of ADM and FDM. Especially, due to the influence of truncation error, the accuracy of FDM is far inferior to that of Derivative-NN and ADM. Fig.\ref{fig:ho2} further visualizes the analytical solution and the absolute error distribution of the three methods.

\begin{table}[!htb]
	\centering
	\caption{Comparison of calculation accuracy and efficiency.}\label{tab:ho2}
	\begin{tabular}{cccc}
		\hline
		& Derivative-NN & ADM & FDM \\ \hline
		$L^2$ relative error & $5.342\times10^{-16}$   &$9.856\times10^{-15}$   &$2.970\times10^{-6}$  \\
		Time/s               & 0.139   & 9.422  &  0.267 \\ \hline
	\end{tabular}
\end{table}

\begin{figure}[!htb]
	\centering
	\begin{minipage}[t]{0.8\textwidth}
		\centering
		\includegraphics[width=\textwidth]{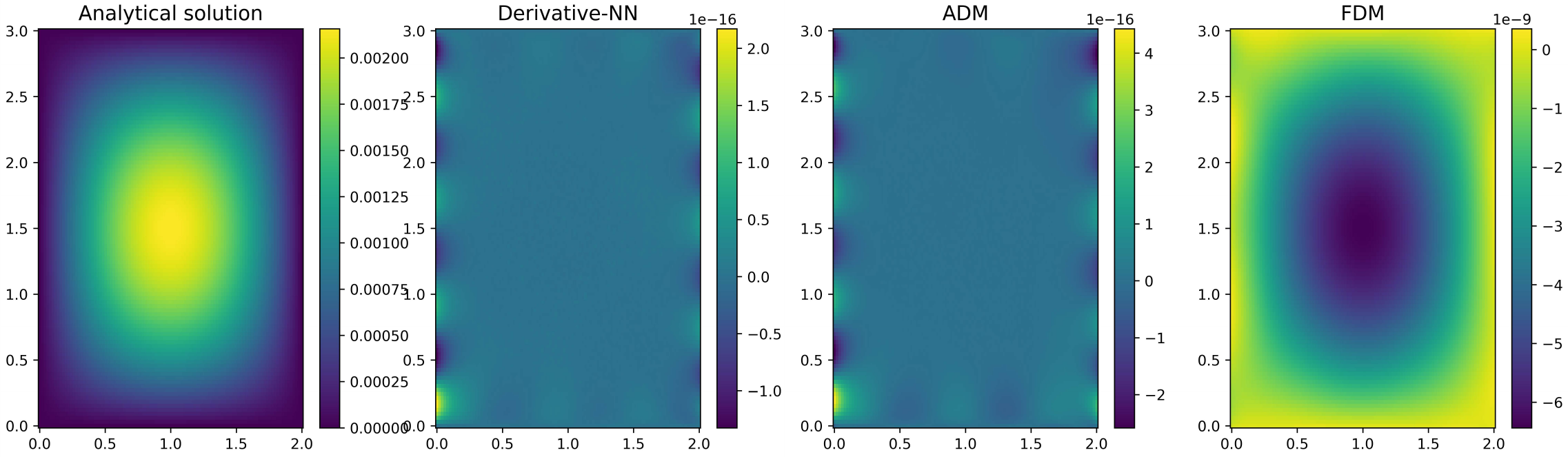}
		(a)
	\end{minipage}
	\hfill
	\begin{minipage}[t]{0.8\textwidth}
		\centering
		\includegraphics[width=\textwidth]{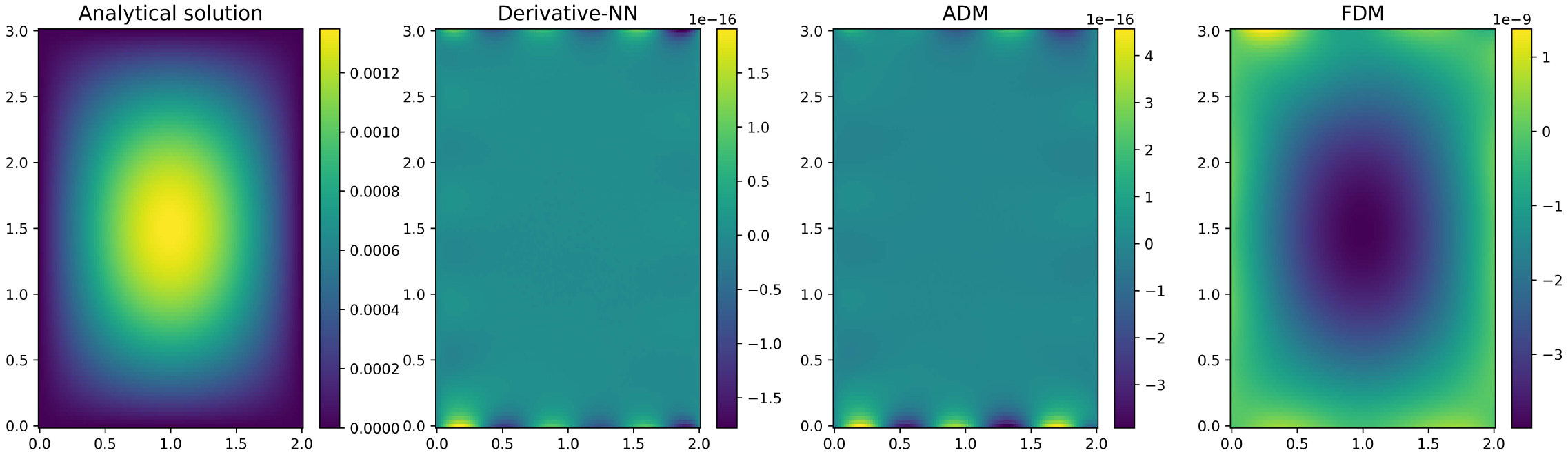}
		(b)
	\end{minipage}
	\hfill
	\begin{minipage}[t]{0.8\textwidth}
		\centering
		\includegraphics[width=\textwidth]{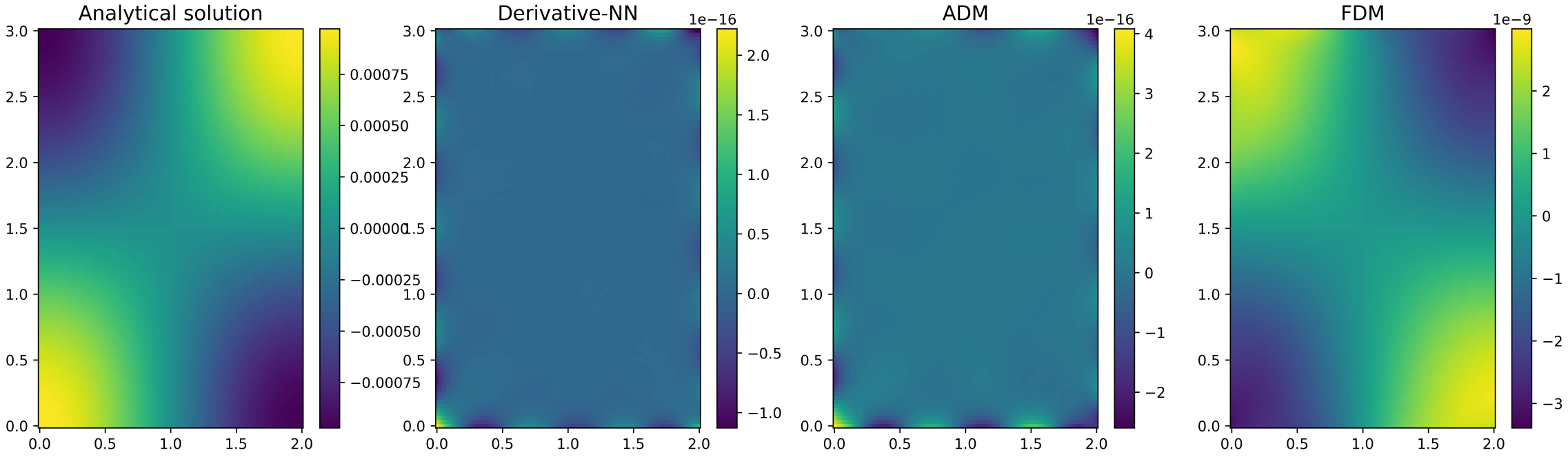}
		(c)
	\end{minipage}
	\hfill
	\begin{minipage}[t]{0.8\textwidth}
		\centering
		\includegraphics[width=\textwidth]{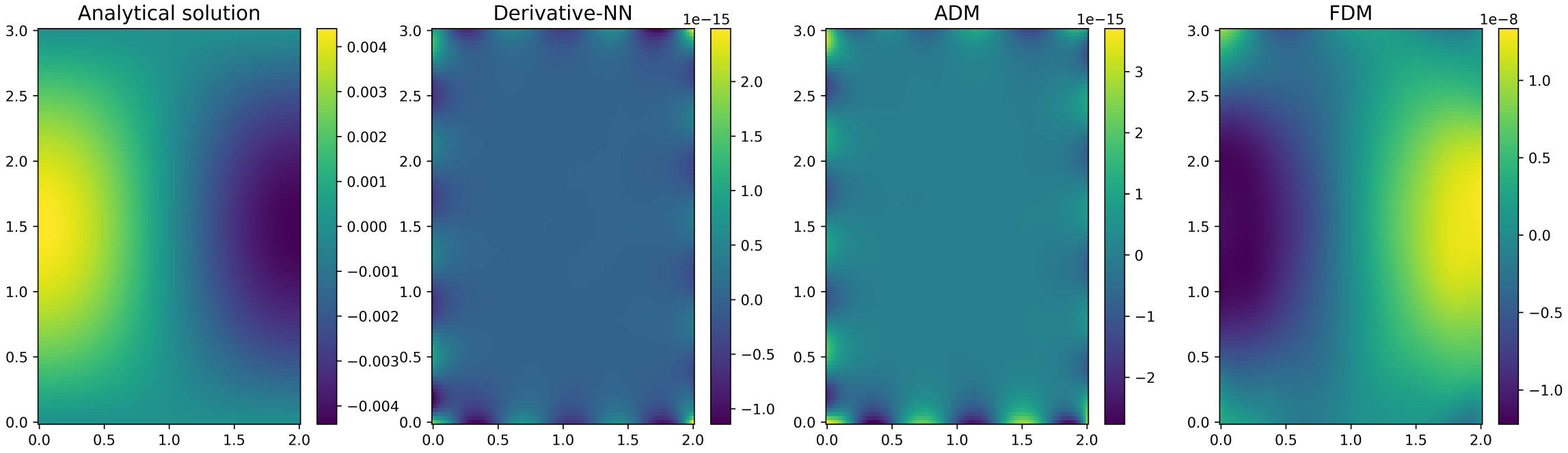}
		(d)
	\end{minipage}
	\hfill
	\begin{minipage}[t]{0.8\textwidth}
		\centering
		\includegraphics[width=\textwidth]{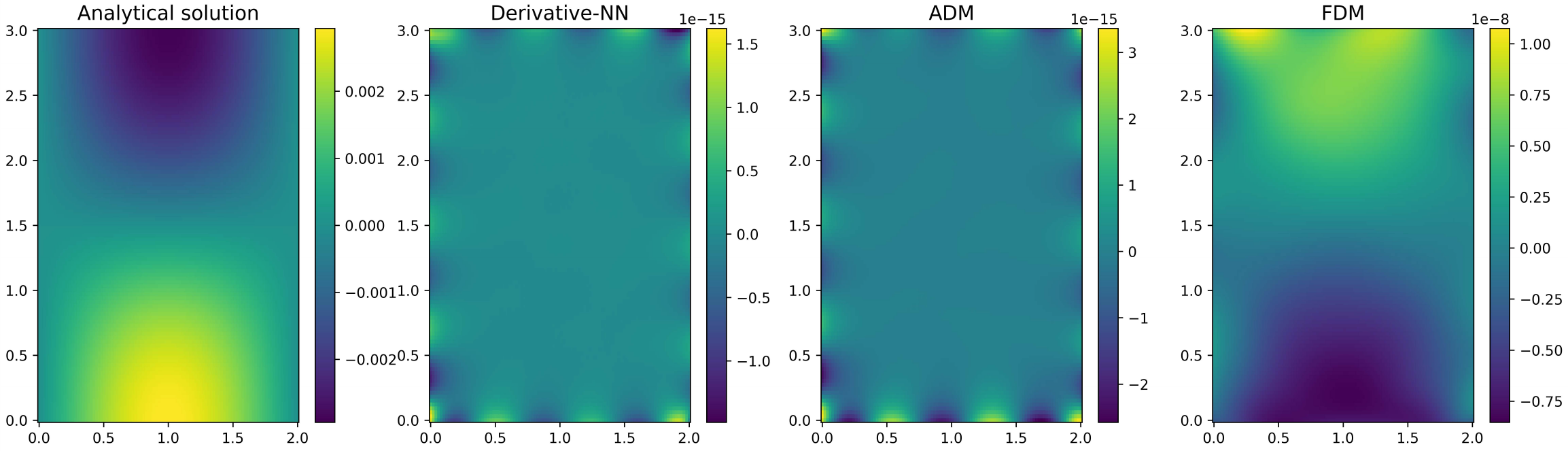}
		(e)
	\end{minipage}
	\caption{The analytical solution (first column) of Kirchhoff-Love thin plate problems and absolution error (last three columns). (a) $M_{x_1}$. (b) $M_{x_2}$. (c) $M_{x_1x_2}$. (d) $Q_{x_1}$. (e) $Q_{x_2}$.}
	\label{fig:ho3}
\end{figure}

Subsequently, the bending moments, twisting moments and shear forces are computed by differentiating the obtained deflection field. Derivative-NN, ADM and FDM are used to differentiate the solution of deflection. Fig.\ref{fig:ho3} illustrates the numerical results for the bending moments, twisting moments and shear forces. Table \ref{tab:ho3} summarizes the relative errors of the solutions. The results demonstrate that the Derivative-NN method achieves the higher accuracy in deriving these mechanical quantities. The Derivative-NN not only efficiently and accurately computes solutions, but also effectively captures the derivative properties of solutions.

\begin{table}[!htb]
	\centering
	\caption{The bending moment, twisting moment and shearing forces of calculation accuracy.}\label{tab:ho3}
	\begin{tabular}{cccc}
		\hline
		 & Derivative-NN & ADM & FDM \\ \hline
		$M_{x_1}$  &$1.510\times10^{-14}$   &$3.497\times10^{-14}$   &$2.975\times10^{-6}$     \\
		$M_{x_2}$   & $2.445\times10^{-14}$   &$6.810\times10^{-14}$   &$2.980\times10^{-6}$   \\
		$M_{x_1x_2}$   & $2.265\times10^{-14}$   &$5.820\times10^{-14}$   &$2.973\times10^{-6}$      \\
		$Q_{x_1}$     & $8.049\times10^{-14}$   &$1.787\times10^{-13}$   &$2.983\times10^{-6}$   \\
		$Q_{x_2}$   & $1.149\times10^{-13}$   &$2.682\times10^{-13}$   &$3.015\times10^{-6}$    \\ \hline
	\end{tabular}
\end{table}

\subsection{Example 4: high-dimensional problems}

We delve into the high-dimensional Poisson equation on the domain $\Omega = [-1, 1]^d$,
\begin{equation}
	\label{eq:hd1}
	\begin{aligned}
		- \Delta u\left( \bm{x} \right) & = f\left( \bm{x} \right), \quad \bm{x} \in \Omega,\\
		u\left( \bm{x} \right) & = g\left( \bm{x} \right), \quad \bm{x} \in \partial \Omega,
	\end{aligned}
\end{equation}
where $\Delta = \sum_{i=1}^{d} \frac{\partial^2}{\partial x_i^2}$, the source term $f(\bm{x}) = \frac{1}{d} \left( \sin\left(\frac{1}{d}\sum_{i=1}^{d} x_i\right) - 2 \right)$, the boundary term $h(\bm{x}) = \Big( \frac{1}{d} \sum_{i=1}^{d} x_i \Big)^2 + \sin\Big( \frac{1}{d} \sum_{i=1}^{d} x_i \Big)$. The analytical solution can be expressed as:
\begin{equation}
	u(\bm{x}) = \Big(\frac{1}{d} \sum_{i=1}^{d} x_i \Big)^2 + \sin\Big(\frac{1}{d} \sum_{i=1}^{d} x_i\Big).
\end{equation}

\begin{figure}[!htb]
	\centering
	\begin{minipage}[t]{0.24\textwidth}
		\centering
		\includegraphics[width=\textwidth]{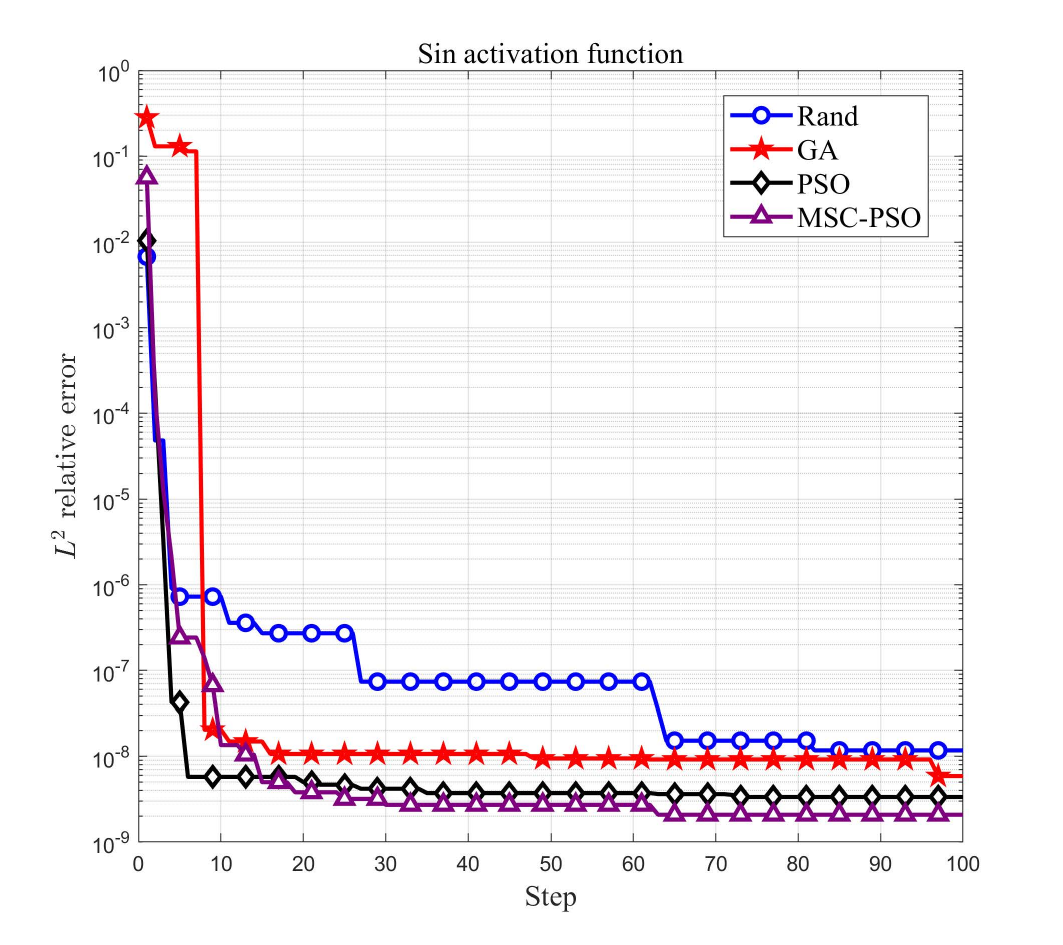}
		(a)
	\end{minipage}
	\hfill
	\begin{minipage}[t]{0.24\textwidth}
		\centering
		\includegraphics[width=\textwidth]{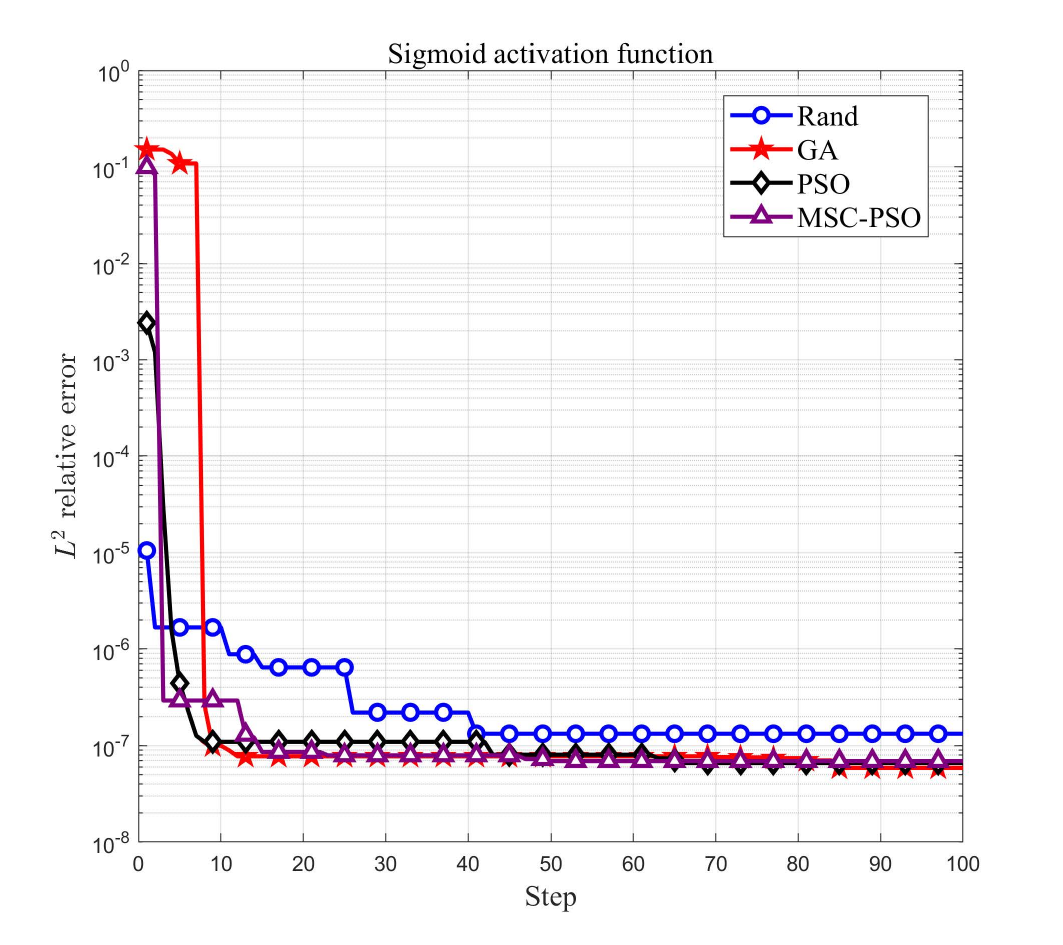}
		(b)
	\end{minipage}
	\hfill
	\begin{minipage}[t]{0.24\textwidth}
		\centering
		\includegraphics[width=\textwidth]{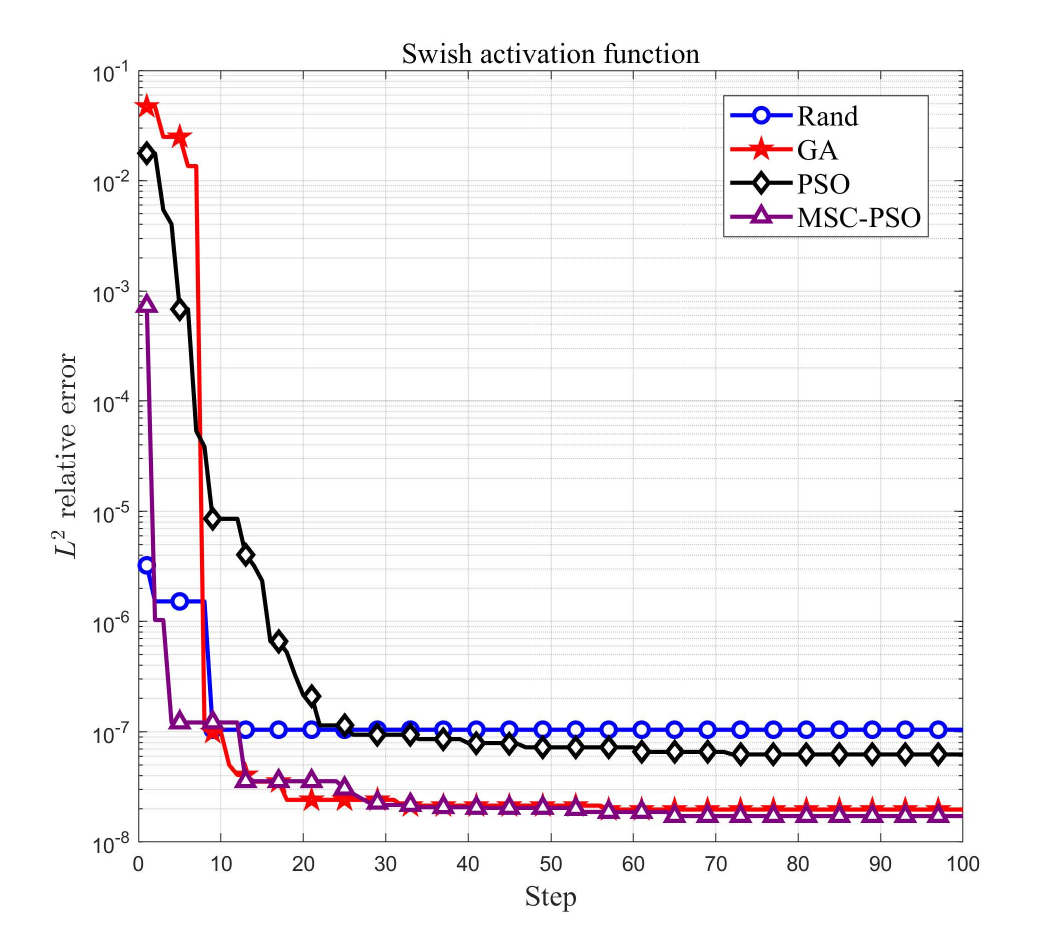}
		(c)
	\end{minipage}
	\hfill
	\begin{minipage}[t]{0.24\textwidth}
		\centering
		\includegraphics[width=\textwidth]{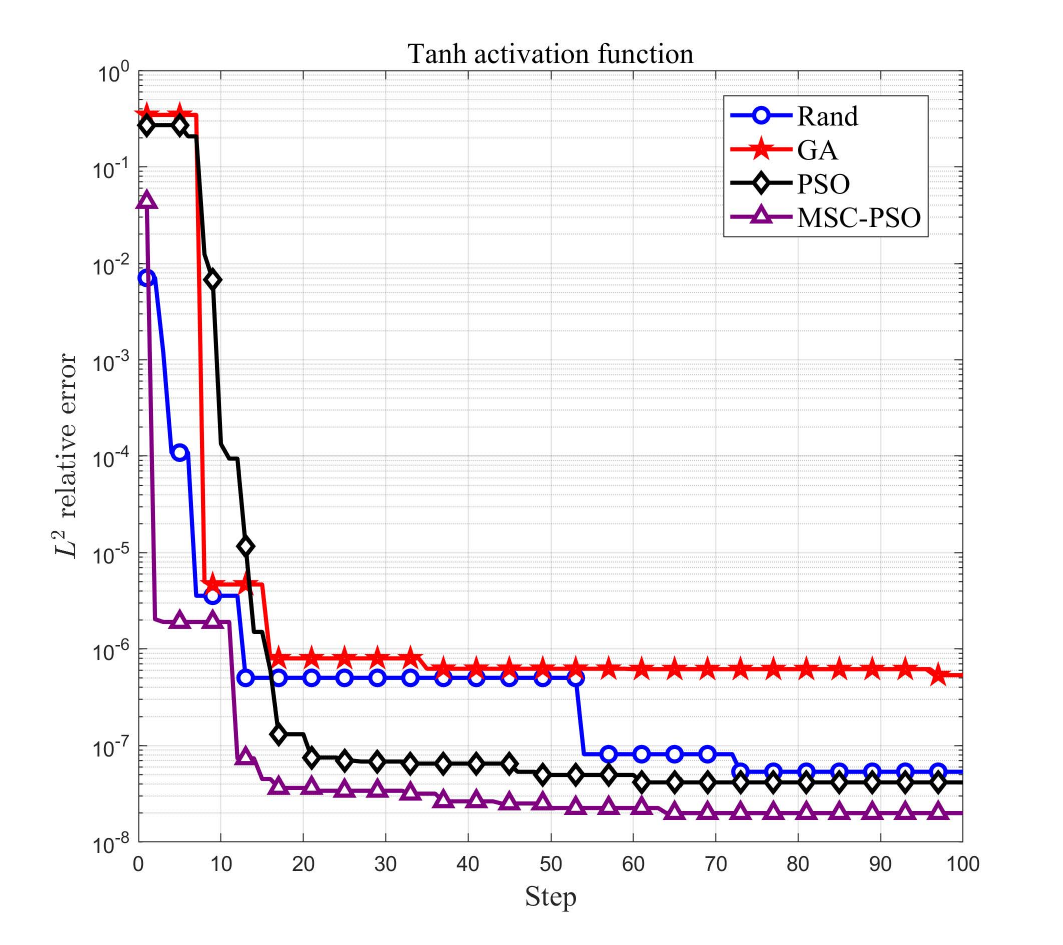}
		(d)
	\end{minipage}
	\caption{The $L^2$ relative error in dimension $d = 5$ of different activation functions under different intelligent optimization algorithms. (a) Sin. (b) Sigmoid. (c) Swish. (d) Tanh.}
	\label{fig:hd1}
\end{figure}

\begin{figure}[!htb]
	\centering
	\begin{minipage}[t]{0.24\textwidth}
		\centering
		\includegraphics[width=\textwidth]{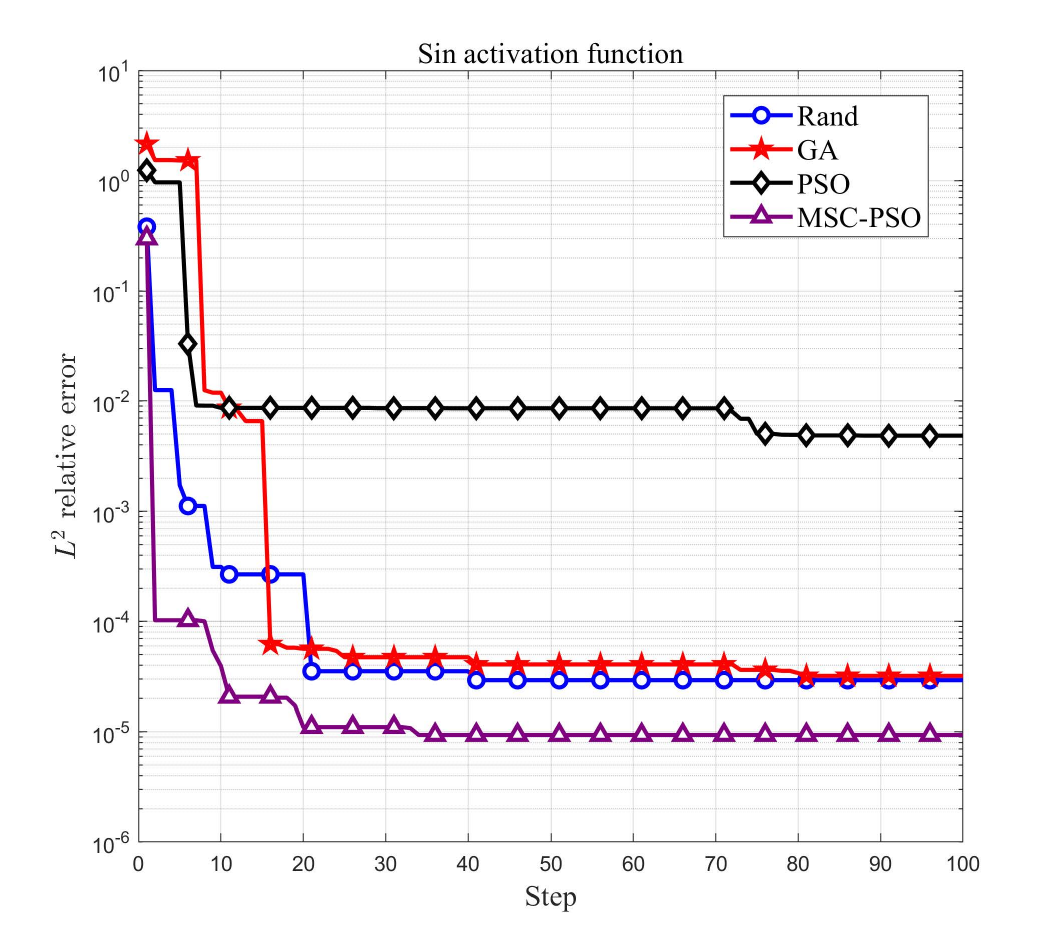}
		(a)
	\end{minipage}
	\hfill
	\begin{minipage}[t]{0.24\textwidth}
		\centering
		\includegraphics[width=\textwidth]{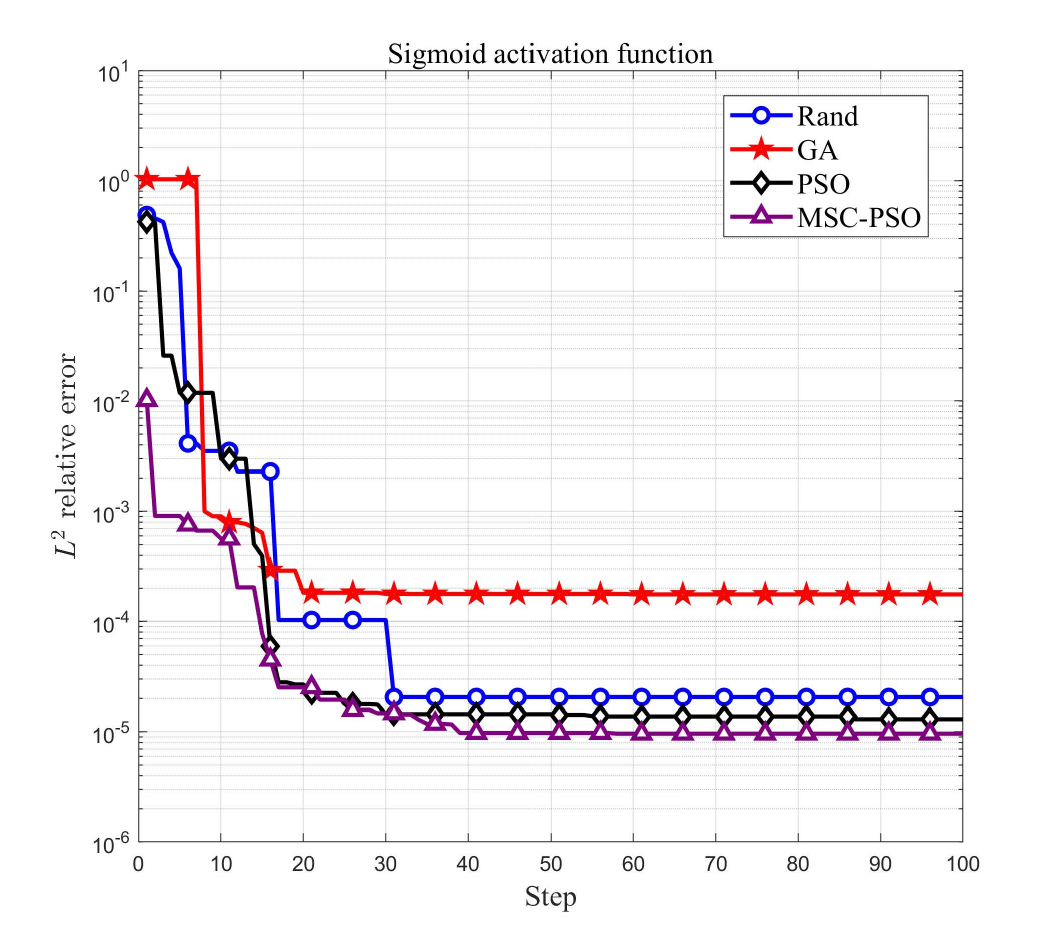}
		(b)
	\end{minipage}
	\hfill
	\begin{minipage}[t]{0.24\textwidth}
		\centering
		\includegraphics[width=\textwidth]{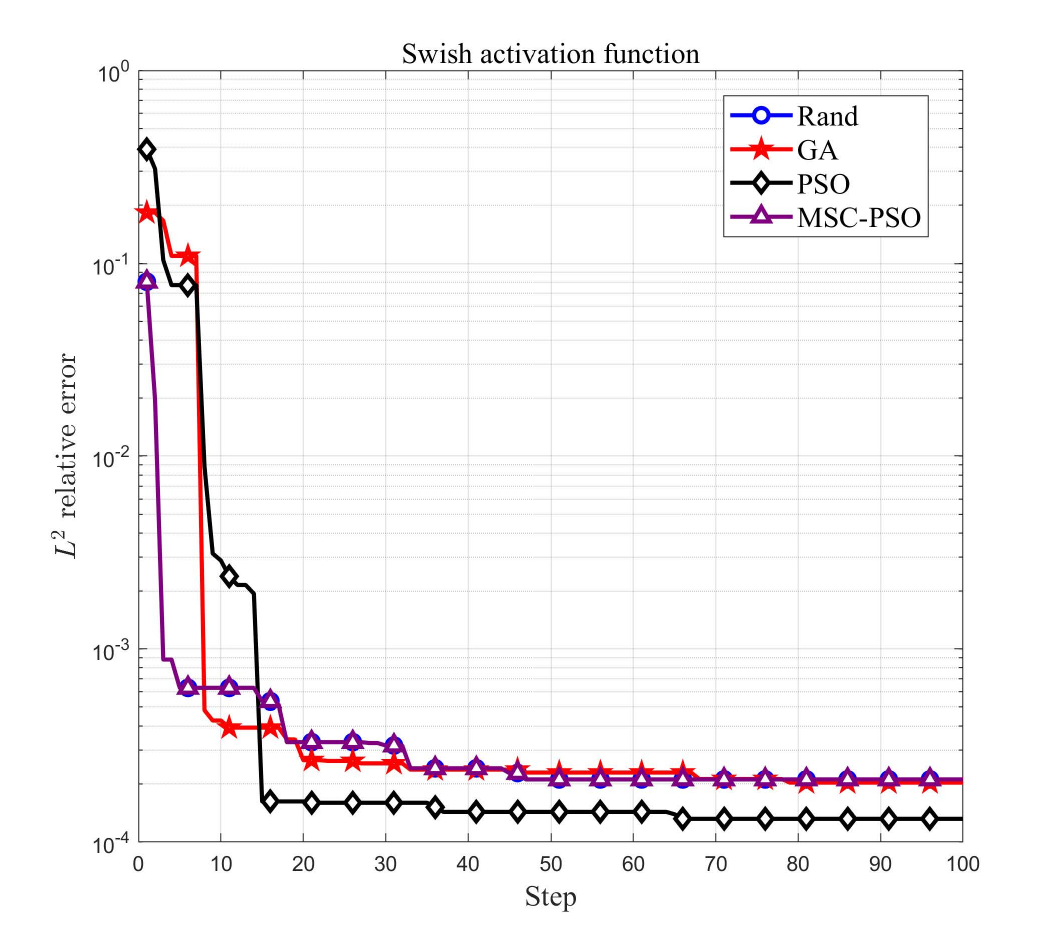}
		(c)
	\end{minipage}
	\hfill
	\begin{minipage}[t]{0.24\textwidth}
		\centering
		\includegraphics[width=\textwidth]{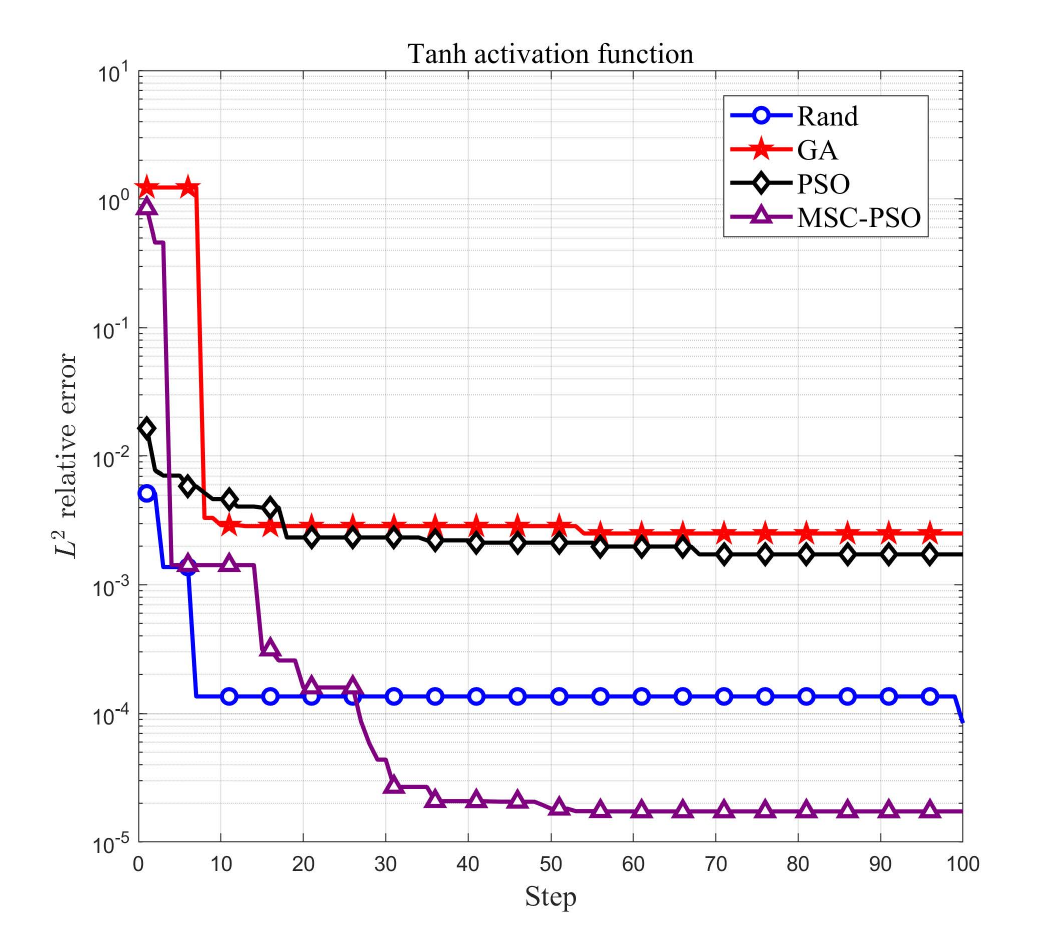}
		(d)
	\end{minipage}
	\caption{The $L^2$ relative error in dimension $d = 10$ of different activation functions under different intelligent optimization algorithms. (a) Sin. (b) Sigmoid. (c) Swish. (d) Tanh.}
	\label{fig:hd2}
\end{figure}

For the high-dimensional Poisson equation \eqref{eq:hd1}, the hyperparameter configuration of this linear system is the same as described in subsection \ref{sec:31}, the equation can be discretized into a linear system \eqref{eq:3}. To validate the ability of SO-PIFRNN to solve high-dimensional problems, this section constructs experimental groups with dimensional parameters $d=5$ and $d=10$. The corresponding hyperparameter search space is shown in Table \ref{tab:hd1}. Figs.\ref{fig:hd1} and \ref{fig:hd2} show the decrease curves of $L^2$ relative error for $d=5$ and $d=10$, respectively. As can be seen from the figures, the hyperparameter set found by MCS-PSO method is optimal in most cases. Notably, the adoption of the sin activation function demonstrates sustained superior approximation capabilities in high-dimensional spaces. The optimal hyperparameters of the SO-PIFRNN method are: $N = 1983$, $\omega=0.219$, $\lambda_1 = 4895.903$, $N_1 = 8510$ and $N_2 = 13196$ when $d=5$. The optimal hyperparameters of the SO-PIFRNN method are: $N = 4793$, $\omega=0.091$, $\lambda_1 = 165.799$, $N_1 = 19889$ and $N_2 = 14342$ when $d=10$. In the case of $d=10$, the optimal values of sin and sigmoid activation functions optimized by MSC-PSO are $9.376\times10^{-6}$ and $9.623\times10^{-6}$, respectively. It can be seen that for high-dimensional problems with relatively smooth solutions, the advantage of the sin activation function is not significant. Table \ref{tab:hd2} systematically compares the comprehensive performance of Derivative-NN, ADM and FDM in solving high-dimensional PDEs. The results reveal that Derivative-NN demonstrates significant advantages in both computational efficiency and solution accuracy for high-dimensional problems. Of particular note is that as the problem dimension increases from $d=5$ to $d=10$, the computation time required by ADM increases dramatically.

\begin{table}[!htb]
	\centering
	\caption{The optimization range of hyperparameter.}\label{tab:hd1}
	\begin{tabular}{ccccc}
		\hline
		Hyperparameters	& $N$ & $\omega$ & $\lambda_1$ & $N_1$, $N_2$ \\ \hline
		Range of $d=5$	& $\left[ {10,{\rm{ 2000}}} \right]$ &  $\left[ {0.0001,{\rm{ 10}}}\right]$  &  $\left[ {0.0001,{\rm{ 10000}}} \right]$   &    $\left[  {10,{\rm{ 20000}}}\right]$ \\
		Range of $d=10$	& $\left[ {10,{\rm{ 5000}}} \right]$ &  $\left[ {0.0001,{\rm{ 10}}}\right]$  &  $\left[ {0.0001,{\rm{ 10000}}} \right]$   &    $\left[  {10,{\rm{ 20000}}}\right]$ \\ \hline
	\end{tabular}
\end{table}

\begin{table}[!htb]
	\centering
	\caption{Comparison of calculation accuracy and efficiency.}\label{tab:hd2}
	\begin{tabular}{ccccc}
		\hline
		&                      & Derivative-NN & ADM & FDM \\ \hline
		\multirow{2}{*}{$d=5$}  & $L^2$ relative error &   $2.076\times10^{-9}$   & $2.067\times10^{-9}$    &  $2.103\times10^{-7}$   \\
		& time                 &  2.017    &  224.375 &   3.212   \\
		\multirow{2}{*}{$d=10$} & $L^2$ relative error &   $9.375\times10^{-6}$   & $9.375\times10^{-6}$    &  $2.055\times10^{-5}$    \\
		& time                 & 34.434   & 3713.210   & 48.730\\ \hline
	\end{tabular}
\end{table}

\subsection{Example 5: Lam\'e equations}

In this subsection, we investigate a plane strain problem in elastic mechanics involving a thick-walled cylindrical structure subjected to combined mechanical loading. As schematically illustrated in Fig.\ref{fig:L1}, the system configuration consists of an annular domain with inner radius $a$ and outer radius $b$, where the outer boundary $\Gamma_1$ experiences superimposed uniform pressure $q_1$ and tangential shear traction $q_2$, while the inner boundary $\Gamma_2$ remains fully constrained. The governing equilibrium equations are formulated in displacement formalism through Lam\'e elasticity theory, constituting a system of coupled partial differential equations:

\begin{figure}[!htb]
	\centering
	\includegraphics[width=0.4\linewidth]{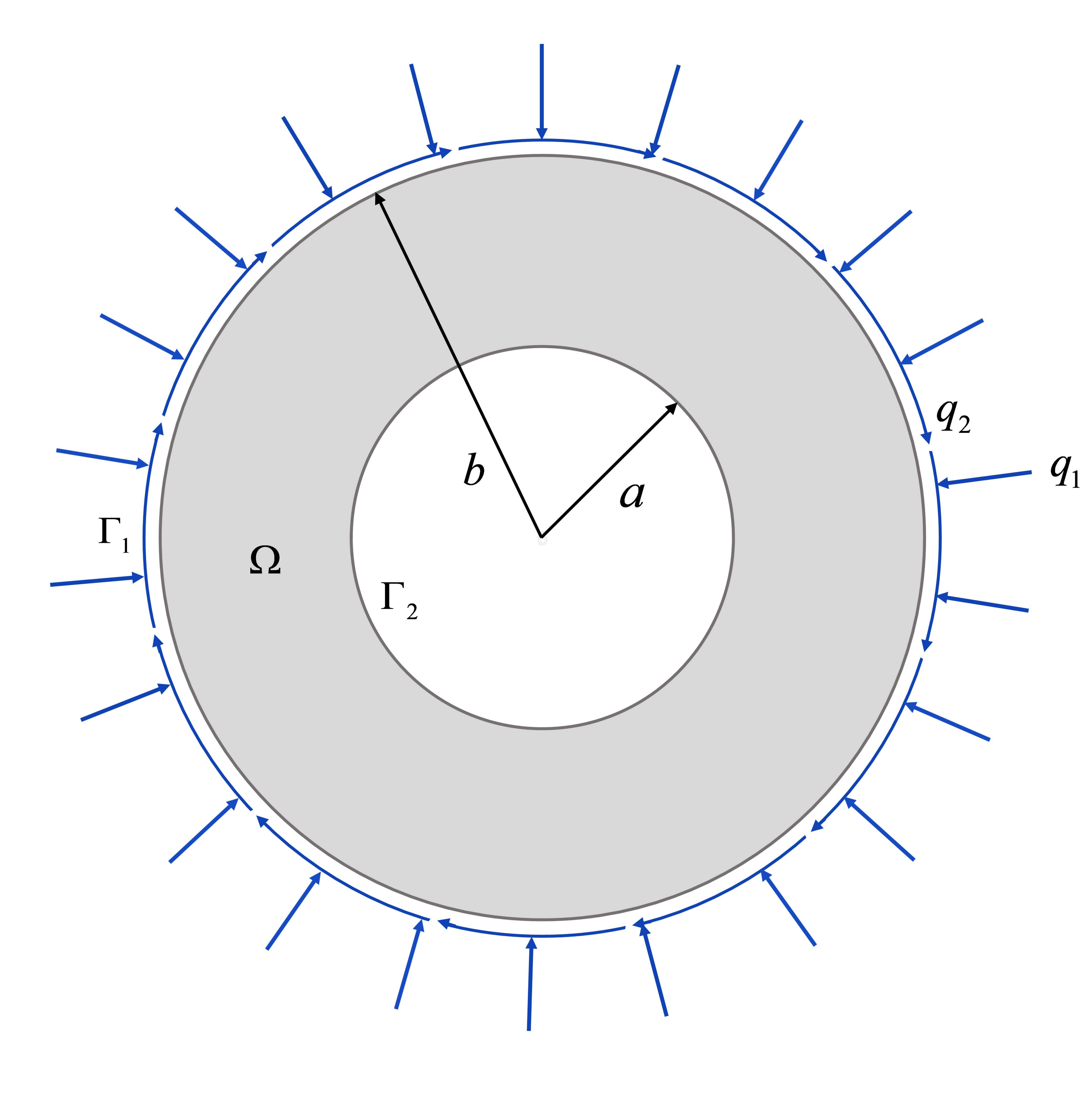}
	\caption{The thick-walled cylinderical structure.}
	\label{fig:L1}
\end{figure}

\begin{equation}
	\label{eq:L1}
	\begin{aligned}
		& \frac{E}{1-\mu^2} \left( \frac{\partial^2 u}{\partial x_1^2} + \frac{1-\mu}{2} \frac{\partial^2 u}{\partial x_2^2} + \frac{1+\mu}{2} \frac{\partial^2 v}{\partial x_1 x_2} \right)(\bm{x}) = 0, \quad \bm{x} \in \Omega, \\
		& \frac{E}{1-\mu^2} \left( \frac{\partial^2 v}{\partial x_2^2} + \frac{1-\mu}{2} \frac{\partial^2 v}{\partial x_1^2} + \frac{1+\mu}{2} \frac{\partial^2 u}{\partial x_1 x_2} \right)(\bm{x}) = 0, \quad \bm{x} \in \Omega, \\
		& \frac{E}{1-\mu^2} \left[ n_1 \left(\frac{\partial u}{\partial x_1} + \mu \frac{\partial v}{\partial x_2}\right) + n_2 \frac{1-\mu}{2} \left(\frac{\partial u}{\partial x_2} + \frac{\partial v}{\partial x_1}\right)\right](\bm{x}) = h_1(\bm{x}), \quad \bm{x} \in \partial \Gamma_1, \\
		& \frac{E}{1-\mu^2} \left[ n_2 \left(\frac{\partial v}{\partial x_2} + \mu \frac{\partial u}{\partial x_1}\right) + n_1 \frac{1-\mu}{2} \left(\frac{\partial v}{\partial x_1} + \frac{\partial u}{\partial x_2}\right)\right](\bm{x}) = h_2(\bm{x}), \quad \bm{x} \in \partial \Gamma_1, \\
		& u(\bm{x}) = 0, \quad \bm{x} \in \partial \Gamma_2, \\
		& v(\bm{x}) = 0, \quad \bm{x} \in \partial \Gamma_2.
	\end{aligned}
\end{equation}
Here, $u$ and $v$ denote the displacements in $x_1$- and $x_2$-directions, respectively. $E$ represents the Young's modulus and $\mu$ represents the Poisson's ratio. The vector $(n_1, n_2)^T$ is the unit outward normal vector on $\Gamma_1$. $h_1(\bm{x}) = -q_1n_1 + q_2n_2$, $h_2(\bm{x}) = -q_1n_2 - q_2n_1$. Owing to the symmetry of the problem, the Airy stress function approach \cite{sadd2009elasticity} can be employed to derive the analytical solutions as follows:
\begin{equation}
	\label{eq:L2}
	\begin{aligned}
		&u(\bm{x}) = A \left( \frac{a^2}{x_1^2 + x_2^2} - 1 \right) x_1 + B \left( 1 - \frac{a^2}{x_1^2 + x_2^2} \right) x_2, \\
		&v(\bm{x}) = A \left( \frac{a^2}{x_1^2 + x_2^2} - 1 \right) x_2 - B \left( 1 - \frac{a^2}{x_1^2 + x_2^2} \right) x_1, \\
		&A = \frac{1}{E}\frac{ q_1 (1 - \mu^2) b^2}{b^2 (1 + \mu) + a^2 (1 - \mu)}, \\
		&B = \frac{1}{E} \frac{q_2 (1 + \mu) b^2}{a^2}.
	\end{aligned}
\end{equation}

Within the finite difference method (FDM) framework, this work employs the Eq.\eqref{eq:la2} to approximate two-order mixed partial derivatives. Set the step $\Delta h = 1\times10^{-5}$.
\begin{equation}
	\label{eq:la2}
	\begin{aligned}
		&\frac{\partial^2 u(x_1, x_2)}{\partial x_1x_2} = \frac{\mathfrak{u}_1 - \mathfrak{u}_2}{\Delta 4h ^2},\\
		&\mathfrak{u}_1 = u(x_1 + \Delta h, x_2 + \Delta h) - u(x_1 + \Delta h, x_2 - \Delta h),\\
		&\mathfrak{u}_2 = u(x_1 - \Delta h, x_2 + \Delta h) - u(x_1 - \Delta h, x_2 - \Delta h).
	\end{aligned}
\end{equation}

We approximate the displacement fields $u$ and $v$ using networks ${u_\rho }(\bm{x}) = \sum\limits_{i = 1}^{N_u} {{\alpha _i}{\varphi^u _i}(\bm{x})}$ and ${v_\rho }(\bm{x}) = \sum\limits_{i = 1}^{N_v} {{\alpha _i}{\varphi^v _i}(\bm{x})}$, respectively, where $N_u$ and $N_v$ denote the number of neurons in the hidden layers of ${u_\rho }(\bm{x})$ and ${v_\rho }(\bm{x})$. Randomly select $N_1$, $N_2$ and $N_3$ points in the region $\Omega$, outer boundary $\Gamma_1$ and inner boundary $\Gamma_2$. The Eq.\eqref{eq:L1} is discretized to formulate the following linear system:
\begin{equation}
	\label{eq:L3}
	\begin{bmatrix}
		\mathcal{F}^1\\
		\mathcal{F}^2\\
		{\lambda _1}\mathcal{B}^1\\
		{\lambda _1}\mathcal{B}^2\\
		{\lambda _2}\mathcal{B}^3\\
		{\lambda _2}\mathcal{B}^4
	\end{bmatrix} \bm{\alpha}  = \begin{bmatrix}
		\mathfrak{F}^1\\
		\mathfrak{F}^2\\
		{\lambda _1}\mathfrak{B}^1\\
		{\lambda _1}\mathfrak{B}^2\\
		{\lambda _2}\mathfrak{B}^3\\
		{\lambda _2}\mathfrak{B}^4
	\end{bmatrix},
\end{equation}
where ${\lambda _1}$ and ${\lambda _2}$ represent the weights of the outer and inner boundary conditions, respectively. $\mathcal{F}^1$, $\mathcal{F}^2$, $\mathcal{B}^1$, $\mathcal{B}^2$, $\mathcal{B}^3$ and $\mathcal{B}^4$ are matrices of order $N_1\times(N_u+N_v)$, $N_1\times(N_u+N_v)$, $N_2\times(N_u+N_v)$, $N_2\times(N_u+N_v)$, $N_3\times(N_u+N_v)$ and $N_3\times(N_u+N_v)$, respectively. $\mathfrak{F}^1$, $\mathfrak{F}^2$, $\mathfrak{B}^1$, $\mathfrak{B}^2$, $\mathfrak{B}^3$ and $\mathfrak{B}^4$ are vectors of order $N_1$, $N_1$, $N_2$, $N_2$, $N_3$ and $N_3$. Specifically,
\begin{equation*}
	\mathcal{F}^1_{n,m}=\left\{
	\begin{aligned}
		&\frac{E}{1-\mu^2} \left( \frac{\partial^2 \varphi^u_m}{\partial x_{n1}^2} + \frac{1-\mu}{2} \frac{\partial^2 \varphi^u_m}{\partial x_{n2}^2} \right)(\bm{x}_n),\;\;\bm{x}_n\in \Omega,\;1\leq m\leq N_u, \\
		&\frac{E}{1-\mu^2} \left( \frac{1+\mu}{2} \frac{\partial^2 \varphi^v_m}{\partial x_{n1} x_{n2}} \right)(\bm{x}_n),\;\;\bm{x}_n\in \Omega,\;N_u+1\leq m\leq N_u+N_v,
	\end{aligned}\right.
\end{equation*}
\begin{equation*}
	\mathcal{F}^2_{n,m}=\left\{
	\begin{aligned}
		&\frac{E}{1-\mu^2} \left( \frac{1+\mu}{2} \frac{\partial^2 \varphi^u_m}{\partial x_{n1} x_{n2}} \right)(\bm{x}_n),\;\;\bm{x}_n\in \Omega,\;1\leq m\leq N_u, \\
		&\frac{E}{1-\mu^2} \left( \frac{\partial^2 \varphi^v_m}{\partial x_{n2}^2} + \frac{1-\mu}{2} \frac{\partial^2 \varphi^v_m}{\partial x_{n1}^2} \right)(\bm{x}_n),\;\;\bm{x}_n\in \Omega,\;N_u+1\leq m\leq N_u+N_v,
	\end{aligned}\right.
\end{equation*}
\begin{equation*}
	\mathcal{B}^1_{n,m}=\left\{
	\begin{aligned}
		&\frac{E}{1-\mu^2} \left(n_1 \frac{\partial \varphi^u_m}{\partial x_{n1}} + n_2 \frac{1-\mu}{2} \frac{\partial \varphi^u_m}{\partial x_{n2}}\right)(\bm{x}_n),\;\;\bm{x}_n\in \Omega,\;1\leq m\leq N_u, \\
		&\frac{E}{1-\mu^2} \left( n_1 \mu \frac{\partial \varphi^v_m}{\partial x_{n2}} + n_2 \frac{1-\mu}{2} \frac{\partial \varphi^v_m}{\partial x_{n1}}\right)(\bm{x}_n),\;\;\bm{x}_n\in \Omega,\;N_u+1\leq m\leq N_u+N_v,
	\end{aligned}\right.
\end{equation*}
\begin{equation*}
	\mathcal{B}^2_{n,m}=\left\{
	\begin{aligned}
		&\frac{E}{1-\mu^2} \left( n_2 \mu \frac{\partial \varphi^u_m}{\partial x_{n1}} + n_1 \frac{1-\mu}{2} \frac{\partial \varphi^u_m}{\partial x_{n2}}\right)(\bm{x}_n),\;\;\bm{x}_n\in \Omega,\;1\leq m\leq N_u, \\
		&\frac{E}{1-\mu^2} \left( n_2 \frac{\partial \varphi^v_m}{\partial x_{n2}} + n_1 \frac{1-\mu}{2} \frac{\partial \varphi^v_m}{\partial x_{n1}}\right)(\bm{x}_n),\;\;\bm{x}_n\in \Omega,\;N_u+1\leq m\leq N_u+N_v,
	\end{aligned}\right.
\end{equation*}
\begin{equation*}
	\mathcal{B}^3_{n,m}=\left\{
	\begin{aligned}
		&\varphi^u_m(\bm{x}_n),\;\;\bm{x}_n\in \Omega,\;1\leq m\leq N_u, \\
		&0,\;\;\bm{x}_n\in \Omega,\;N_u+1\leq m\leq N_u+N_v,
	\end{aligned}\right.
\end{equation*}
\begin{equation*}
	\mathcal{B}^4_{n,m}=\left\{
	\begin{aligned}
		&0,\;\;\bm{x}_n\in \Omega,\;1\leq m\leq N_u, \\
		&\varphi^v_m(\bm{x}_n),\;\;\bm{x}_n\in \Omega,\;N_u+1\leq m\leq N_u+N_v,
	\end{aligned}\right.
\end{equation*}
\begin{equation*}
	\mathfrak{F}^1_n = 0, \;\; \mathfrak{F}^2_n = 0, \;\; \mathfrak{B}^1_n = h_1(\bm{x}_n), \;\; \mathfrak{B}^2_n = h_2(\bm{x}_n), \;\; \mathfrak{B}^3_n = 0, \;\; \mathfrak{B}^4_n = 0.
\end{equation*}

\begin{figure}[!htb]
	\centering
	\begin{minipage}[t]{0.24\textwidth}
		\centering
		\includegraphics[width=\textwidth]{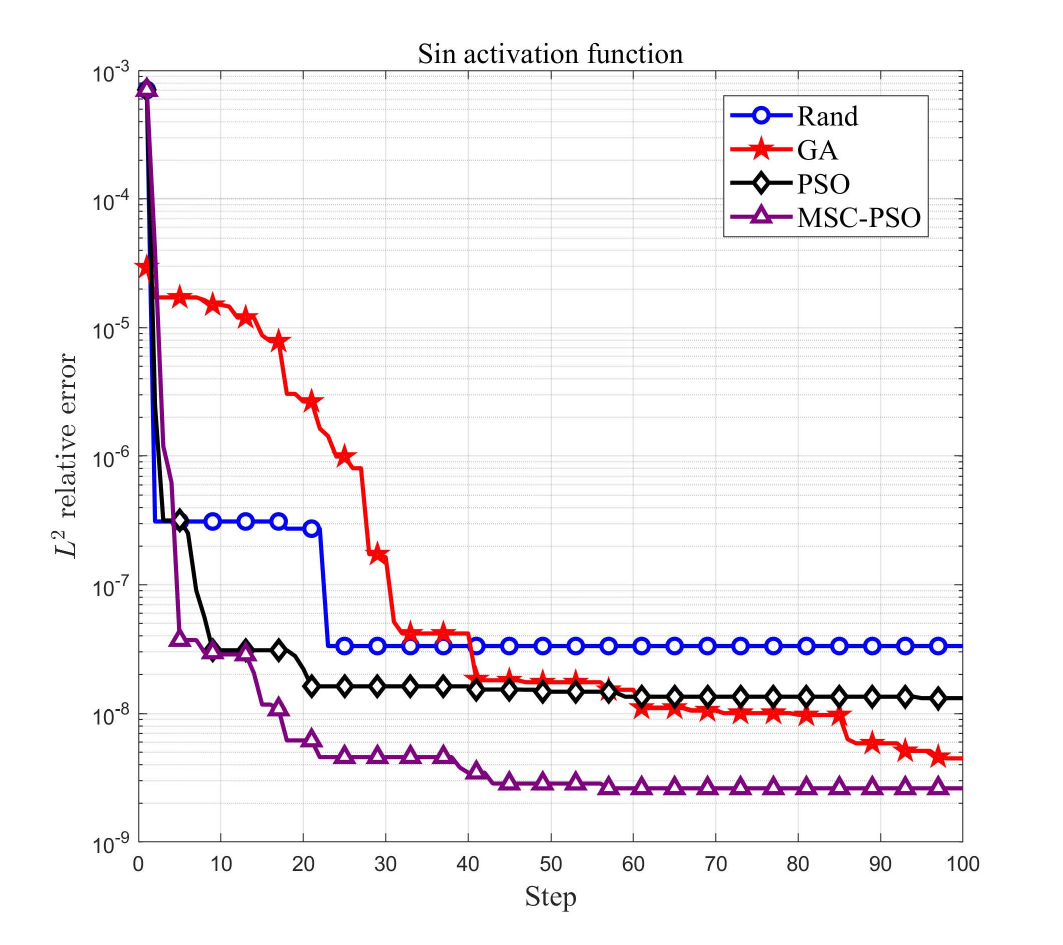}
		(a)
	\end{minipage}
	\hfill
	\begin{minipage}[t]{0.24\textwidth}
		\centering
		\includegraphics[width=\textwidth]{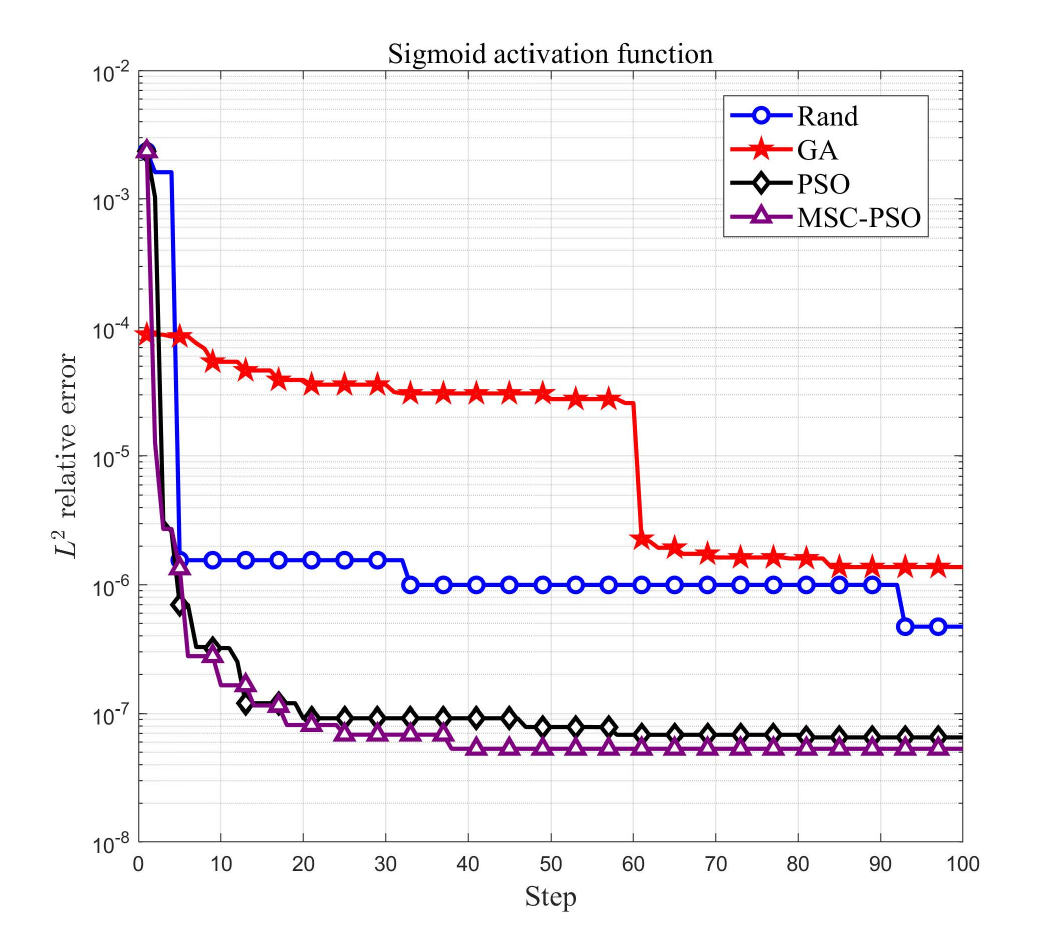}
		(b)
	\end{minipage}
	\hfill
	\begin{minipage}[t]{0.24\textwidth}
		\centering
		\includegraphics[width=\textwidth]{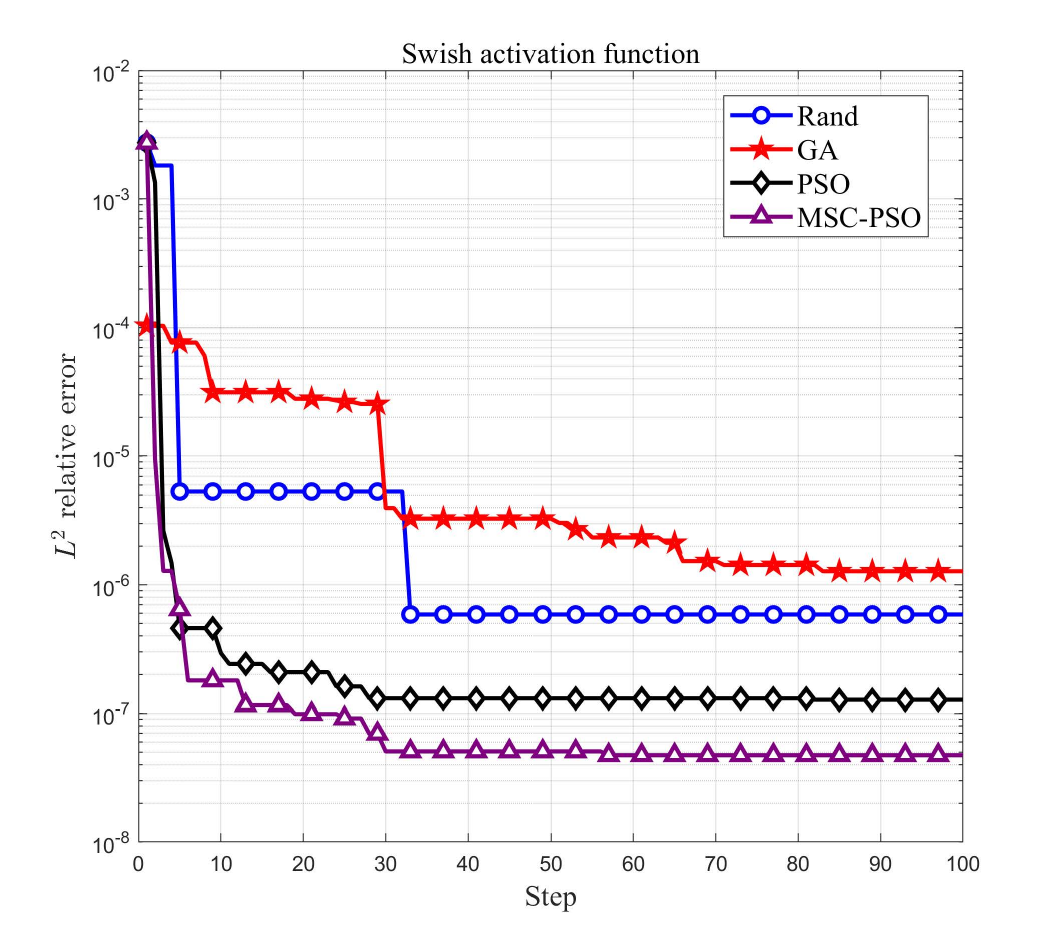}
		(c)
	\end{minipage}
	\hfill
	\begin{minipage}[t]{0.24\textwidth}
		\centering
		\includegraphics[width=\textwidth]{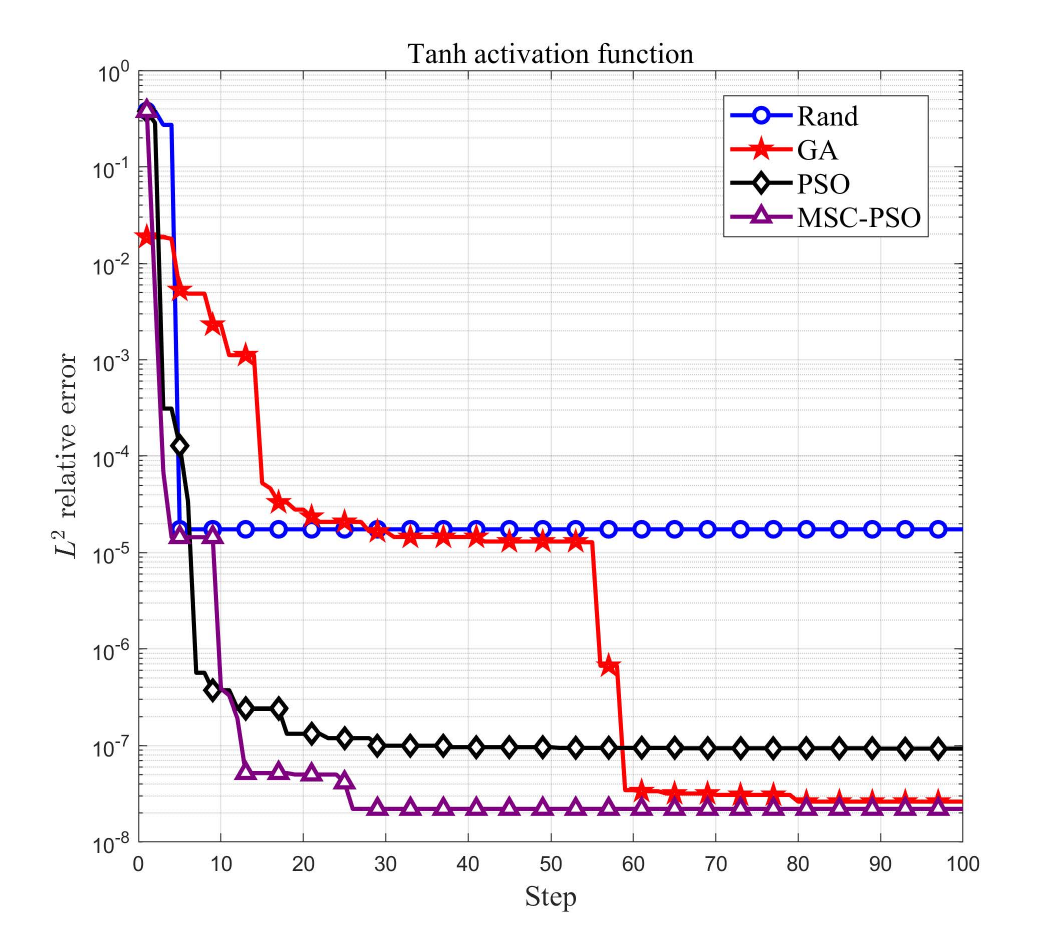}
		(d)
	\end{minipage}
	\caption{The $L^2$ relative error of different activation functions under different intelligent optimization algorithms. (a) Sin. (b) Sigmoid. (c) Swish. (d) Tanh.}
	\label{fig:L2}
\end{figure}

\begin{figure}[!htb]
	\centering
	\includegraphics[width=1\linewidth]{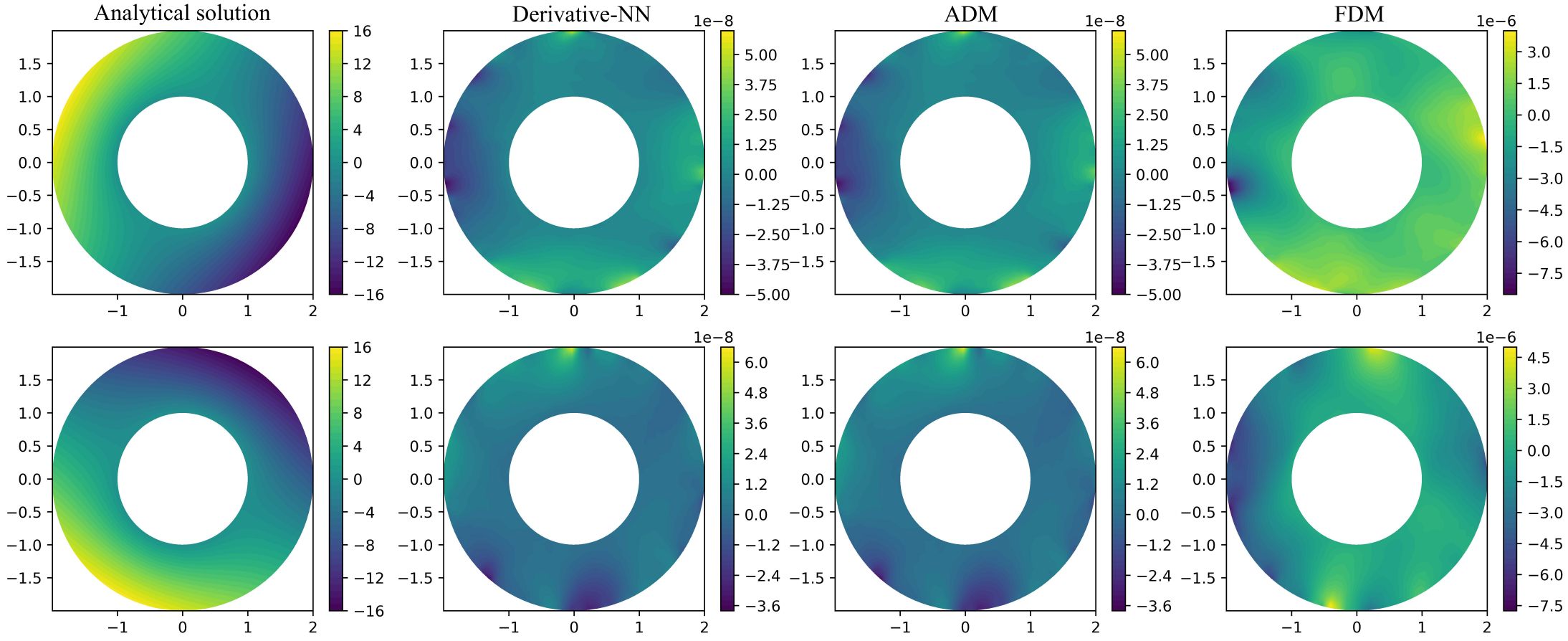}
	\caption{The analytical solution (first column) of Lam\'e equations and absolution error (last three columns), the first line is the variable $u$, and the second line is the variable $v$.}
	\label{fig:L3}
\end{figure}

In this experiment, set $E=2.1$, $\mu=0.25$, $q_1=30$, $q_2=2$, $a=1$ and $b=2$. Table \ref{tab:L1} outlines the parameter optimization ranges for all hyperparameters. In this experiment, the outer-layer optimization objective is defined as the sum of the $L^2$ relative errors of variables $u$ and $v$. Fig.\ref{fig:L2} presents the convergence profiles of the $L^2$ relative error for solutions to the Lam\'e equations obtained using distinct activation functions and intelligent optimization algorithms. As shown, the MSC-PSO algorithm achieves accelerated convergence rates and identifies hyperparameter combinations that enhance computational accuracy. It is worth noting that when the SO-PIFRNN framework is combined with the frequency domain feature extraction ability of the sin activation function, the computational accuracy is better than traditional activation functions. The optimized hyperparameters for SO-PIFRNN are determined as follows: $N_u = 1374$, $N_v = 1526$, $\omega_1=20.358$, $\omega_2=20.882$, $\lambda_1 = 8819.374$, $\lambda_2 = 7173.623$, $N_1 = 5000$, $N_2 = 2619$ and $N_3 = 1513$. Table \ref{tab:L2} systematically compares the computational accuracy and efficiency of Derivative-NN against the ADM and FDM under optimal hyperparameter configurations. Fig.\ref{fig:L3} additionally illustrates the analytical solution, along with the absolute errors between the three numerical methods and the analytical solution. The numerical experimental results demonstrate that Derivative-NN exhibits advantages in both computational accuracy and efficiency. While ADM retains comparable numerical precision to Derivative-NN, it demands 35 times longer computational time. Notably, due to the accumulation of truncation errors during the discretization process, FDM exhibits errors two orders of magnitude higher than those of Derivative-NN and ADM respectively.
\begin{table}[!htb]
	\centering
	\caption{The optimization range of hyperparameter.}\label{tab:L1}
	\begin{tabular}{cccccc}
		\hline
		Hyperparameters	& $N_u$, $N_v$ & $\omega_1$, $\omega_2$ & $\lambda_1$, $\lambda_2$ & $N_1$ & $N_2$, $N_3$ \\ \hline
		Range	& $\left[ {10,{\rm{ 2000}}} \right]$ &  $\left[ {0.0001,{\rm{ 50}}}\right]$  &  $\left[ {0.0001,{\rm{ 10000}}} \right]$   &    $\left[  {10,{\rm{ 5000}}}\right]$   &    $\left[  {10,{\rm{ 3000}}}\right]$\\ \hline
	\end{tabular}
\end{table}

\begin{table}[!htb]
	\centering
	\caption{Comparison of calculation accuracy and efficiency.}\label{tab:L2}
	\begin{tabular}{cccc}
		\hline
		& Derivative-NN & ADM & FDM \\ \hline
		$L^2$ relative error of $u$ & $1.534\times10^{-9}$   &$1.524\times10^{-9}$   &$1.898\times10^{-7}$  \\
		$L^2$ relative error of $v$ & $1.082\times10^{-9}$   &$1.091\times10^{-9}$   &$2.516\times10^{-7}$  \\
		Time/s               & 4.215    & 141.531  & 4.537   \\ \hline
	\end{tabular}
\end{table}

\subsection{Example 6: nonlinear Helmholtz equation}

We consider the two-dimensional nonlinear Helmholtz equation on the domain $\Omega = [0, 1]^2$,
\begin{equation}
	\label{eq:no1}
	\begin{aligned}
		& \frac{\partial^2 u(\bm{x})}{\partial x_1^2} + \frac{\partial^2 u(\bm{x})}{\partial x_2^2} - 100u(\bm{x}) + 10\cos(u(\bm{x})) = f(\bm{x}), \quad \bm{x} \in \Omega,\\
		& u(x_1, 0) = g_1(\bm{x}), \quad u(x_1, 1) = g_2(\bm{x}), \quad u(0, x_2) = g_3(\bm{x}), \quad u(1, x_2)= g_4(\bm{x}),
	\end{aligned}
\end{equation}
where $u(\bm{x})$ represents the field function to be determined, $f(\bm{x})$ denotes the prescribed source term and $g_i(\bm{x})(1\leq i \leq4)$ specify the Dirichlet boundary conditions. Through proper selection of the source term $f(\bm{x})$ and boundary conditions $g_i(\bm{x})$, this system permits an analytical solution expressed as:
\begin{equation*}
	u(\bm{x}) = 4 \cos(3\pi x_1^2) \sin(3\pi x_2^2).
\end{equation*}

This study employs the Newton iteration method \cite{dong2022numerical} to compute the numerical solution of nonlinear equations, the computational process is controlled by restricting the maximum number of iterations to 10. Following an analogous discretization scheme, $N_1$ and $N_2$ points are randomly selected in the region and on the boundary. The nonlinear Helmholtz equation is subsequently solved using the SO-PIFRNN method combined with the Newton iteration method. Table \ref{tab:no1} provides the optimization ranges corresponding to each hyperparameter.
\begin{table}[!htb]
	\centering
	\caption{The optimization range of hyperparameter.}\label{tab:no1}
	\begin{tabular}{ccccc}
		\hline
		Hyperparameters	& $N$ & $\omega$ & $\lambda$ & $N_1$, $N_2$  \\ \hline
		Range	& $\left[ {10,{\rm{ 2000}}} \right]$ &  $\left[ {0.0001,{\rm{ 100}}}\right]$  &  $\left[ {0.0001,{\rm{ 10000}}} \right]$   &    $\left[  {10,{\rm{ 3000}}}\right]$  \\ \hline
	\end{tabular}
\end{table}

\begin{figure}[!htb]
	\centering
	\begin{minipage}[t]{0.24\textwidth}
		\centering
		\includegraphics[width=\textwidth]{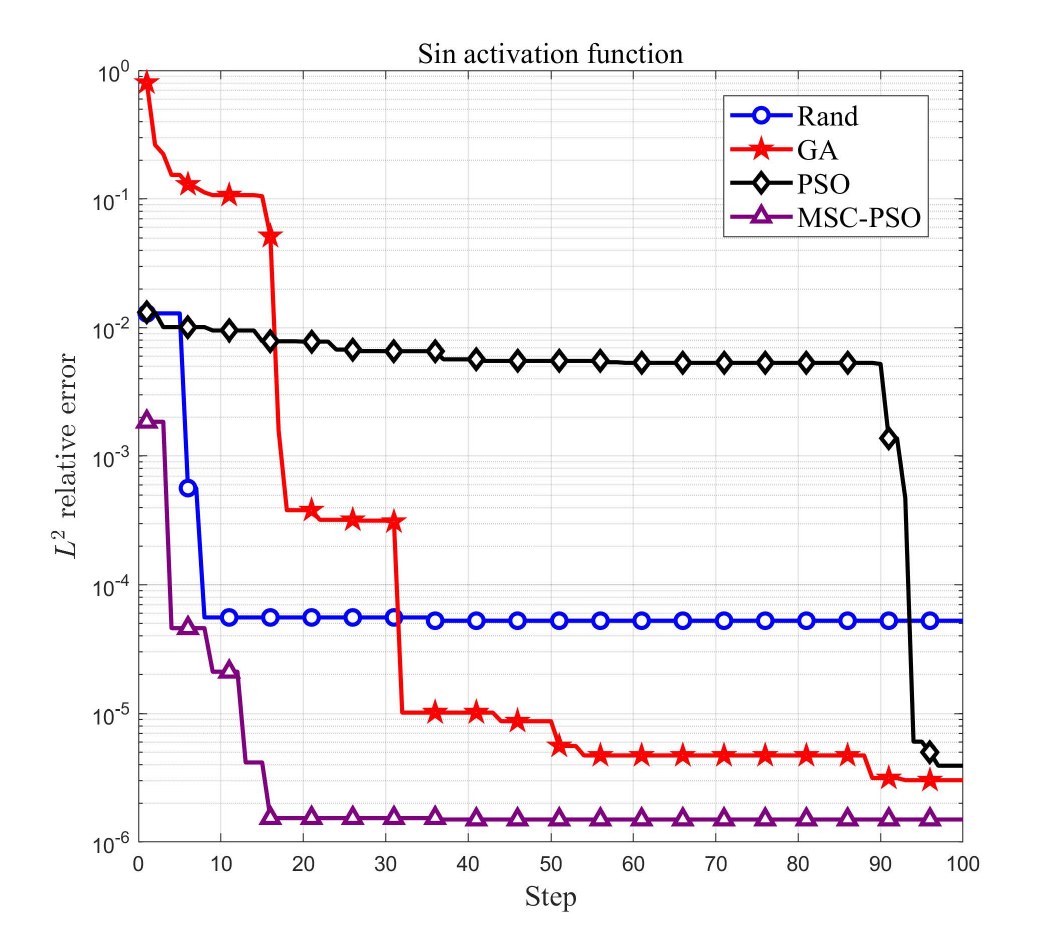}
		(a)
	\end{minipage}
	\hfill
	\begin{minipage}[t]{0.24\textwidth}
		\centering
		\includegraphics[width=\textwidth]{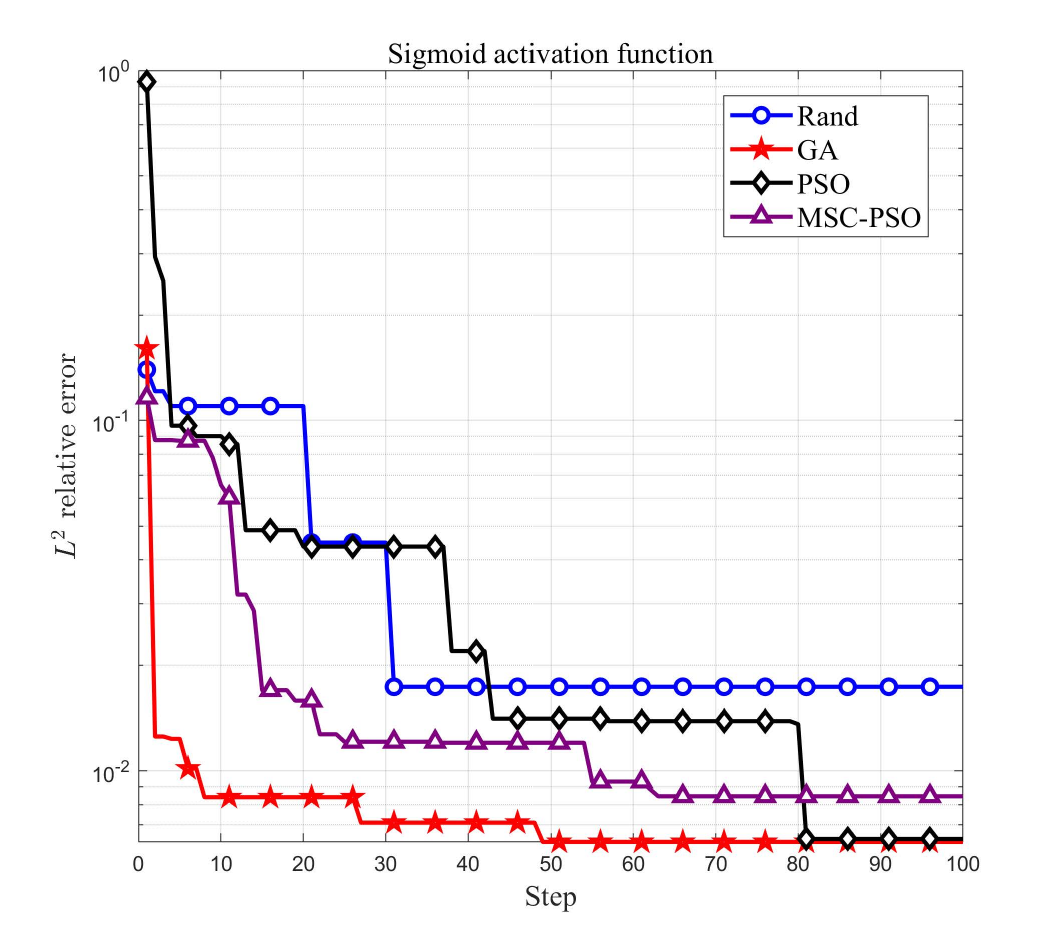}
		(b)
	\end{minipage}
	\hfill
	\begin{minipage}[t]{0.24\textwidth}
		\centering
		\includegraphics[width=\textwidth]{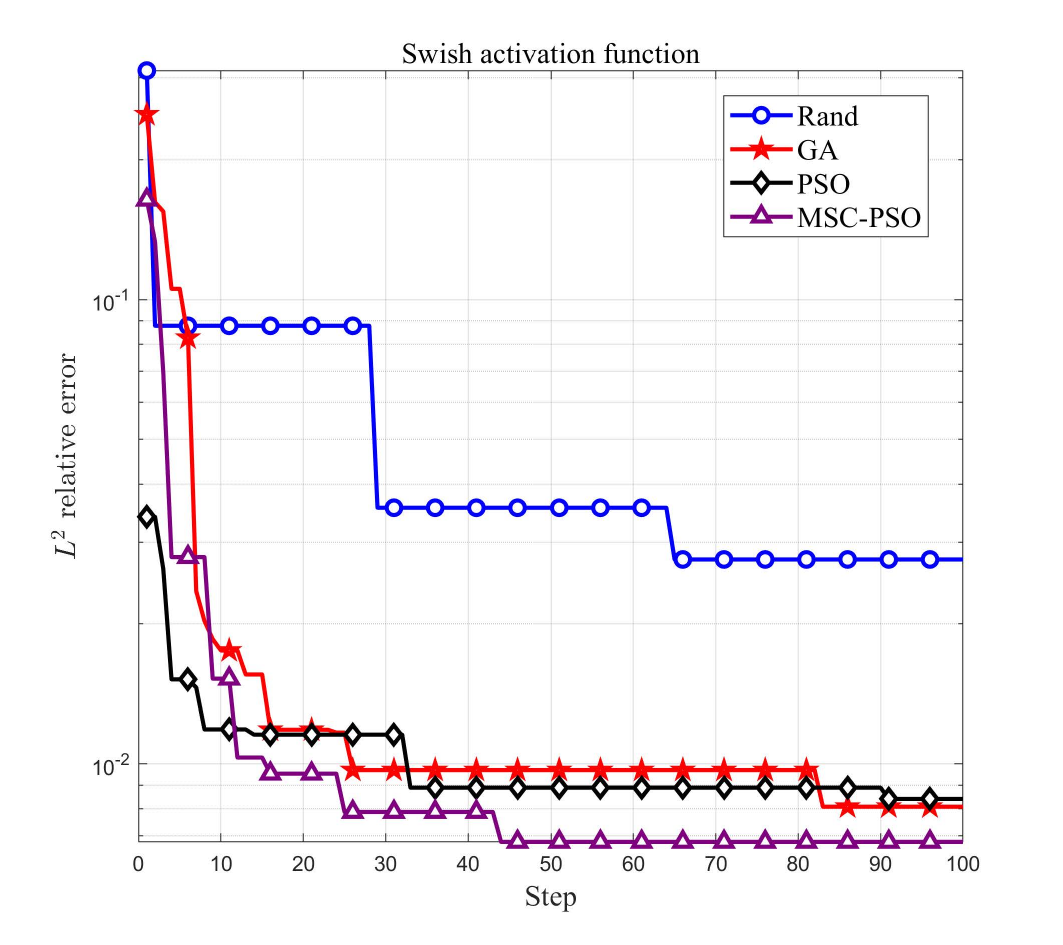}
		(c)
	\end{minipage}
	\hfill
	\begin{minipage}[t]{0.24\textwidth}
		\centering
		\includegraphics[width=\textwidth]{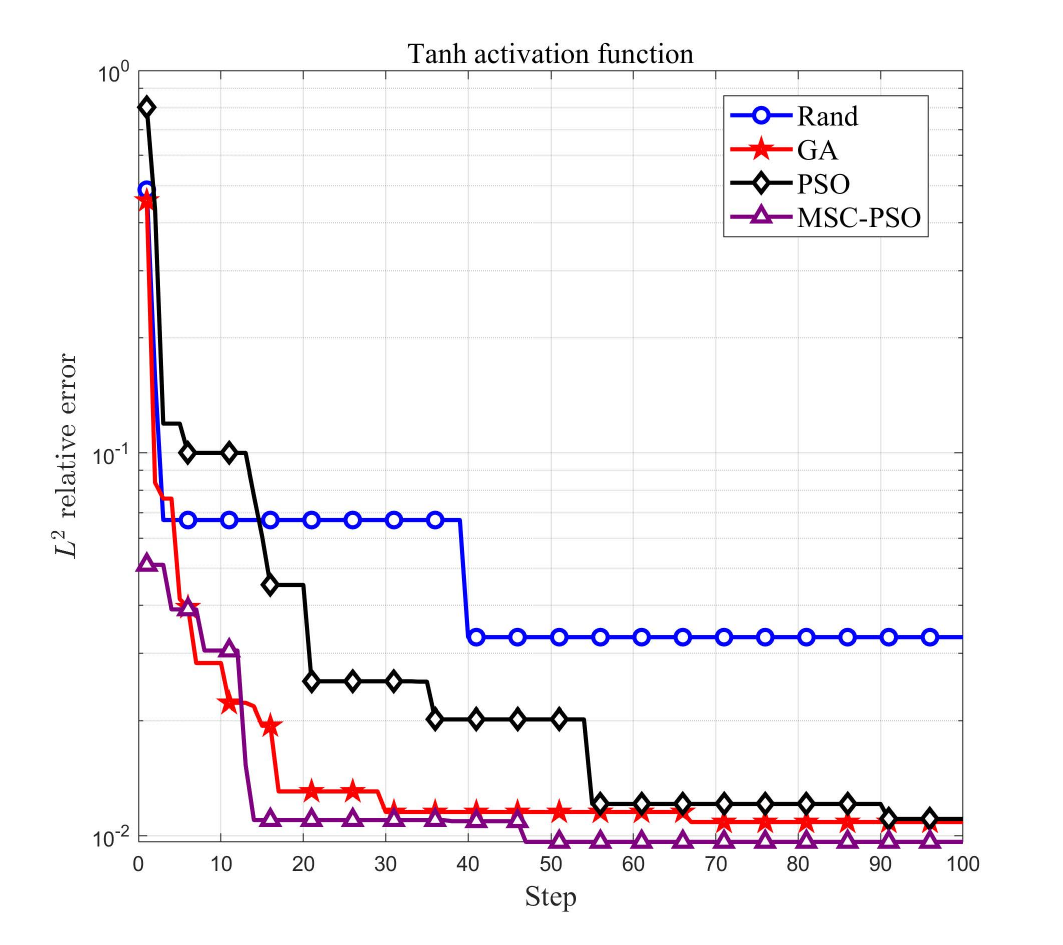}
		(d)
	\end{minipage}
	\caption{The $L^2$ relative error of different activation functions under different intelligent optimization algorithms. (a) Sin. (b) Sigmoid. (c) Swish. (d) Tanh.}
	\label{fig:no1}
\end{figure}

\begin{figure}[!htb]
	\centering
	\includegraphics[width=1\linewidth]{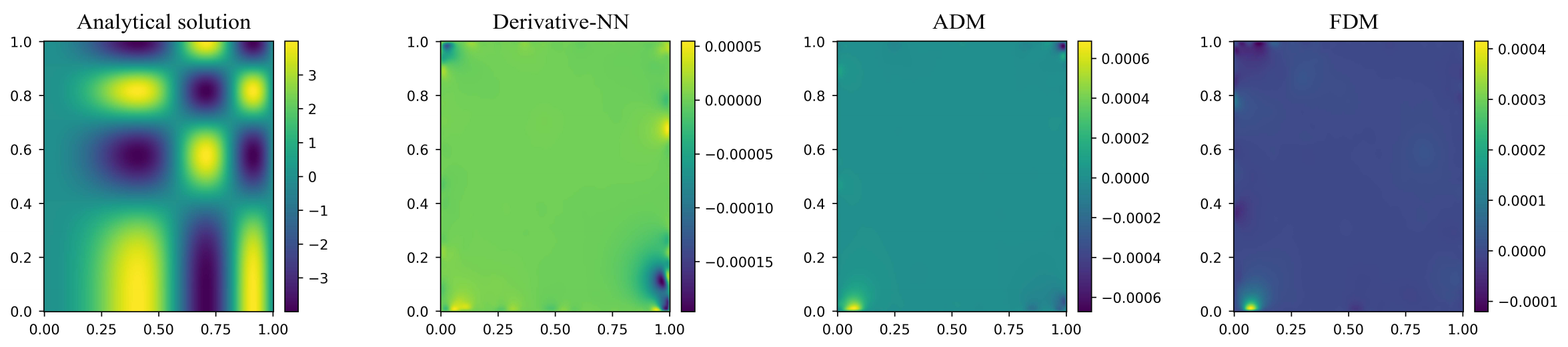}
	\caption{The analytical solution (first column) of plate deflection of nonlinear Helmholtz equation and absolution error (last three columns).}
	\label{fig:no2}
\end{figure}

Fig.\ref{fig:no1} systematically compares the convergence characteristics of $L^2$ relative error for solutions of the Helmholtz equation obtained using various activation functions and intelligent optimization algorithms. As depicted in the figure, the MSC-PSO algorithm identifies hyperparameter combinations that enhance computational accuracy in most cases. Notably, the SO-PIFRNN architecture integrated with sin activation functions demonstrates about four orders of magnitude improvement in numerical precision over traditional activation functions, attributable to the frequency-domain feature extraction capability of the sin activation function. The optimized hyperparameters for SO-PIFRNN are determined as $N = 2000$, $\omega =87.857$, $\lambda = 585.640$, $N_1 = 2995$ and $N_2 = 2974$. Table \ref{tab:no2} provides a systematic comparison of the computational accuracy and efficiency of the Derivative-NN method with the ADM of neural networks and the FDM under the optimal hyperparameter configuration. Fig.\ref{fig:no2} presents the analytical solution, alongside the absolute errors computed between the three numerical methods and the analytical solution. The numerical experiments demonstrate that the Derivative-NN method maintains a certain advantage in terms of computational efficiency. However, due to the constraints imposed by the truncation errors from the linearization of the Newton iteration method, the three numerical methods do not exhibit significant differences in terms of computational accuracy.

\begin{table}[!htb]
	\centering
	\caption{Comparison of calculation accuracy and efficiency.}\label{tab:no2}
	\begin{tabular}{cccc}
		\hline
		& Derivative-NN & ADM & FDM \\ \hline
		$L^2$ relative error & $1.494\times10^{-6}$   &$1.793\times10^{-5}$   &$8.689\times10^{-6}$  \\
		Time/s               & 18.340   & 142.001  &  26.378 \\ \hline
	\end{tabular}
\end{table}

\section{Summary}
\label{sec:4}

This paper proposes a self-optimization physics-informed Fourier-features randomized neural network (SO-PIFRNN) framework that significantly enhances the computational accuracy of PDEs through automated search for optimal hyperparameter combinations in PIFRNN. The framework adopts a bi-level optimization architecture: the outer-level automatically optimizes PIFRNN hyperparameters via the MSC-PSO algorithm, while the inner-level determines the output layer weights of the PIFRNN through least square method. The main innovations in algorithm design include: (1) Fourier activation function is proposed as the network activation function. Experimental results show that compared with traditional activation functions (sigmoid, tanh, swish), Fourier activation function significantly enhances the ability of the network to capture multi-frequency components of the solution function. (2) A novel derivative neural network method is developed, and the differential operator is calculated by an improved forward propagation mechanism, which improves the calculation accuracy and efficiency of PIFRNN method. (3) Design MSC-PSO, which integrates dynamic parameter adjustment, elitist and mutation strategies. Experimental data show that compared with random search, genetic algorithm and standard PSO, MSC-PSO achieves the optimal combination of hyperparameters in most test cases.

Future research will focus on three directions. Firstly, to address the challenge of balancing computational efficiency and accuracy in randomized neural networks for solving PDE, we will introduce multi-objective optimization algorithms. This involves constructing the multi-objective space in the loss function that incorporates numerical accuracy, computational time and memory consumption, aiming to achieve optimal accuracy under computational resource constraints. Secondly, based on spectral analysis theory, we will determine the initialization range of neural network weights, leveraging prior knowledge to significantly enhance the convergence speed of intelligent algorithms. Finally, to tackle the strongly nonlinear characteristics of multiscale, multiphysics coupled systems, we intend to develop the SO-PIFRNN algorithm with adaptive domain decomposition capabilities.

\section*{CRediT authorship contribution statement}
\textbf{Jiale Linghu}: Methodology, Software, Writing-original draft, Visualization, Conceptualization, Investigation.
\textbf{Weifeng Gao}: Conceptualization, Funding acquisition, Investigation, Supervision, Writing-review \& editing.
\textbf{Hao Dong}: Methodology, Software, Writing-review \& editing, Conceptualization, Investigation, Funding acquisition, Supervision.
\textbf{Yufeng Nie}: Methodology, Conceptualization, Supervision.

\section*{Data availability}
No data was used for the research described in the article.

\section*{Declarations}
The authors affirm that they have no known financial or interpersonal conflicts that might have looked to have influenced the research presented in this study.

\section*{Acknowledgments}
The authors gratefully acknowledge the support of the National Natural Science Foundation of China (Nos.\hspace{1mm}62276202 and 12471387), the Natural Science Basic Research Plan in Shaanxi Province of China under Grant (No.\hspace{1mm}2022JQ-670), the Young Talent Fund of Association for Science and Technology in Shaanxi, China (No.\hspace{1mm}20220506), the Fundamental Research Funds for the Central Universities (Nos.\hspace{1mm}ZYTS24071, ZYTS24072 and YJSJ24002), the Xidian University Specially Funded Project for Interdisciplinary Exploration (No.\hspace{1mm}TZJH2024008).

\section*{References}

\begin{thebibliography}{41}
\expandafter\ifx\csname natexlab\endcsname\relax\def\natexlab#1{#1}\fi
\providecommand{\url}[1]{\texttt{#1}}
\providecommand{\href}[2]{#2}
\providecommand{\path}[1]{#1}
\providecommand{\DOIprefix}{doi:}
\providecommand{\ArXivprefix}{arXiv:}
\providecommand{\URLprefix}{URL: }
\providecommand{\Pubmedprefix}{pmid:}
\providecommand{\doi}[1]{\href{http://dx.doi.org/#1}{\path{#1}}}
\providecommand{\Pubmed}[1]{\href{pmid:#1}{\path{#1}}}
\providecommand{\bibinfo}[2]{#2}
\ifx\xfnm\relax \def\xfnm[#1]{\unskip,\space#1}\fi
\bibitem[{Cai et~al.(2021)Cai, Mao, Wang, Yin, and
  Karniadakis}]{cai2021physics}
\bibinfo{author}{S.~Cai}, \bibinfo{author}{Z.~Mao}, \bibinfo{author}{Z.~Wang},
  \bibinfo{author}{M.~Yin}, \bibinfo{author}{G.~E. Karniadakis},
\newblock \bibinfo{title}{Physics-informed neural networks (pinns) for fluid
  mechanics: A review},
\newblock \bibinfo{journal}{Acta Mechanica Sinica} \bibinfo{volume}{37}
  (\bibinfo{year}{2021}) \bibinfo{pages}{1727--1738}.
\bibitem[{Linghu et~al.(2025{\natexlab{a}})Linghu, Gao, Dong, and
  Nie}]{linghu2025higher}
\bibinfo{author}{J.~Linghu}, \bibinfo{author}{W.~Gao},
  \bibinfo{author}{H.~Dong}, \bibinfo{author}{Y.~Nie},
\newblock \bibinfo{title}{Higher-order multi-scale physics-informed neural
  network (homs-pinn) method and its convergence analysis for solving elastic
  problems of authentic composite materials},
\newblock \bibinfo{journal}{Journal of Computational and Applied Mathematics}
  \bibinfo{volume}{456} (\bibinfo{year}{2025}{\natexlab{a}})
  \bibinfo{pages}{116223}.
\bibitem[{Linghu et~al.(2025{\natexlab{b}})Linghu, Dong, Nie, and
  Cui}]{linghu2025higherDRM}
\bibinfo{author}{J.~Linghu}, \bibinfo{author}{H.~Dong},
  \bibinfo{author}{Y.~Nie}, \bibinfo{author}{J.~Cui},
\newblock \bibinfo{title}{Higher-order multi-scale deep ritz method (homs-drm)
  and its convergence analysis for solving thermal transfer problems of
  composite materials},
\newblock \bibinfo{journal}{Computational Mechanics} \bibinfo{volume}{75}
  (\bibinfo{year}{2025}{\natexlab{b}}) \bibinfo{pages}{71--95}.
\bibitem[{Khan and Lowther(2022)}]{khan2022physics}
\bibinfo{author}{A.~Khan}, \bibinfo{author}{D.~A. Lowther},
\newblock \bibinfo{title}{Physics informed neural networks for electromagnetic
  analysis},
\newblock \bibinfo{journal}{IEEE Transactions on Magnetics}
  \bibinfo{volume}{58} (\bibinfo{year}{2022}) \bibinfo{pages}{1--4}.
\bibitem[{Raissi et~al.(2019)Raissi, Perdikaris, and
  Karniadakis}]{raissi2019physics}
\bibinfo{author}{M.~Raissi}, \bibinfo{author}{P.~Perdikaris},
  \bibinfo{author}{G.~E. Karniadakis},
\newblock \bibinfo{title}{Physics-informed neural networks: A deep learning
  framework for solving forward and inverse problems involving nonlinear
  partial differential equations},
\newblock \bibinfo{journal}{Journal of Computational physics}
  \bibinfo{volume}{378} (\bibinfo{year}{2019}) \bibinfo{pages}{686--707}.
\bibitem[{Meng et~al.(2020)Meng, Li, Zhang, and Karniadakis}]{meng2020ppinn}
\bibinfo{author}{X.~Meng}, \bibinfo{author}{Z.~Li}, \bibinfo{author}{D.~Zhang},
  \bibinfo{author}{G.~E. Karniadakis},
\newblock \bibinfo{title}{Ppinn: Parareal physics-informed neural network for
  time-dependent pdes},
\newblock \bibinfo{journal}{Computer Methods in Applied Mechanics and
  Engineering} \bibinfo{volume}{370} (\bibinfo{year}{2020})
  \bibinfo{pages}{113250}.
\bibitem[{Yang et~al.(2021)Yang, Meng, and Karniadakis}]{yang2021b}
\bibinfo{author}{L.~Yang}, \bibinfo{author}{X.~Meng}, \bibinfo{author}{G.~E.
  Karniadakis},
\newblock \bibinfo{title}{B-pinns: Bayesian physics-informed neural networks
  for forward and inverse pde problems with noisy data},
\newblock \bibinfo{journal}{Journal of Computational Physics}
  \bibinfo{volume}{425} (\bibinfo{year}{2021}) \bibinfo{pages}{109913}.
\bibitem[{Jagtap and Karniadakis(2020)}]{jagtap2020extended}
\bibinfo{author}{A.~D. Jagtap}, \bibinfo{author}{G.~E. Karniadakis},
\newblock \bibinfo{title}{Extended physics-informed neural networks (xpinns): A
  generalized space-time domain decomposition based deep learning framework for
  nonlinear partial differential equations},
\newblock \bibinfo{journal}{Communications in Computational Physics}
  \bibinfo{volume}{28} (\bibinfo{year}{2020}).
\bibitem[{Yu et~al.(2022)Yu, Lu, Meng, and Karniadakis}]{yu2022gradient}
\bibinfo{author}{J.~Yu}, \bibinfo{author}{L.~Lu}, \bibinfo{author}{X.~Meng},
  \bibinfo{author}{G.~E. Karniadakis},
\newblock \bibinfo{title}{Gradient-enhanced physics-informed neural networks
  for forward and inverse pde problems},
\newblock \bibinfo{journal}{Computer Methods in Applied Mechanics and
  Engineering} \bibinfo{volume}{393} (\bibinfo{year}{2022})
  \bibinfo{pages}{114823}.
\bibitem[{Yu et~al.(2018)}]{yu2018deep}
\bibinfo{author}{B.~Yu}, et~al.,
\newblock \bibinfo{title}{The deep ritz method: a deep learning-based numerical
  algorithm for solving variational problems},
\newblock \bibinfo{journal}{Communications in Mathematics and Statistics}
  \bibinfo{volume}{6} (\bibinfo{year}{2018}) \bibinfo{pages}{1--12}.
\bibitem[{Liao and Ming(2019)}]{liao2019deep}
\bibinfo{author}{Y.~Liao}, \bibinfo{author}{P.~Ming},
\newblock \bibinfo{title}{Deep nitsche method: Deep ritz method with essential
  boundary conditions},
\newblock \bibinfo{journal}{arXiv preprint arXiv:1912.01309}
  (\bibinfo{year}{2019}).
\bibitem[{Sun et~al.(2023)Sun, Liu, Wang, Yao, and Zheng}]{sun2023binn}
\bibinfo{author}{J.~Sun}, \bibinfo{author}{Y.~Liu}, \bibinfo{author}{Y.~Wang},
  \bibinfo{author}{Z.~Yao}, \bibinfo{author}{X.~Zheng},
\newblock \bibinfo{title}{Binn: A deep learning approach for computational
  mechanics problems based on boundary integral equations},
\newblock \bibinfo{journal}{Computer Methods in Applied Mechanics and
  Engineering} \bibinfo{volume}{410} (\bibinfo{year}{2023})
  \bibinfo{pages}{116012}.
\bibitem[{Sukumar and Srivastava(2022)}]{sukumar2022exact}
\bibinfo{author}{N.~Sukumar}, \bibinfo{author}{A.~Srivastava},
\newblock \bibinfo{title}{Exact imposition of boundary conditions with distance
  functions in physics-informed deep neural networks},
\newblock \bibinfo{journal}{Computer Methods in Applied Mechanics and
  Engineering} \bibinfo{volume}{389} (\bibinfo{year}{2022})
  \bibinfo{pages}{114333}.
\bibitem[{Sheng and Yang(2021)}]{sheng2021pfnn}
\bibinfo{author}{H.~Sheng}, \bibinfo{author}{C.~Yang},
\newblock \bibinfo{title}{Pfnn: A penalty-free neural network method for
  solving a class of second-order boundary-value problems on complex
  geometries},
\newblock \bibinfo{journal}{Journal of Computational Physics}
  \bibinfo{volume}{428} (\bibinfo{year}{2021}) \bibinfo{pages}{110085}.
\bibitem[{Wang et~al.(2021)Wang, Wang, and Perdikaris}]{wang2021eigenvector}
\bibinfo{author}{S.~Wang}, \bibinfo{author}{H.~Wang},
  \bibinfo{author}{P.~Perdikaris},
\newblock \bibinfo{title}{On the eigenvector bias of fourier feature networks:
  From regression to solving multi-scale pdes with physics-informed neural
  networks},
\newblock \bibinfo{journal}{Computer Methods in Applied Mechanics and
  Engineering} \bibinfo{volume}{384} (\bibinfo{year}{2021})
  \bibinfo{pages}{113938}.
\bibitem[{Ng et~al.(2024)Ng, Wang, and Lai}]{ng2024spectrum}
\bibinfo{author}{J.~Ng}, \bibinfo{author}{Y.~Wang}, \bibinfo{author}{C.-Y.
  Lai},
\newblock \bibinfo{title}{Spectrum-informed multistage neural networks:
  Multiscale function approximators of machine precision},
\newblock \bibinfo{journal}{arXiv preprint arXiv:2407.17213}
  (\bibinfo{year}{2024}).
\bibitem[{Dwivedi and Srinivasan(2020)}]{dwivedi2020physics}
\bibinfo{author}{V.~Dwivedi}, \bibinfo{author}{B.~Srinivasan},
\newblock \bibinfo{title}{Physics informed extreme learning machine (pielm)--a
  rapid method for the numerical solution of partial differential equations},
\newblock \bibinfo{journal}{Neurocomputing} \bibinfo{volume}{391}
  (\bibinfo{year}{2020}) \bibinfo{pages}{96--118}.
\bibitem[{Dong and Li(2021)}]{dong2021local}
\bibinfo{author}{S.~Dong}, \bibinfo{author}{Z.~Li},
\newblock \bibinfo{title}{Local extreme learning machines and domain
  decomposition for solving linear and nonlinear partial differential
  equations},
\newblock \bibinfo{journal}{Computer Methods in Applied Mechanics and
  Engineering} \bibinfo{volume}{387} (\bibinfo{year}{2021})
  \bibinfo{pages}{114129}.
\bibitem[{Dong and Yang(2022)}]{dong2022numerical}
\bibinfo{author}{S.~Dong}, \bibinfo{author}{J.~Yang},
\newblock \bibinfo{title}{Numerical approximation of partial differential
  equations by a variable projection method with artificial neural networks},
\newblock \bibinfo{journal}{Computer Methods in Applied Mechanics and
  Engineering} \bibinfo{volume}{398} (\bibinfo{year}{2022})
  \bibinfo{pages}{115284}.
\bibitem[{Chi et~al.(2024)Chi, Chen, and Yang}]{chi2024random}
\bibinfo{author}{X.~Chi}, \bibinfo{author}{J.~Chen}, \bibinfo{author}{Z.~Yang},
\newblock \bibinfo{title}{The random feature method for solving interface
  problems},
\newblock \bibinfo{journal}{Computer Methods in Applied Mechanics and
  Engineering} \bibinfo{volume}{420} (\bibinfo{year}{2024})
  \bibinfo{pages}{116719}.
\bibitem[{Chen et~al.(2022)Chen, Chi, Yang et~al.}]{chen2022bridging}
\bibinfo{author}{J.~Chen}, \bibinfo{author}{X.~Chi}, \bibinfo{author}{Z.~Yang},
  et~al.,
\newblock \bibinfo{title}{Bridging traditional and machine learning-based
  algorithms for solving pdes: the random feature method},
\newblock \bibinfo{journal}{J Mach Learn} \bibinfo{volume}{1}
  (\bibinfo{year}{2022}) \bibinfo{pages}{268--298}.
\bibitem[{Sun et~al.(2024)Sun, Dong, and Wang}]{sun2024local}
\bibinfo{author}{J.~Sun}, \bibinfo{author}{S.~Dong}, \bibinfo{author}{F.~Wang},
\newblock \bibinfo{title}{Local randomized neural networks with discontinuous
  galerkin methods for partial differential equations},
\newblock \bibinfo{journal}{Journal of Computational and Applied Mathematics}
  \bibinfo{volume}{445} (\bibinfo{year}{2024}) \bibinfo{pages}{115830}.
\bibitem[{Shang et~al.(2023)Shang, Wang, and Sun}]{shang2023randomized}
\bibinfo{author}{Y.~Shang}, \bibinfo{author}{F.~Wang},
  \bibinfo{author}{J.~Sun},
\newblock \bibinfo{title}{Randomized neural network with petrov--galerkin
  methods for solving linear and nonlinear partial differential equations},
\newblock \bibinfo{journal}{Communications in Nonlinear Science and Numerical
  Simulation} \bibinfo{volume}{127} (\bibinfo{year}{2023})
  \bibinfo{pages}{107518}.
\bibitem[{Li and Wang(2025)}]{li2025local}
\bibinfo{author}{Y.~Li}, \bibinfo{author}{F.~Wang},
\newblock \bibinfo{title}{Local randomized neural networks with finite
  difference methods for interface problems},
\newblock \bibinfo{journal}{Journal of Computational Physics}
  (\bibinfo{year}{2025}) \bibinfo{pages}{113847}.
\bibitem[{Li and Talwalkar(2020)}]{li2020random}
\bibinfo{author}{L.~Li}, \bibinfo{author}{A.~Talwalkar},
\newblock \bibinfo{title}{Random search and reproducibility for neural
  architecture search},
\newblock in: \bibinfo{booktitle}{Uncertainty in artificial intelligence},
  \bibinfo{organization}{PMLR}, \bibinfo{year}{2020}, pp.
  \bibinfo{pages}{367--377}.
\bibitem[{Linghu et~al.(2024)Linghu, Dong, Gao, and Nie}]{linghu2024self}
\bibinfo{author}{J.~Linghu}, \bibinfo{author}{H.~Dong},
  \bibinfo{author}{W.~Gao}, \bibinfo{author}{Y.~Nie},
\newblock \bibinfo{title}{Self-optimization wavelet-learning method for
  predicting nonlinear thermal conductivity of highly heterogeneous materials
  with randomly hierarchical configurations},
\newblock \bibinfo{journal}{Computer Physics Communications}
  \bibinfo{volume}{295} (\bibinfo{year}{2024}) \bibinfo{pages}{108969}.
\bibitem[{Zoph and Le(2016)}]{zoph2016neural}
\bibinfo{author}{B.~Zoph}, \bibinfo{author}{Q.~V. Le},
\newblock \bibinfo{title}{Neural architecture search with reinforcement
  learning},
\newblock \bibinfo{journal}{arXiv preprint arXiv:1611.01578}
  (\bibinfo{year}{2016}).
\bibitem[{Wang and Zhong(2024)}]{wang2024pinn}
\bibinfo{author}{Y.~Wang}, \bibinfo{author}{L.~Zhong},
\newblock \bibinfo{title}{Nas-pinn: Neural architecture search-guided
  physics-informed neural network for solving pdes},
\newblock \bibinfo{journal}{Journal of Computational Physics}
  \bibinfo{volume}{496} (\bibinfo{year}{2024}) \bibinfo{pages}{112603}.
\bibitem[{Escapil-Inchausp{\'e} and Ruz(2023)}]{escapil2023hyper}
\bibinfo{author}{P.~Escapil-Inchausp{\'e}}, \bibinfo{author}{G.~A. Ruz},
\newblock \bibinfo{title}{Hyper-parameter tuning of physics-informed neural
  networks: Application to helmholtz problems},
\newblock \bibinfo{journal}{Neurocomputing} \bibinfo{volume}{561}
  (\bibinfo{year}{2023}) \bibinfo{pages}{126826}.
\bibitem[{Zhang and Yang(2024)}]{zhang2024discovering}
\bibinfo{author}{B.~Zhang}, \bibinfo{author}{C.~Yang},
\newblock \bibinfo{title}{Discovering physics-informed neural networks model
  for solving partial differential equations through evolutionary computation},
\newblock \bibinfo{journal}{Swarm and Evolutionary Computation}
  \bibinfo{volume}{88} (\bibinfo{year}{2024}) \bibinfo{pages}{101589}.
\bibitem[{Nabian et~al.(2021)Nabian, Gladstone, and
  Meidani}]{nabian2021efficient}
\bibinfo{author}{M.~A. Nabian}, \bibinfo{author}{R.~J. Gladstone},
  \bibinfo{author}{H.~Meidani},
\newblock \bibinfo{title}{Efficient training of physics-informed neural
  networks via importance sampling},
\newblock \bibinfo{journal}{Computer-Aided Civil and Infrastructure
  Engineering} \bibinfo{volume}{36} (\bibinfo{year}{2021})
  \bibinfo{pages}{962--977}.
\bibitem[{Wu et~al.(2023)Wu, Zhu, Tan, Kartha, and Lu}]{wu2023comprehensive}
\bibinfo{author}{C.~Wu}, \bibinfo{author}{M.~Zhu}, \bibinfo{author}{Q.~Tan},
  \bibinfo{author}{Y.~Kartha}, \bibinfo{author}{L.~Lu},
\newblock \bibinfo{title}{A comprehensive study of non-adaptive and
  residual-based adaptive sampling for physics-informed neural networks},
\newblock \bibinfo{journal}{Computer Methods in Applied Mechanics and
  Engineering} \bibinfo{volume}{403} (\bibinfo{year}{2023})
  \bibinfo{pages}{115671}.
\bibitem[{Tancik et~al.(2020)Tancik, Srinivasan, Mildenhall, Fridovich-Keil,
  Raghavan, Singhal, Ramamoorthi, Barron, and Ng}]{tancik2020fourier}
\bibinfo{author}{M.~Tancik}, \bibinfo{author}{P.~Srinivasan},
  \bibinfo{author}{B.~Mildenhall}, \bibinfo{author}{S.~Fridovich-Keil},
  \bibinfo{author}{N.~Raghavan}, \bibinfo{author}{U.~Singhal},
  \bibinfo{author}{R.~Ramamoorthi}, \bibinfo{author}{J.~Barron},
  \bibinfo{author}{R.~Ng},
\newblock \bibinfo{title}{Fourier features let networks learn high frequency
  functions in low dimensional domains},
\newblock \bibinfo{journal}{Advances in neural information processing systems}
  \bibinfo{volume}{33} (\bibinfo{year}{2020}) \bibinfo{pages}{7537--7547}.
\bibitem[{Melin et~al.(2013)Melin, Olivas, Castillo, Valdez, Soria, and
  Valdez}]{melin2013optimal}
\bibinfo{author}{P.~Melin}, \bibinfo{author}{F.~Olivas},
  \bibinfo{author}{O.~Castillo}, \bibinfo{author}{F.~Valdez},
  \bibinfo{author}{J.~Soria}, \bibinfo{author}{M.~Valdez},
\newblock \bibinfo{title}{Optimal design of fuzzy classification systems using
  pso with dynamic parameter adaptation through fuzzy logic},
\newblock \bibinfo{journal}{Expert Systems with Applications}
  \bibinfo{volume}{40} (\bibinfo{year}{2013}) \bibinfo{pages}{3196--3206}.
\bibitem[{Lim and Isa(2014)}]{lim2014adaptive}
\bibinfo{author}{W.~H. Lim}, \bibinfo{author}{N.~A.~M. Isa},
\newblock \bibinfo{title}{An adaptive two-layer particle swarm optimization
  with elitist learning strategy},
\newblock \bibinfo{journal}{Information Sciences} \bibinfo{volume}{273}
  (\bibinfo{year}{2014}) \bibinfo{pages}{49--72}.
\bibitem[{Koch(1904)}]{koch1904courbe}
\bibinfo{author}{H.~Koch},
\newblock \bibinfo{title}{Sur une courbe continue sans tangente, obtenue par
  une construction g{\'e}om{\'e}trique {\'e}l{\'e}mentaire},
\newblock \bibinfo{journal}{Arkiv for Matematik, Astronomi och Fysik}
  \bibinfo{volume}{1} (\bibinfo{year}{1904}) \bibinfo{pages}{681--704}.
\bibitem[{Evans(2022)}]{evans2022partial}
\bibinfo{author}{L.~C. Evans}, \bibinfo{title}{Partial differential equations},
  volume~\bibinfo{volume}{19}, \bibinfo{publisher}{American Mathematical
  Society}, \bibinfo{year}{2022}.
\bibitem[{Timoshenko and Woinowsky-Krieger(1959)}]{timoshenko1959theory}
\bibinfo{author}{S.~Timoshenko}, \bibinfo{author}{S.~Woinowsky-Krieger},
\newblock \bibinfo{title}{Theory of plates and shells}  (\bibinfo{year}{1959}).
\bibitem[{Reddy(2006)}]{reddy2006theory}
\bibinfo{author}{J.~N. Reddy}, \bibinfo{title}{Theory and analysis of elastic
  plates and shells}, \bibinfo{publisher}{CRC press}, \bibinfo{year}{2006}.
\bibitem[{Roy et~al.(2023)Roy, Bose, Sundararaghavan, and
  Arr{\'o}yave}]{roy2023deep}
\bibinfo{author}{A.~M. Roy}, \bibinfo{author}{R.~Bose},
  \bibinfo{author}{V.~Sundararaghavan}, \bibinfo{author}{R.~Arr{\'o}yave},
\newblock \bibinfo{title}{Deep learning-accelerated computational framework
  based on physics informed neural network for the solution of linear
  elasticity},
\newblock \bibinfo{journal}{Neural Networks} \bibinfo{volume}{162}
  (\bibinfo{year}{2023}) \bibinfo{pages}{472--489}.
\bibitem[{Sadd(2009)}]{sadd2009elasticity}
\bibinfo{author}{M.~H. Sadd}, \bibinfo{title}{Elasticity: theory, applications,
  and numerics}, \bibinfo{publisher}{Academic Press}, \bibinfo{year}{2009}.

\end{thebibliography}



	
	
	

\end{document}